\documentclass{ws-procs11x85}

\usepackage{ws-procs-thm}  % comment this line when `amsthm / theorem / ntheorem` package is used
\let\captionbox\relax
\usepackage[caption]{subfig} % This loads caption without overwriting existing \caption
\usepackage{amsmath}% http://ctan.org/pkg/amsmath
\makeatletter
\newcommand{\xleftrightarrow}[2][]{\ext@arrow 3359\leftrightarrowfill@{#1}{#2}}
\newcommand{\xdashrightarrow}[2][]{\ext@arrow 0359\rightarrowfill@@{#1}{#2}}
\newcommand{\xdashleftarrow}[2][]{\ext@arrow 3095\leftarrowfill@@{#1}{#2}}
\newcommand{\xdashleftrightarrow}[2][]{\ext@arrow 3359\leftrightarrowfill@@{#1}{#2}}
\def\rightarrowfill@@{\arrowfill@@\relax\relbar\rightarrow}
\def\leftarrowfill@@{\arrowfill@@\leftarrow\relbar\relax}
\def\leftrightarrowfill@@{\arrowfill@@\leftarrow\relbar\rightarrow}
\def\arrowfill@@#1#2#3#4{%
  $\m@th\thickmuskip0mu\medmuskip\thickmuskip\thinmuskip\thickmuskip
   \relax#4#1
   \xleaders\hbox{$#4#2$}\hfill
   #3$%
}
\makeatother
\usepackage{adjustbox}
\usepackage{cancel}
\usepackage{array}
\usepackage{makecell}
\usepackage{tikz}
\usepackage{float}
\usepackage{graphicx}
\usepackage{latexsym}
\usepackage{commath}
\usepackage{float}
\usepackage{ifthen}
\usepackage{amsbsy}
\usepackage{dsfont}
\usepackage{bbm}
\usepackage{graphicx}
\usepackage{algorithm}
\usepackage{algpseudocode}
\usepackage{setspace}
\usepackage{esint}
\usepackage{comment}
\usepackage{mathtools}
\usepackage{xcolor}

\let\Pr\undefined

\DeclareMathOperator{\Pr}{Pr}

\DeclareMathOperator*{\argmin}{arg\,min} % * Places subscript directly under operator

\def\ddefloop#1{\ifx\ddefloop#1\else\ddef{#1}\expandafter\ddefloop\fi}
\def\ddef#1{\expandafter\def\csname bb#1\endcsname{\ensuremath{\mathbb{#1}}}}
\ddefloop ABCDEFGHIJKLMNOPQRSTUVWXYZ\ddefloop
\def\ddefloop#1{\ifx\ddefloop#1\else\ddef{#1}\expandafter\ddefloop\fi}
\def\ddef#1{\expandafter\def\csname b#1\endcsname{\ensuremath{\mathbf{#1}}}}
\ddefloop ABCDEFGHIJKLMNOPQRSTUVWXYZ\ddefloop
\def\ddef#1{\expandafter\def\csname c#1\endcsname{\ensuremath{\mathcal{#1}}}}
\ddefloop ABCDEFGHIJKLMNOPQRSTUVWXYZ\ddefloop
\def\ddef#1{\expandafter\def\csname h#1\endcsname{\ensuremath{\widehat{#1}}}}
\ddefloop ABCDEFGHIJKLMNOPQRSTUVWXYZ\ddefloop
\def\ddef#1{\expandafter\def\csname hc#1\endcsname{\ensuremath{\widehat{\mathcal{#1}}}}}
\ddefloop ABCDEFGHIJKLMNOPQRSTUVWXYZ\ddefloop
\def\ddef#1{\expandafter\def\csname t#1\endcsname{\ensuremath{\widetilde{#1}}}}
\ddefloop ABCDEFGHIJKLMNOPQRSTUVWXYZ\ddefloop
\def\ddef#1{\expandafter\def\csname tc#1\endcsname{\ensuremath{\widetilde{\mathcal{#1}}}}}
\ddefloop ABCDEFGHIJKLMNOPQRSTUVWXYZ\ddefloop

\usepackage{prettyref}

\usepackage{nicefrac}
\usepackage{booktabs}
\usepackage{hyperref}
\newcommand{\ci}{\perp \!\!\! \perp}
\usepackage{parskip}
\usepackage{color-edits}
\addauthor{vs}{blue}
\addauthor{vsr}{red}
\addauthor{hl}{red}
\addauthor{kl}{red}

\usepackage{makecell}

\usepackage{url}
\usepackage{hyperref}
\newcommand{\kibitz}[2]{\ifnum\Comments=1{\color{#1}{#2}}\fi}

\newcommand{\E}{\mathbb{E}}

\newcommand{\R}{\mathbb{R}}

\newcommand{\D}{\mathcal D}

\renewcommand{\Pr}{\ensuremath{\mathrm{Pr}}}

\newcommand{\gen}{\ensuremath{\text{gen}}}
 
\newcommand{\cov}{\ensuremath{\text{Cov}}}

\newcommand{\ba}{\begin{array}}
\newcommand{\ea}{\end{array}}
\newcommand{\bs}{\begin{align}\begin{split}\nonumber}
\newcommand{\bsnumber}{\begin{align}\begin{split}}
\newcommand{\es}{\end{split}\end{align}}

\def\defeq{\triangleq} %
 
\def\balign#1\ealign{\begin{align}#1\end{align}}
\def\balignat#1\ealign{\begin{alignat}#1\end{alignat}}
\def\bitemize#1\eitemize{\begin{itemize}#1\end{itemize}}
\def\benumerate#1\eenumerate{\begin{enumerate}#1\end{enumerate}}
\newenvironment{talign}
 {\let\displaystyle\textstyle\csname align\endcsname}
 {\endalign}
\def\balignt#1\ealignt{\begin{talign}#1\end{talign}}%
\usepackage{graphicx}
\usepackage{float}

\usepackage{tikz}
\usetikzlibrary{positioning, arrows.meta}
\usepackage[utf8]{inputenc}
\usepackage{tikz}
\usetikzlibrary{positioning, arrows.meta}
\usetikzlibrary{positioning, arrows.meta}
\usepackage{longtable}
\begin{document}

\title{Detecting clinician implicit biases in diagnoses using proximal causal inference}

\author{Kara Liu$^\dag$, Russ Altman, Vasilis Syrgkanis}

\address{Computer Science Department, Stanford University,\\
Stanford, CA 94305, USA\\
$^\dag$E-mail: karaliu@stanford.edu\\
}

\begin{abstract}
Clinical decisions to treat and diagnose patients are affected by implicit biases formed by racism, ableism, sexism, and other stereotypes. These biases reflect broader systemic discrimination in healthcare and risk marginalizing already disadvantaged groups. Existing methods for measuring implicit biases require controlled randomized testing and only capture individual attitudes rather than outcomes. However, the "big-data" revolution has led to the availability of large observational medical datasets, like EHRs and biobanks, that provide the opportunity to investigate discrepancies in patient health outcomes. In this work, we propose a causal inference approach to detect the effect of clinician implicit biases on patient outcomes in large-scale medical data. Specifically, our method uses proximal mediation to disentangle pathway-specific effects of a patient's sociodemographic attribute on a clinician's diagnosis decision. We test our method on real-world data from the UK Biobank. Our work can serve as a tool that initiates conversation and brings awareness to unequal health outcomes caused by implicit biases.\footnote{Our method is available at \url{https://github.com/syrgkanislab/hidden_mediators}}

\end{abstract}
\vspace{-.2cm}

\keywords{Implicit bias, proximal causal inference, fairness, healthcare}

% required, do-not-remove
\copyrightinfo{\copyright\ 2024 The Authors. Open Access chapter published by World Scientific Publishing Company and distributed under the terms of the Creative Commons Attribution Non-Commercial (CC BY-NC) 4.0 License.}

\vspace{-.2cm}

\section{Introduction}
Implicit bias refers to unconscious and automatic associations that affect how we perceive, evaluate, and interact with people from different social groups \cite{holroyd2016heterogeneity}. Outside of mere cognitive distortions, these biases held by healthcare professionals influence clinical decisions and alter a patient’s quality of care. Implicit biases have been shown to be both harmful and pervasive in modern-day medicine, exacerbating existing inequality in the treatment and health outcomes of marginalized groups \cite{vela2022eliminating, gopal2021implicit}. For instance, unconscious attitudes held by clinicians result in disparate outcomes where women are less likely then men to be diagnosed with myocardial infarction \cite{gopal2021implicit}, Black women in the UK and US experience higher maternal mortality than White women \cite{saluja2021implicit}, and low socioeconomic (SES) and non-White patients receive sub-optimal pain management treatment compared to high SES and White patients \cite{anastas2020unique, sabin2012influence}.

The recent integration of machine learning (ML) models into clinical decision-making has highlighted the prevalence of biases in medicine. By replicating the patterns from real-world medical data, ML models perpetuate and risk amplifying existing disparities in the medical treatment of marginalized groups \cite{ueda2024fairness, pfohl2021empirical}. While much attention has been given to the statistical objectives of fairness and the development of fair models, there has been comparatively less focus on investigating the biases present in the underlying data. A method capable of detecting implicit clinician bias in observational datasets would prevent ML models from unintentionally perpetuating biased decisions.

However, measuring implicit bias is challenging. Existing methods for quantifying implicit bias rely on the Implicit Association Test (IAT) \cite{iat} and randomized psychological experiments like affective priming \cite{arif2021gaps}. While these tests are useful for initiating dialogue, they only provide a snapshot of individual clinician attitudes and do not guarantee a causal link to behavior or larger systemic discrepancies of care \cite{arif2021gaps}. %Furthermore, these experiments require organized administration that might not be scalable across large institutions. 

In this work, we propose a computational tool to detect clinician implicit bias in observational datasets by measuring the causal effect of patient attributes, like race, SES, and other social determinants of health (SDoH), on medical diagnoses. By decomposing the causal effect into two pathways, we can separate the \textit{biological effect} (the influence of a demographic attribute on diagnosis as mediated by valid biological traits) from the \textit{implicit bias effect} (how the patient's attribute affects a clinician's judgement independent of their actual health state). As it is unlikely to observe a patient's true health state, we use observed medical data as proxies using proximal causal inference \cite{tchetgen2020introduction}. To estimate the effect of implicit bias, we propose a novel proximal mediation method that guarantees identifiability under several assumptions. Using real patient data from the UK Biobank, we validate our method can robustly detect several clinician implicit biases identified from prior works. We aim for the proposed method to serve as a bias-detection tool in dataset audits and initiate discussion on reducing systemic discrimination in medicine.

\textit{Disclaimer:} While we use the UK Biobank data for method validation, we emphasize that this work is not a commentary on specific examples of discrimination within the UK healthcare system. Additionally, it is crucial to clarify that our method of estimating implicit bias is not intended to target clinicians but rather reflect on clinician behaviors within the context of discriminatory healthcare systems.
\vspace{-.3cm}
\section{Method}
\vspace{-.05cm}
\subsection{Background} 
\vspace{-.05cm}
\subsubsection{Overview}

According to the Hippocratic Oath, clinicians should base their diagnostic decisions on each patient's history and current health status, unaffected by biases or stereotypes of the perceived patient identity. However, even in the ideal scenario of unbiased treatment, patient sociodemographic attributes will still influence diagnosis. Attributes including race, sex, or SES have been shown to influence a patient's true health status via mechanisms like genetics, lifestyle, and weathering from systemic oppression \cite{forde2019weathering, williams2018stress, cui2021income}. These biologically-mediated effects increase the risk of certain medical conditions. For instance, patients from lower SES backgrounds experience higher levels of stress and reduced access to healthcare, increasing their risk of cardiovascular disease \cite{minhas2023family}. In light of these known biological influences, the causal effect of a patient's sociodemographic attribute on their diagnosis by a clinician is therefore comprised of two pathway effects: the \textit{biological effect} and the \textit{implicit bias effect}, the latter referring to the clinician's subjective biases of the sociodemographic attribute not mediated through the patient's actual health state. 

\begin{figure}[H]
\centering
\centerline{\includegraphics[width=.75\textwidth]{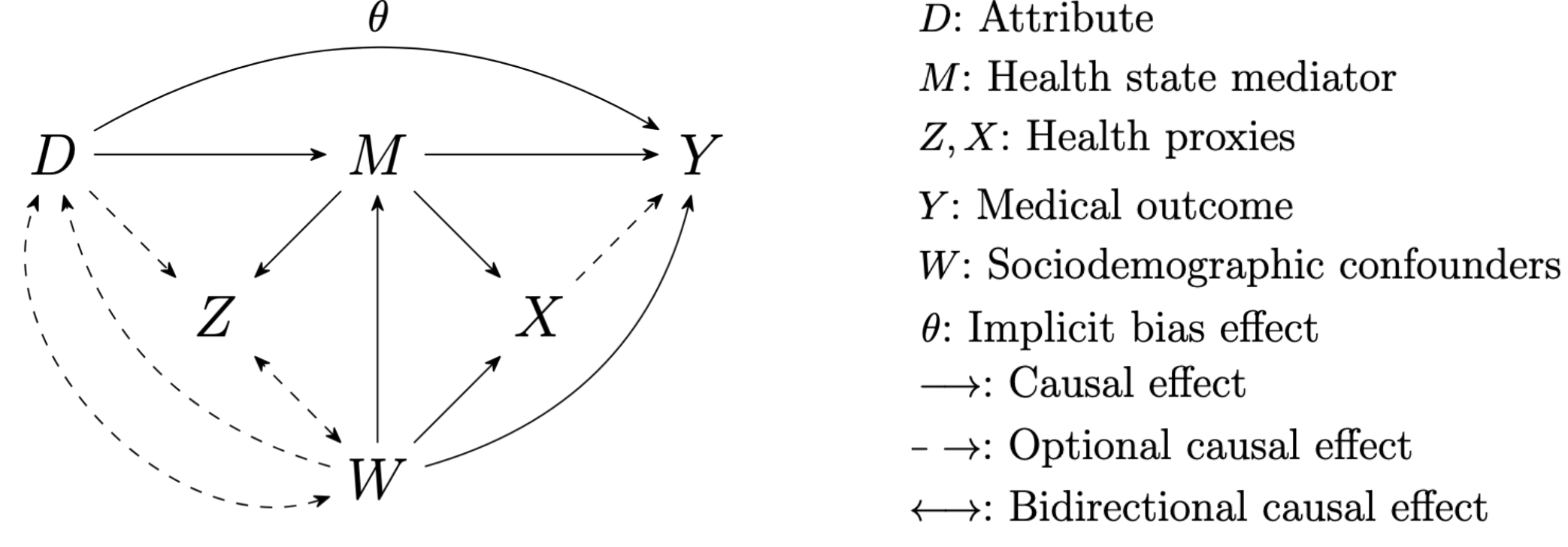}}

\caption{Assumed causal graph.}
\label{fig:cg}
\end{figure}

We present the assumed causal relationships between variables as the directed acyclic graph (DAG) in Figure \ref{fig:cg}. Dashed arrows denote optional edges, and bi-directional arrows denote indirect confounding paths through latent variables. Let $D$ be the binary sociodemographic attribute and $Y$ the diagnosis decision we wish to measure implicit bias with respect to. $M$ represents the latent variables encoding a patient's true underlying health state. However, as $M$ is typically unknown, we instead observe $Z$ and $X$ as multivariate proxies of $M$. We differentiate these proxies into the variables $Z$ that do not affect the diagnostic decision $Y$ but could be affected by the attribute $D$; and the variables $X$ which are not directly affected by the attribute $D$ but can influence diagnosis $Y$. For example, $X$ could be recent lab reports a clinician uses to make their diagnosis, and $Z$ could be a patient's survey responses to a sleep questionnaire (assuming the survey does not influence the clinician's diagnosis). Finally, let $W$ be sociodemographic confounders to control for.

We can now reframe \textit{biological} and \textit{implicit bias effects} using pathway causal effects. The \textit{biological effect} of attribute $D$ on diagnosis $Y$ is the indirect effect as mediated through the true underlying health state $M$: $D \rightarrow M \rightarrow Y$. The \textit{implicit bias effect} we wish to measure is the direct effect of $D \rightarrow Y$ that flows through the edge $\theta$ and is defined as the residual of the biological effect. We formally define bias in terms of controlled direct effects in Equation \eqref{eq:cde}.
\vspace{-.3cm}

\vspace{-1.1cm}

\subsubsection{Related works}
Measuring implicit biases requires detecting the unconscious and automatic attitudes that shape behavior. The predominant method for implicit bias measurement thus far has been the Implicit Association Test (IAT)\cite{iat}, a questionnaire developed in 1998 intended to measure group association through word categorization. To capture clinician biases, several works have linked clinician attitudes via their IAT score to behavioral manifestation\cite{arif2021gaps, greenwald2022implicit}. Other methods for detecting implicit clinician bias include affective priming, which measures biased associations after stimulus priming; and the assumption method, which surveys clinicians' decisions after reading patient vignettes \cite{fitzgerald2017implicit}. While association tests like the IAT have been integral in bringing awareness to medical biases, they are criticized for their arbitrary scoring system, inability to predict real-world patient outcomes, and context-dependency \cite{arif2021gaps,gopal2021implicit, greenwald2022implicit,fitzgerald2017implicit}. Furthermore, administering these controlled tests in every clinical encounter is impractical and unscalable.

Computational methods present a promising and scalable alternative for detecting implicit bias in real-world medical data. While the field of ML fairness has explored bias detection, the focus has been on identifying and mitigating bias in models rather than the data \cite{ueda2024fairness}. In causal inference, disentangling a causal effect into natural indirect and direct pathway effects has led to methods that control for ``fair'' and ``unfair'' causal pathway effects. [\citen{path:chiappa,path:nabi,path:zhang, cause:naimi}] propose metrics for measuring fair pathway influence on outcomes and develop methods that mitigate the effect of unfair pathways on the predicted outcome. [\citen{jun2023quantifying}] leveraged the Fairness-Aware Causal paThs (FACTS) \cite{fact} algorithm to quantify disparate pathway influence of SDoH attributes on mortality using real-world health data. While these methods recognize an attribute's influence on an outcome contains both fair and unfair effects, prior works are limited to simple scenarios where all variables are known and observed. Our work is the first to extend pathway inference to large-scale observational data with potentially unobserved variables.

Finally, a few recent methods have explored proximal mediation analysis, where pathway effects can be measured despite unobserved mediators by using proxy variables \cite{cause:ghassami21,cause:ghassami23}. However, by relying on natural direct and indirect pathway effects, these works rely on more stringent assumptions, require learning complicated bridge functions, and limit their analysis to simple datasets. In comparison, our method makes several relaxations that enable application to observational data. First, we identify controlled instead of natural effects, which presents an equally good measurement of a biased decision yet lends to a much simpler statistical problem. Additionally, we assume partially linear equations instead of requiring the identification of a complex bridge function. Finally, we do not require uniqueness of the parameters unrelated to implicit bias (i.e., the nuisance parameters for the outcome bridge function). These relaxations enable our approach to be effective at analyzing large-scale real-world medical data.
\vspace{-.4cm}
\subsection{Our method}
Our goal is to identify and estimate the following controlled direct effect:
\begin{align}\label{eq:cde}
\theta =~& \int_{m,w} \E[Y(1, m) - Y(0, m)\mid W=w] \text{ }p(m, w) \text{ }dm\text{ } dw
\end{align}
where $Y(d, m)$ is the potential (or counterfactual) outcome when we intervene on the attribute $D$ and the mediator $M$ and set them to values $(d, m)$; and $p(m, w)$ is the natural probability distribution in the data. If the controlled direct effect is nonzero, then there exists a direct influence of the attribute $D$ on the outcome $Y$, which is evidence of implicit bias. 

If we observe $M$, the above controlled direct effect can be identified by a simple g-formula that ``controls'' for $M$ and $W$: $\theta = \E[\E[Y\mid D=1, M, W] - \E[Y\mid D=0, M, W]]$. Unfortunately, this equation is intractable if $M$ is unobserved. However, we show that under a few reasonable assumptions the controlled direct effect is still identifiable. 

\begin{theorem}[Identification]\label{thm:id}\footnote{We present more intuitive interpretations of each theorem and lemma in the Appendix.}
    Consider a non-parametric structural causal model (SCM) that respects the causal relationships encoded in Figure~\ref{fig:cg} (see Appendix \ref{appendix:cg_ass}) and assume there exists a ``bridge function'' $q$ that solves $\E[Y\mid D, M, W] = \E[q(D, X, W)\mid D, M, W]$. Then $q$ also solves the Non-Parametric Instrumental Variable (NPIV) problem defined by the set of conditional moment restrictions
    \begin{align}\label{eqn:npiv}
     \E[Y - q(D, X, W) \mid D, Z, W] = 0
    \end{align}
    and the controlled direct effect can be identified as $ \theta = \E[q(1, X, W) - q(0, X, W)]$.
\end{theorem}

Identifying parameters $\theta$ using a bridge function $q$ (where $q$ also solves an NPIV problem) has been extensively studied in proximal causal inference literature \cite{chen2012estimation,ai2003efficient,ai2012semiparametric,lewis2018adversarial,dikkala2020minimax,bennett2019deep,bennett2020variational,miao2018a,cui2020semiparametric,bennett2022inference,bennett2023source,bennett2023minimax,zhang2023proximal}. However, these approaches rely on solving saddle-point problems with adversarial training or require learning conditional density functions, both of which are statistically daunting. 

% For example, it has been shown that $\theta$ is point identified even when the outcome bridge function is not unique, so long as there exists a solution to a dual instrumental variable regression problem that defines what is referred to as a treatment bridge function. Moreover, modern machine learning techniques have been developed for the estimation of solutions to the NPIV problem and to conducting inference on functionals of the solution to NPIV problems \cite{bennett2022inference,bennett2023source,bennett2023minimax,zhang2023proximal}.

We can avoid these difficult statistical tasks if we assume that the bridge function is partially linear in $D$ and $X$. The following lemma shows that partial linearity of $q$ is implied by a more primitive assumption of partial linearity of two other functions (proof in Appendix \ref{appendix:lemma1}).

\begin{lemma}[Identification under partial linearity]\label{lem:plr}
    Consider a non-parametric SCM that respects the constraints encoded in Figure~\ref{fig:cg} and assume that $X$ has dimension $p_X$ at least as large as the dimension $p_M$ of $M$. Moreover, assume that the following functions are partially linear:
\begin{align}\label{eqn:plr}
    \E[Y\mid D, M, X, W] =~& D\,c + M^\top b + X^\top g + f_Y(W)\\
    \E[X\mid M, W] =~& F\, M + f_X(W)
\end{align}
where $F$ is a $p_X \times p_M$ matrix, $b$ is a $p_M$-dimensional vector, $g$ is a $p_X$-dimensional vector and $f_Y, f_X$ are arbitrary non-parametric functions. If we assume the matrix $F$ has full column rank, then there exists a partially linear outcome bridge function 
\begin{align}\label{eqn:linearq}
q(D, X, W) = D\,\theta + X^\top h + f(W)
\end{align}
that satisfies Equation~\eqref{eqn:npiv}, where parameter $h=F^+b + g$\footnote{$F^+$ is the Moore-Penrose pseudoinverse of $F$.} and $\theta=c$.
\end{lemma}

% \begin{lemma}
%     The desired controlled direct effect $\theta$ is equal to the constant $c$.
% \end{lemma}
\vspace{-.2cm}

Under the assumption of partial linearity, we can simplify the estimation problem by first removing the effect of $W$ from all the remaining variables (see Appendix~\ref{app:res-w}), where for any variable $V$ we define the residual $\tilde{V} = V - \E[V\mid W]$. Partial linearity  of $q$ from Equation \eqref{eqn:linearq}, when combined with the NPIV Equation \eqref{eqn:npiv}, implies that $\theta$ can be identified using linear instrumental variable (IV) regression where $(\tilde{Z};\tilde{D})$\footnote{We denote $(A;B)$ to be concatenation of vectors $A$ and $B$.} are the instruments and $(\tilde{X}; \tilde{D})$ are the treatments:
\begin{align}\label{eq:primal}
    \E\left[(\tilde{Y} - \tilde{X}^\top h - \tilde{D}\,\theta) \begin{pmatrix}\tilde{Z}\\ \tilde{D}\end{pmatrix}\right] =~& 0 \tag{Primal Equation}
\end{align}
\vspace{-.2cm}

Unique identification of $\theta$ seemingly requires unique identification of the other ``nuisance'' parameters like $h$, which might be difficult to achieve as the covariance matrix $\E[(\tilde{X}; \tilde{D})\, (\tilde{Z};\tilde{D})^\top ]$ is usually not full rank\footnote{This could be the case if the number of proxies is much larger than the dimensionality of the latent mediator $M$.}. We invoke and simplify ideas from the recent proximal inference literature\cite{chen2021robust,bennett2022inference} to show that $\theta$ can be point-identified even if $h$ is not. To achieve this, we construct a moment restriction equation that is Neyman orthogonal to the nuisance parameters $h$ but still point-identifies $\theta$, given sufficient quality of the proxy $Z$. Intuitively, we learn a new instrument $V = (\tilde{D} - \gamma^\top  \tilde{Z})$ such that $V$ is uncorrelated with $\tilde{X}$ and thus estimation of $\theta$ is not sensitive to $h$. Existence of such a $\gamma$ is sufficient for point-identification of $\theta$. We provide the proof for the point identification of $\theta$ in Appendix \ref{app:unique-id} and for Neyman orthogonality in \ref{app:neyman}. 
 % This moment also reveals the sufficient conditions for point-identification of $\theta$, which relate to the quality of the treatment proxy $Z$. Crucially, this requires the existence of such a $\gamma$ that achieves de-correlation. One can argue that, under the assumptions of Lemma~\ref{lem:plr}, existence of $\gamma$ is guaranteed if the covariance matrix $\text{Cov}(M, Z)$ has full row rank. The function $\gamma^\top  \tilde{Z}$ relates to the notion of a treatment bridge function\cite{bennett2022inference}, whose existence is known to imply uniqueness of the target parameter $\theta$ in proximal inference\cite{kallus2021causal}. 
\vspace{-.3cm}
\begin{theorem} Let $h_*$ be the minimum norm solution to the \eqref{eq:primal} and assume that the following dual equation also admits 
 a solution $\gamma_*$:
\begin{align}
    \E[\tilde{X}\, (\tilde{D} - \gamma^\top \tilde{Z})] = 0 \tag{Dual Equation}\label{eq:dual}
\end{align}
Furthermore, assume $\E[\tilde{D}\,(\tilde{D}-\gamma_*^\top Z)]\neq 0$. Then the solution $\theta_0$ to the equation:
\begin{equation}\label{eq:finalnew}
    \E[(\tilde{Y} - \tilde{X}^\top h_* - \tilde{D}\, \theta)\, (\tilde{D} - \gamma_*^\top \tilde{Z})] = 0
\end{equation}
uniquely identifies the controlled direct effect $\theta$. Furthermore, this moment restriction is Neyman orthogonal with respect to nuisance parameters $\gamma_*, h_*$. 
\end{theorem}\label{thm2}
\vspace{-.3cm}

Theorem~\ref{thm2} allows us to invoke the general framework of [\citen{chernozhukov2017double}] to construct an estimate and confidence interval for the controlled direct effect $\theta$. The full estimation algorithm is presented in Appendix \ref{app:algo}.
\vspace{-.6cm}
\subsection{Testing and Removing Weak Instruments}\label{main:tests}
Our method for uniquely identifying the controlled direct effect $\theta$ relies on several assumptions, e.g., $(\tilde{Z}; \tilde{D})$ are good instruments for $(\tilde{X}; \tilde{D})$. To assess the validity of these assumptions, we developed a suite of tests that must pass for the estimate $\theta$ to be valid and can be used as validity checks by practitioners. These tests are further described in Appendix \ref{appendix:tests}: 
\begin{enumerate}
    \item \textit{Primal equation violation} - We develop a $\chi^2$-test to check if the primal equation admits a solution, i.e., $\E[(\tilde{Y} - \tilde{X}^\top h_* - \tilde{D}\,\theta_0) (\tilde{Z};\tilde{D})] \approx 0$. Intuitively, violation of the primal test implies either the variables $X$ are insufficient proxies of the health state $M$ or the residual proxy $\tilde{Z}$ has a direct path to $\tilde{Y}$. 
    \item \textit{Dual equation violation} - We develop a $\chi^2$-test to check if the dual equation admits a solution, i.e., $\E[\tilde{X}(\tilde{D}-\gamma_*^\top \tilde{Z})] \approx 0$. Violation of the dual implies the variables $Z$ are insufficient proxies of the health state $M$ or that the residual proxy $\tilde{X}$ has a direct path from $\tilde{D}$. 
    \item \textit{Strength of identification} -  We perform two tests to check if $V = (\tilde{D}-\gamma_*^\top \tilde{Z})$ is a good instrument for (i.e., retains enough information about) $\tilde{D}$. (a) We develop an effective F-test\cite{olea2013robust,andrews2005inference} to check the correlation strength of $V$ with $\tilde{D}$. (b) We develop a z-test to check if the quantity $\E[\tilde{D}(\tilde{D}-\gamma_*^\top \tilde{Z})]$ is substantially bounded away from zero (see assumption in Theorem \ref{thm2}). Intuitively, these tests will fail if the hidden mediator is a very deterministic function of the attribute $D$.
    \item \textit{Proxy covariance rank test} - To ensure the health proxies are sufficiently related, we check the rank of the covariance matrix of $\tilde{X}$ and $\tilde{Z}$ by identifying the number of statistically significant singular values. This rank can be viewed as an upper bound on the dimensionality of the hidden mediator $M$ that we can control for.
    % For this we use a conservative test based on Weyl's theorem\cite{weyl1912asymptotische} and a high probability bound on the Forbenius norm of the error between the sample covariance and the population covariance, using a characterization of the asymptotic distribution of the norm as a sum of $\chi^2$ random variables. kledit: we can put this in the appendix.  
\end{enumerate}
\vspace{-.3cm}
\subsubsection{Proxy selection algorithm}\label{main:proxyrm}
In practice, the initial selection of proxies $X,Z$ may violate key assumptions, which can be detected by the failure of one or more of the aforementioned tests. In Appendix \ref{appendix:proxyrm}, we provide an algorithm for identifying subsets of $X$ and $Z$ that satisfy the necessary assumptions and thus produce valid estimates. This proxy selection algorithm should be performed on a separate dataset from the one used to estimate $\theta$.

\vspace{-.3cm}
\section{Experiments}
\subsection{Data}\label{data}
To validate our approach, we use the UK Biobank, a rich and accessible repository containing genomic, imaging, and tabular health data from over 500,000 patients. Our work uses its tabular data, which includes survey questions and biometrics collected upon an individual's enrollment into the biobank. In addition, several health outcomes, including medical diagnoses via ICD10 codes, have been linked to most patients. We note and discuss the caveats of applying our method to biobank data in Section \ref{discussion}. 
\vspace{-.3cm}

\begin{table}[ht]
\centering
\resizebox{.7\textwidth}{!}{%
\begin{tabular}{@{}llcc@{}}
\toprule
{\textbf{}} & {} & {\textbf{\begin{tabular}[c]{@{}c@{}}Prevalence in \\ UK Biobank \\ (n=502411)\end{tabular}}} & {\textbf{\begin{tabular}[c]{@{}c@{}}Prior works \\ on implicit bias\end{tabular}}} \\ \midrule
{\textbf{Sociodemographic attribute $D$}} & {\textit{Race} - Asian} & {2.4\%} & {\citen{mcmurtry2019discrimination, bougie2019influence, wu2021anti}} \\
 & \textit{Race} - Black & 1.8\% &  \citen{garb2021race, gopal2021implicit, saluja2021implicit, anastas2020unique, taylor2018pain} \\
{\textbf{}} & {\textit{Gender} - Female} & {54.4\%} & {\citen{gopal2021implicit, maserejian2009disparities}} \\
{\textbf{}} & {\textit{Disability status} - On disability allowance} & {6.2\%} & {\citen{mcclendon2021cumulative, vanpuymbrouck2020explicit}} \\
{\textbf{}} & {\textit{Income} - Household income \textless 18,000£} & {20.3\%} & {\citen{anastas2020unique,stepanikova2017perceived, minhas2023family, cui2021income}} \\
{\textbf{}} & {\textit{Education} - No post-secondary education} & {67.3\%} & {\citen{anastas2020unique, stepanikova2017perceived}} \\
{\textbf{}} & {\textit{Weight} - BMI \textgreater 30} & {24.3\%} & {\citen{Phelan2015, fulton2023obesity}} \\
{\textbf{}} & {\textit{Insurance} - Not on private insurance} & {31.4\%} & {\citen{han2015reports}} \\ \bottomrule
{\textbf{Medical diagnosis $Y$}} & {Osteoarthritis} & {18.0\%} & {\citen{taylor2018pain, mcclendon2021cumulative}} \\
{} & {Rheumatoid arthritis} & {1.9\%} & {\citen{mcburney2012racial}} \\
{} & {Chronic kidney disease} & {5.0\%} & {\citen{evans2011race, jenkins2022perspectives}} \\
{} & {Complications during labor} & {2.4\%} & {\citen{gopal2021implicit, saluja2021implicit}} \\
{} & {Heart disease} & {10.7\%} & { \citen{minhas2023family, gopal2021implicit, maserejian2009disparities}} \\
{} & {Depression} & {6.0\%} & {\citen{garb2021race, kim2014racial}} \\
{} & {Melanoma} & {1.2\%} & {\citen{rizvi2022bias, krueger2023clinical}} \\ \bottomrule
\end{tabular}
}
\caption{Selected sociodemographic attributes $D$ and diagnoses $Y$} 
\label{table:dy}
\end{table}
\vspace{-.6cm}

Prior works have proposed sociodemographic attributes that might bias clinical decisions. For example, [\citen{maserejian2009disparities}] showed that clinicians exhibited greater uncertainty when diagnosing coronary heart disease in women compared to men. We list in Table \ref{table:dy} most of the attributes $D$ and diagnoses $Y$ we test for implicit bias, and present the full list of the 102 $(D,Y)$ pairs in Appendix \ref{appendix:alldy}. To highlight the influence of clinician subjectivity, we concentrate on diagnoses that require clinician interpretation of patient-reported symptoms, e.g., chronic pain.

Selecting health proxies for $Z$ and $X$ relies user intuition and medical expertise to determine which variables have a direct relationship with attribute $D$ and outcome $Y$, respectively. In general, proxies $X$ could be observed by the clinician during their diagnostic decision, and proxies $Z$ are not accessible during diagnosis but might have a direct causal relationship with attribute $D$. In the UK Biobank, we select $X$ to be the biometric variables collected by the biobank at patient enrollment, which includes lab results and blood pressure readings. For $Z$, we use survey responses of self-reported pain levels, mental health, and sleep. We list all variables, including the sociodemographic confounders $W$, in Appendix \ref{appendix:wzx}. Note our data contains a mix of binary, integer, and continuous variable types.
\vspace{-.3cm}

\subsection{Evaluation metrics}
\subsubsection{Semi-synthetic data validation}\label{main:semisyn}
We test if our method can retrieve a known implicit bias effect using semi-synthetic data. We use real data from the UK Biobank for attribute $D$, confounders $W$, and health proxies $X$,$Z$. We develop a model that computes $M$ and a synthetic diagnosis $Y$ with a known implicit bias effect $\theta_0$ using linear structural equations. We test against fully continuous (Experiment 1) and both binary and continuous (Experiment 2) semi-synthetic data, the latter being more realistic in real-world medical data. Our semi-synthetic data generation method is fully described in Appendix \ref{appendix:semisynthetic}. As a baseline, we compare two variants of ordinary least squares (OLS): (a) given we know $M$, we fit an OLS model over $W, D, X,$ and $M$ to predict $Y$; (b) in the more realistic scenario where $M$ isn't known, we learn over $W, D, X,$ and $Z$. We compute the average effect estimate $\bar{\theta}$ and confidence interval based on $\pm1.96\text{ }\sigma$ where the average and standard deviation $\sigma$ is taken over $K=100$ iterations. 
 \vspace{-.2cm}
\subsubsection{Calculating the implicit bias effect in the UK Biobank}
We next run our method on the full UK Biobank data. We compute the residuals of $Z, X, Y, D$ fitted on $W$ using Lasso regression. For all models, the regularization term is chosen via semi-cross fitting \cite{crossfit, chernozhukov2024applied} over 3 splits. We fit all models using the \verb|scikit-learn| Python package. For nuisance parameters $h_*$ and $\gamma_*$ we used regularized adversarial IV estimation \cite{adviv,bennett2022inference} with linear functions and a theoretically driven penalty choice that decays faster than the root of the number of samples. 

In cases where the data may not meet the method’s assumptions, we developed a proxy selection algorithm (see Section \ref{main:proxyrm}) that identifies an optimal subset of $X,Z$ proxies for each $(D,Y)$ pair using the assumption tests from Section \ref{main:tests}. Although we recommend separate data splits for proxy selection and effect estimation, we use the same dataset as our intent is method demonstration rather than robust effect estimates. Details of the hyperparameters used for the selection algorithm are provided in Appendix \ref{appendix:proxyrm}.

For each of the 102 pairs of attribute $D$ and diagnosis $Y$, we report seven metrics: the implicit bias effect $\theta$, the 95\% confidence interval, as well as our five proposed tests from Section \ref{main:tests}: (1) the primal and (2) dual violation, (3-4) the strength of identification, and (5) the $\tilde{Z},\tilde{X}$ covariance rank test. In addition, we also run the following five analyses:
\vspace{-.2cm}
\paragraph{Weak identification confidence interval} - If the instrument identification tests from \ref{main:tests} are violated, then effect estimation can be unstable and normality-based confidence intervals inaccurate. We thus compute an alternative confidence interval\cite{chernozhukov2024applied} developed under the assumption of weak instruments (see Appendix \ref{appendix:weakiv} for the description). 
\vspace{-.2cm}
\paragraph{Bootstrapping analyses}\label{main:bs} - We perform several bootstrapping analyses to test the sensitivity of the estimate. In the first analysis, given the computational complexity of recomputing the full estimate, we compare K=10 bootstrapped iterations re-estimating the full pipeline (stage 1); K=100 iterations using the pre-computed residuals but re-estimating all other parameters (stage 2); and K=1000 iterations re-computing only the final Equation \eqref{eq:finalnew} (stage 3). Each iteration samples 50\% of the data without replacement. In the second analysis, we compare sampling $10\%, 25\%, 50\%$ or $75\%$ of the original data for K=10 bootstrapped iterations, re-estimating over full pipeline (stage 1). Finally, we compare different sample sizes for $K=1000$ iterations re-estimating from stage 3 of the pipeline. 
\vspace{-.2cm}
 \paragraph{Influence points} - Inspired by [\citen{cause:finite}], we analyze influence scores, which measure how influential each data point is in the effect estimate. A significant change to the estimate after removing a small set of highly-influential points indicates the implicit bias calculation is highly sensitive to a few (potentially) outlier patients. We also include a preliminary interpretability analysis that explores the distinguishing phenotypes of highly influential patients, which could aid in determining if these subsets of patients correspond to some interpretable outlier group. We describe how we calculate the influence score and identify highly-influential patient sets in Appendix \ref{appendix:inf}. 
 \vspace{-.2cm}
\paragraph{Income stratification} -
To investigate intersectionality in implicit biases, we perform a stratified effect estimate over different income groups where $D\neq$Income.
 \vspace{-.2cm}
\paragraph{Partial non-linearity of $W$} -
Our identification theorem allows for partial non-linearity in the effect of $W$. We thus re-compute the point estimate allowing for non-linear interactions with $W$ using XGBoost\cite{xgb} models instead of Lasso.
\vspace{-.2cm}

\section{Results}

\subsection{Synthetic data validation}
The results in Table \ref{table:synth} demonstrate that our method is able to retrieve the true implicit bias effect $\theta_0=0.5$ with high certainty for both fully continuous and mixed-type data, with comparable performance to the best-case OLS where $M$ is known. We report our method's coverage, RMSE, bias, standard deviation, mean confidence interval, and performance on our five tests (from Section \ref{main:tests}), as well as testing other values of $\theta$, in Appendix \ref{appendix:semisynthetic_results}. 
\begin{table}[ht]
\centering
\resizebox{.7\textwidth}{!}{%
\begin{tabular}{@{}lcccc@{}}\toprule
{} & \textbf{$\theta_0$} & \textbf{Our method} & \textbf{OLS($D, W, M, X$)} & \textbf{OLS($D, W, Z, X$)}\\ \colrule
{Experiment 1: Continuous} & 0.5 & 0.54 $\pm$ 0.003  &  0.5 $\pm$ 0.01  & 1.10 $\pm$ 0.01 \\
{Experiment 2: Continuous and binary} & 0.5 & 0.53 $\pm$ 0.003  & 0.5 $\pm$ 0.01  & 1.385 $\pm$ 0.01   \\ \bottomrule
\end{tabular}
}
\caption{Semi-synthetic data estimates $\bar{\theta}\pm1.96\sigma$ over $K=100$.}
\label{table:synth}
\end{table}
\vspace{-.5cm}
\subsection{Calculating the implicit bias effect in the UK Biobank}
In Appendix \ref{appendix:7metrics_allXZ}, we show the effect estimates for the $(D, Y)$ pairs using all proxies $Z$, $X$, adjusting the confounders $W$ by excluding the column corresponding to the attribute $D$. However, as evidenced by the failure of the dual and primal tests, we found the initial sets of proxies $Z$, $X$ did not meet our method's necessary assumptions. As discussed further in Appendix \ref{appendix:invalidproxies}, we believe these test failures indicate there might exist some features in $X$ with a causal path from $D$ that does not go through $M$ or features within $Z$ with a causal path to $Y$ that doesn't flow through $M$. Such paths invalidate the resulting effect estimates. 

We thus found applying our proxy selection algorithm (see \ref{main:proxyrm}) necessary for producing valid effect estimates. After running the algorithm to select subsets of admissible $X,Z$ proxies (the description and interpretation of the selected proxies can be found in Appendix \ref{appendix:invalidproxies}), we found 34 $(D,Y)$ pairs that pass all tests with narrow confidence intervals. We report six in Table \ref{table:proxyrm} and include the remaining estimates in Appendix \ref{appendix:7metrics_proxyrm}. Note that $\theta > 0$ implies a patient with $D$ is more likely to be diagnosed with $Y$ due to clinician bias, and conversely $\theta < 0$ implies a patient is less likely to be diagnosed. In Section \ref{discussion}, we offer a framework for interpreting the implications of these results. 
\vspace{-.2cm}

\subsubsection{Weak instrument confidence interval} 
\begin{figure}[ht]
\centerline{\includegraphics[width=\textwidth]{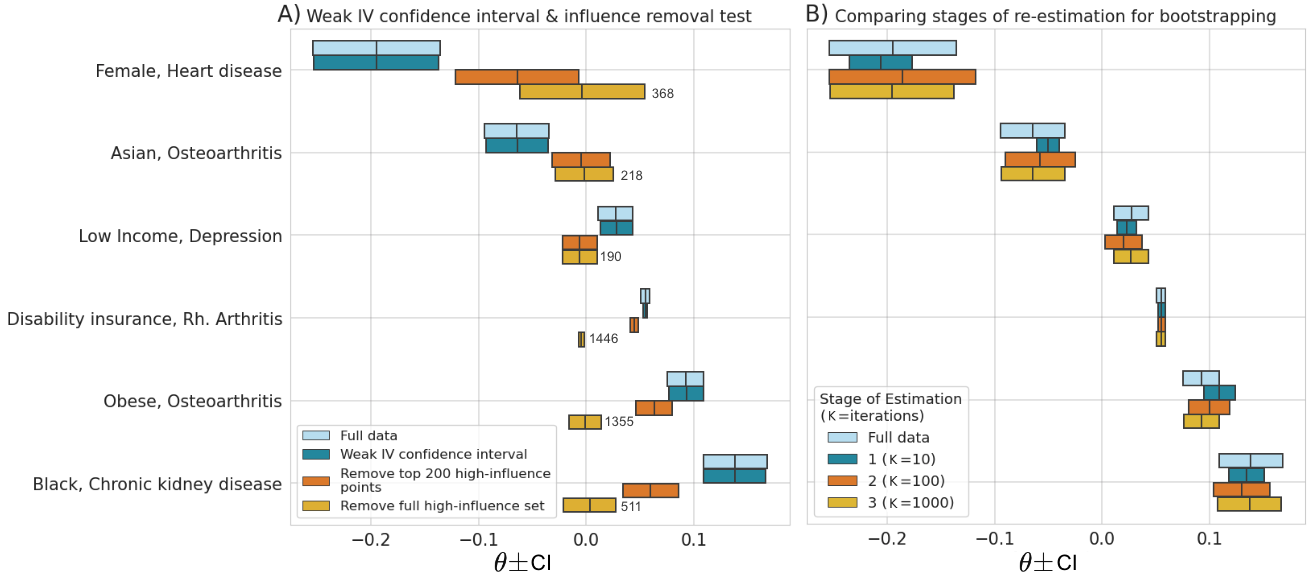}}
\caption{Comparing effect estimates for six $(D,Y)$ pairs using all data with: A) weak instrument and influence set removal (where the numbers next to the yellow bar reflect the set size of high-influence points); B) bootstrapped subsampling 50\% of the data at different stages of re-estimation.}
\label{fig:ci}
\end{figure}

As shown in Figure \ref{fig:ci}A, the confidence interval predicted under the weak instrument regime consistently aligns with the interval under our method, thus indicating our estimate's robustness to weak instruments. 
\vspace{-.1cm}

\begin{table}[ht]
\centering
\resizebox{.99\textwidth}{!}{%
\begin{tabular}{@{}ccccccc@{}}
% \cmidrule(r){1-7}
\toprule
\multicolumn{1}{c}{\textbf{$(D,Y)$}} & \multicolumn{1}{c}{\textbf{$\theta\pm95$\% CI}} & \multicolumn{1}{c}{\textbf{\begin{tabular}[c]{@{}c@{}}(1) Primal \\ statistic $<$ critical  \end{tabular}}} & \multicolumn{1}{c}{\textbf{\begin{tabular}[c]{@{}c@{}}(2) Dual \\ statistic $<$ critical  \end{tabular}}} & \multicolumn{1}{c}{\textbf{\begin{tabular}[c]{@{}c@{}}(3) $\E[\tilde{D}V] \neq 0$\\ statistic $>$ critical  \end{tabular}}} & \multicolumn{1}{c}{\textbf{\begin{tabular}[c]{@{}c@{}}(4) $V$ strength F-test \\ statistic $>$ critical  \end{tabular}}} & \multicolumn{1}{c}{\textbf{(5) Cov($\tilde{X}, \tilde{Z}$) rank}} \\ \midrule
Low income, Depression & 0.03 $\pm$ 0.02 & 59.9$<$60.5 & 31.9$<$40.1 & 84.1$>$0.4 & 3332.1$>$23.1 & 3 \\
Disability insurance, Rh. Arthritis & 0.06 $\pm$ 0.0 & 67.3$<$75.6 & 3.4$<$11.1 & 29.2$>$0.4 & 801.1$>$23.1 & 3 \\
Female, Heart disease & -0.19 $\pm$ 0.06 & 115.8$<$118.8 & 23.3$<$23.7 & 18.8$>$1.3 & 92.5$>$23.1 & 4 \\
Black, Chronic kidney disease & 0.14 $\pm$ 0.03 & 56.9$<$58.1 & 10.6$<$21.0 & 9.8$>$0.3 & 23.3$>$23.1 & 4 \\
Obese, Osteoarthritis & 0.09 $\pm$ 0.02 & 90.5$<$100.7 & 24.8$<$28.9 & 76.5$>$1.7 & 254.9$>$23.1 & 3 \\
Asian, Osteoarthritis & -0.06 $\pm$ 0.03 & 94.7$<$101.9 & 33.1$<$33.9 & 13.9$>$0.3 & 74.6$>$23.1 & 5 \\ \bottomrule
% \cmidrule(r){1-}
\end{tabular}
}
\caption{Six of the 34 valid UK Biobank implicit bias effect estimates after applying our $X,Z$ proxy selection algorithm. Tests (1-5) are detailed in \ref{main:tests}, where \textit{statistic} is the given data's statistic and \textit{critical} is the necessary critical value to be greater or less than to pass. $V = \tilde{D}-\gamma^\top \tilde{Z}$.}
\label{table:proxyrm}
\end{table}
\vspace{-.4cm}
\vspace{-.4cm}

\subsubsection{Bootstrapping analyses}
In Figure \ref{fig:ci}B, we show the results of the first bootstrap analysis comparing different stages of re-estimation. We observe that, regardless of the estimation stage, bootstrapped estimates are consistent with the estimate from the full dataset. The consistency of the bootstrapped estimates over different sample sizes, as shown in Appendix \ref{appendix:bs_samplesize}, further support the robustness of our method. 
\vspace{-.2cm}

\subsubsection{Influence points} 
\begin{figure}[ht]
\centerline{\includegraphics[width=.9\textwidth]{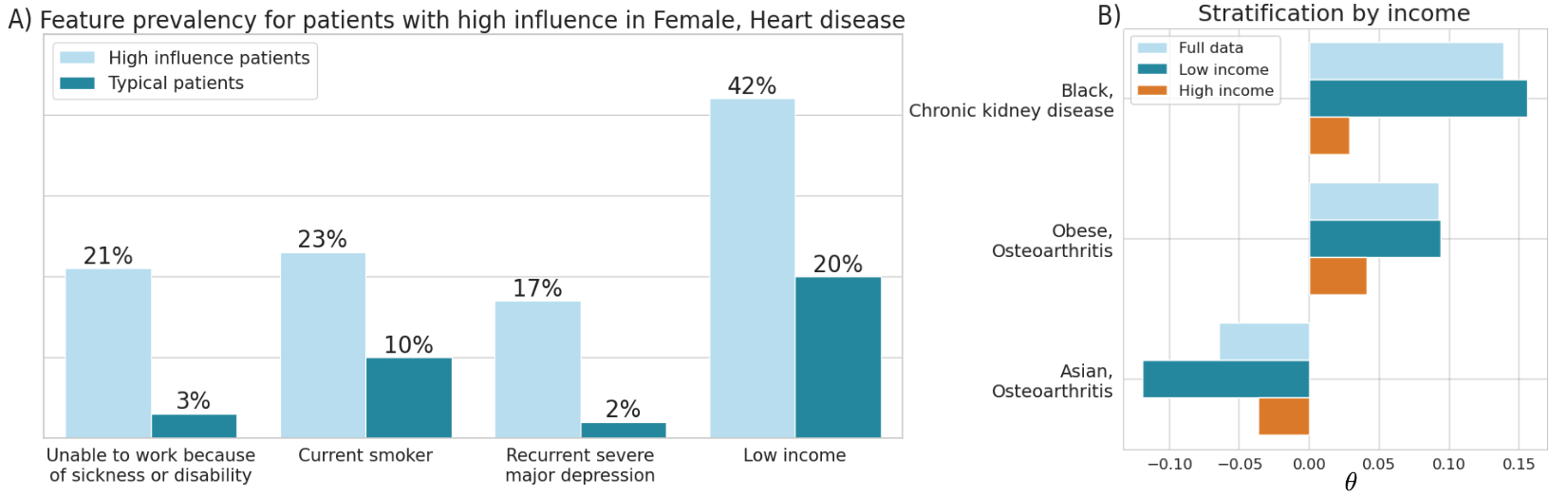}}
\caption{A) Interpretability into high influence points. B) Income stratification}
\label{fig:inf}
\end{figure}

In Figure \ref{fig:ci}A, we see that removing only a few highly-influential points leads to a significant decrease in the magnitude of the estimated effect. To investigate, we run a preliminary interpretabilty analysis where we analyze the univariate differences between patients with high influence and those with low influence. In Figure \ref{fig:inf}A patients that strongly influence the negative implicit bias estimate for ($D$=Female, $Y$=Heart disease) are more likely to be low income, unemployed due to disability, and suffer from depression. It is plausible such patients are the ``outliers'' driving the strong negative bias estimate.

\vspace{-.5cm}

\subsubsection{Income stratification}
In Figure \ref{fig:inf}B we analyze the effect of stratification based on income. We see a general increase in bias effect estimate for the low income strata and a corresponding decrease in effect for high income strata, demonstrating potential evidence of intersectional discrimination\cite{ogungbe2019systematic, anastas2020unique}.
\vspace{-.3cm}

\subsubsection{Partial non-linearity of $W$} 
In Appendix \ref{appendix:nonlinear}, we show our implicit bias estimate with non-linear $W$ interactions leads to a similar effect estimates of $\theta$.
\vspace{-.3cm}

\section{Discussion}
\subsection{Limitations}\label{limitations}
In this work, we propose a robust causal inference method designed to detect clinician implicit bias by estimating pathway-specific causal effects. We demonstrate the applicability of our approach to large-scale medical data by validating on both semi-synthetic and real-world datasets. 

However, our work contains several limitations. First, while the UK Biobank is a rich and accessible source of medical data, most patient information is collected once upon signing up for the biobank. Although UK Biobank has synced their records to a handful of outcomes provided by EHR data (like ICD10 codes), it is unclear to what extent the available proxies for $X$ (which were collected at patient enrollment) are used by clinicians for diagnoses. Additionally, the synced ICD10 codes are from hospital records, thus excluding primary care visits. We plan to validate our method with time-series EHR data in follow-up work. 

Second, while the assumption of partially linear structural equations is crucial for enabling better identifiability of the outcome bridge function under minimal conditions, it is possible the ground truth equations are non-linear.

Finally, it is well known that intersectional identities shape complex patterns of discrimination in healthcare \cite{ogungbe2019systematic, anastas2020unique}. A more comprehensive analysis on the effect of implicit bias from intersectional attributes on patient treatment would be valuable for improving equity in healthcare outcomes.

\vspace{-.3cm}
\subsection{Interpretation and application of results}\label{discussion}
While we re-iterate the intent of this work is not to diagnose specific cases of implicit bias in the UK Biobank, our method did flag several areas of clinical inequity that have been reported in literature. For instance, many works have reported gender-based inequality in cardiovascular health \cite{bosomworth2023analysis}, and we similarly detected an estimate of $\theta$ = -0.19 indicating clinicians are less likely, due to implicit biases, to diagnose $D$=Females with $Y$=heart disease. In another example, our estimate $\theta$ = -0.06 suggested clinicians are less likely to diagnose $D$=Asian patients with $Y$=osteoarthritis, and many works have highlighted both patient- and clinician-stigmas regarding pain-associated disorders, like osteoarthritis, in Asians \cite{yang2023qualitative, gee2007nationwide, kumar2019health}. 

However, we did find several estimates contrary to what we expected. For example, our estimate $\theta=0.14$ indicated clinicians are  \textit{positively} biased towards diagnosing Black patients with chronic kidney disease. However, at the time of UK Biobank data collection, many doctors relied on a race-based equation for kidney function now known to have under-detected kidney disease in Black patients \cite{marzinke2022limited}. 

To understand a discrepancy between a produced estimate and literature (or user intuition), we recommend (1) ensuring the data used contains sufficient health proxies and satisfy all assumptions (e.g., see biobank data limitations in \ref{limitations}); (2) investigating all mechanisms creating the medical outcome $Y$ (e.g., hospital-specific diagnosis protocol); and (3) exploring how the discovered bias estimate fits in context, rather than opposed, to those found in literature. While our method does not offer a solution on \textit{how} to tackle implicit biases, by bringing awareness to potential areas of discrimination within a given healthcare system, detecting biases is the first step towards creating systemic-level change through interdisciplinary collaboration and targeted anti-bias training programs.
\vspace{-.3cm}
\section{Appendix}
The appendix can be found at \url{https://github.com/syrgkanislab/hidden_mediators}.

\vfill
\newpage
% Appendix: \url{https://www.overleaf.com/read/bmfjkswtgqrn#d4bdf7}

\bibliographystyle{ws-procs11x85}
\bibliography{ws-pro-sample}
\vfill
\newpage

\appendix{Semi-synthetic data generation method}\label{appendix:semisynthetic}

We want to generate semi-synthetic data $(W_{\gen}, M_\gen, D_{\gen}, Z_{\gen}, X_{\gen}, Y_{\gen})$ that incorporates real data, generates all variables using partially linear structural equations, and allows for a user-specified implicit bias effect $\theta$. 

For confounders $W_{\gen}$, we use a sample from the empirical distribution of the real confounders. In particular, $W_{\gen}$ is generated by sampling with replacement uniformly from the real confounders $W$. Subsequently, we want to learn from the real data a partially linear low-rank latent factor model where we are treating the mediator $M$ as the latent factor. In particular, our goal is to learn a generative model that is of the form:
\begin{equation}
\begin{aligned}
    D_{\gen} =~& \hat{f}_D(W_\gen, \hat{\epsilon}_D)\\
    M_\gen =~& a D_{\gen} + \hat{\epsilon}_M\\
    Z_{\gen} =~& \hat{G} M_{\gen} + \hat{f}_Z(W_{\gen}) + \hat{\epsilon}_Z\\
    X_{\gen} =~& \hat{F} M_{\gen} + \hat{f}_X(W_\gen) + \hat{\epsilon}_X
\end{aligned}
\label{eqn:synth-data}
\end{equation}
where $\hat{f}_Z, \hat{f}_X, \hat{G}, \hat{F}$ and the distributions of $\hat{\epsilon}_Z, \hat{\epsilon}_X, \hat{\epsilon}_M$, relate to real-world quantities.

Without loss of generality, the binary sensitive attribute $D$ can be thought of as generated according to a Bernoulli distribution with success probability $p_D(W)=\E[D\mid W]$. Thus, to generate a sample $D_{\gen}$, we learn a binary probabilistic classifier (or ``propensity model") $\hat{p}_D(W)$ by running a logistic regression of $D$ on $W$ using the real data. To generate semi-synthetic $D_{\gen}$ conditioned on the already sampled $W_{\gen}$, we sample based on a Bernoulli distribution with success probability $\hat{p}_D(W_{\gen})$. More formally, 
\begin{align}
D_{\gen} =~& \hat{f}_D(W_\gen, \hat{\epsilon}_D)\sim \text{Bernoulli}(\hat{p}_D(W_\gen))
\end{align}

The functions $\hat{f}_Z$ and $\hat{f}_X$ are LASSO regressions (with L1 penalty selected from cross-validation) trained to predict $Z$ from $W$ and $X$ from $W$, respectively.

For the parameters $\hat{F}, \hat{G}$ as well as the distribution of $M_{\gen}$, we proceed as follows. Consider the real data residuals $\tilde{Z}$ and $\tilde{X}$ after removing their predictable parts from $W$, based on the already trained LASSO regressions (i.e., $\tilde{Z} = Z - \hat{f}_Z(W))$. If the real world data were truly generated by the partially-linear structural equation model as in Equations~\eqref{eqn:synth-data}, then the real data residuals $\tilde{Z}$ and $\tilde{X}$ would obey the generative equations
\begin{align}
  \tilde{Z} =~& G \tilde{M} + \epsilon_Z &
  \tilde{X} =~& F \tilde{M} + \epsilon_X
\end{align}
where $\tilde{M}=M-\E[M\mid W]$. Note that under this structural model:
\begin{align}
  \cov(\tilde{Z}, \tilde{X}) = \E[\tilde{Z} \tilde{X}^\top ] = G \E[\tilde{M}\tilde{M}^\top ] F^\top 
\end{align}
Moreover, if we assume the latent factors of the mediator $M$ are roughly independent, then Var($\tilde{M}) = \E[\tilde{M}\tilde{M}^\top] = \text{diag}(\sigma_1, \ldots, \sigma_K)$, and a singular value decomposition (SVD) of the covariance $\cov(\tilde{Z}, \tilde{X})=U \Sigma V^\top $ would reveal $G$ as the left singular vectors $U$, $F$ as the right singular vectors $V$, and $(\sigma_1, \ldots, \sigma_K)$ as the singular values in $\Sigma$. Thus, to calculate the $\hat{G}$ and $\hat{F}$ that we will use in our model using Equations~\eqref{eqn:synth-data}, we run SVD over the empirical covariance matrix $\E_n[\tilde{Z} \tilde{X}^\top ]=U \Sigma V^\top $. Given $\Sigma = \text{diag}(\hat{\sigma}_1, \ldots, \hat{\sigma}_n)$, we discard all singular values that fall below a threshold calculated based on a measure of statistical significance (see appendix \ref{apendix:test:rank}) and consider only the remaining top $K$ statistically significant left and right eigenvectors $\hat{G}=U_{\cdot, 1:K}$, $\hat{F}=V_{\cdot, 1:K}$ and their corresponding eigenvalues $(\hat{\sigma}_1, \ldots, \hat{\sigma}_K)$. The generated hidden mediator $M_{\gen}$ is taken to be $K$-dimensional and the distribution of the error variable $\hat{\epsilon}_M$ is a multivariate normal distribution with covariance matrix $\text{diag}(\hat{\sigma}_n,\ldots, \hat{\sigma}_K)$.

The noise variables $\hat{\epsilon}_Z$ and $\hat{\epsilon}_X$ are drawn from the empirical marginal distribution of the corresponding real world variables. Specifically, for a given $i$th random semi-synthetic sample $Z^{(i)}_{\gen}$, we sample its noise $\hat{\epsilon}^{(i)}_Z$ by sampling uniformly at random a real data point $Z^{(j)}$, i.e., $\hat{\epsilon}^{(i)}_Z =Z^{(j)}$. Similarly, for the same $i$th random semi-synthetic sample, the noise $\hat{\epsilon}^{(i)}_X$ is generated by sampling uniformly at random another real data point $X^{(k)}$, i.e., $\hat{\epsilon}^{(i)}_X =X^{(k)}$.
Finally, our algorithm generates the synthetic outcome based on the linear model:

\begin{align}
    Y_{\gen} =& \frac{b}{K} \sum_{i=1}^K M_{i,\gen} + \theta D_{\gen} + g X_{0, \gen} + \hat{f}_Y(W_{\gen}) + \sigma_Y \hat{\epsilon}_Y
\end{align} 
where $X_{0, \gen}$ denotes the first variable in $X_{\gen}$ and $M_{i,\gen}$ the $i$th latent variable of $M_{\gen}$. Here $\hat{f}_Y$ is a LASSO regression (with cross-validated L1 penalty) trained to predict $Y$ from $W$. Note that we could generate $Y_{\gen}$ as any linear function of $M,D_{\gen}, X_{\gen}$; we just chose this equation for our experiments. The noise variable $\hat{\epsilon}_Y$ is drawn from the empirical distribution of $Y$ in the real data. Specifically, for a given $i$th random semi-synthetic sample $\tilde{Y}^{(i)}_{\gen}$, we generate a sample from $\hat{\epsilon}_Y$ by sampling uniformly at random a real data point $Y^{(m)}$, i.e., $\hat{\epsilon}^{(i)}_Y =Y^{(m)}$.

Parameters $a,b,\theta, g, \sigma_Y$ are chosen by the user as an input. $a$ controls the degree of influence of the treatment $D$ on the hidden mediator $M$. $b$ controls the degree of direct influence of the hidden mediator $M$ on the outcome $Y$. $\theta$ is the true controlled direct effect of the sensitive attribute $D$ on $Y$. $g$ controls the direct effect of the outcome proxies $X$ on the outcome $Y$. $\sigma_Y$ controls the level of noise of the outcome variable.

To avoid overfitting when training the parameters in our semi-synthetic generation process, we split our real data into a train-test (or train-sampling) split, and estimate $\hat{F}$, $\hat{G}$, $\text{diagonal}(\hat{\sigma}_1, ..., \hat{\sigma}_K)$, $\hat{p}_D$, $\hat{f}_X$, $\hat{f}_Z$
and $\hat{f}_Y$ using the training data. We use the test data to draw random samples from the empirical distributions that were used for $W_\gen, \hat{\epsilon}_X, \hat{\epsilon}_Z, \hat{\epsilon}_Y$.

\appendix{Proxy variable removal algorithm}\label{appendix:proxyrm}
For simplicity of notation, we omit the tilde's on the variables and use the short-hand notation in this section as: $Y\equiv \tilde{Y}$, $X\equiv \tilde{X}$, $Z\equiv \tilde{Z}$, $D\equiv \tilde{D}$, where recall $\tilde{V} = V - \E[V | W]$.

Furthermore, let $\cX\subseteq [p_X]$ and $\cZ\subseteq [p_Z]$ be subsets of indices of proxies\footnote{We use the notation $[n]=\{1,\ldots,n\}$ for any integer $n\geq 0$.}, where $p_X, p_Z$ is the dimension of $X, Z$ correspondingly. Let the variables corresponding to the indices be $X_{\cX}=(X_j)_{j\in \cX}$ and $Z_{\cZ} = (Z_j)_{j\in \cZ}$.

\subappendix{Overview}
We develop an algorithm that selects a subset of proxy features (or variables; we use these terms interchangeably) from $X,Z$ such that the primal and dual violation tests pass. Intuitively, this algorithm identifies pairs of indices $\cX, \cZ$ such that (1) there exists no direct paths from ${D}$ to ${X}_{\cX}$ that are not through $M$, and (2) there are no paths from ${Z}_{\cZ}$ to ${Y}$ that are not through $M$. In other words, there can exist no health proxy features in ${X}_{\cX}$ that have an alternate mediator $M' \neq M$ that is not shared by ${Z}_{\cZ}$, and vice versa. Thus, we only want to keep the proxy features $\cX$ and $\cZ$ that share the mediator $M$.

To estimate violations of $\cZ$ and $\cX$, we use the primal and dual equation violation tests, respectively (see Section \ref{main:tests} and Appendix \ref{appendix:tests} for more details on these tests, and why we believe they measure these violating paths). Let $p(\cX, \cZ), d(\cX, \cZ)$ be the primal and dual test statistics when we use the subsets of proxies ${X}_{\cX}, {Z}_{\cZ}$. However, computing the primal and dual test statistics is expensive as it requires estimating parameters $\theta, h, \gamma$. Thus we also use efficient approximations $\hat{p}(\cX,\cZ), \hat{d}(\cX,\cZ)$. We discuss how we approximate $\hat{p},\hat{d}$ in Appendix \ref{proxyscore_est}. 

We first compute the primal and dual approximate test scores $\hat{p}_* = \hat{p}([p_X],[p_Z]), \hat{d}_* = \hat{d}([p_X],[p_Z])$ using all $X,Z$ proxy features. These are the baseline scores we want to beat by a substantial amount. Our proxy feature selection algorithm then starts by considering all proxy variables of $Z$, i.e. $\cZ=[p_Z]$ and no proxies from $X$, i.e. $\cX=\emptyset$. We increment $\cX$ until we can no longer add more elements without incurring a large test score. At each iteration, we choose a random proxy variable $i \in [p_X]/\cX$ and add it to $\cX$ as long as $\hat{d}(\cX \cup \{i\}, \cZ) < \delta \cdot \hat{d}_*$ for some hyperparamter $\delta \in (0, 1)$; otherwise we discard the element $i$. Finally, when we have the largest candidate set $\cX$ that still satisfies the aforementioned inequality, we check using the full dual test score if $d(\cX, \cZ) < \text{critical-value}$, where critical-value is the rejection critical value calculated for the dual test score in Appendix~\ref{appendix:tests}, and keep $\cX$ if it passes. We repeat this $K$ times over random seeds to get at most $K$ potential candidate $X$ proxy variable sets: $\{\cX^{(1)}, \cX^{(2)}, ...\cX^{(k)}\}$. 

Next, we repeat the method to get candidate proxy variable sets for $Z$. For each candidate set for $X$, $\cX^{(i)}\in \{\cX^{(1)}, \cX^{(2)}, ...\cX^{(K)}\}$, we start with no $Z$ proxy variables: $\cZ =\emptyset$. Then, we randomly add some feature $j\in [p_Z]$ into $\cZ$ and continue adding $Z$ variables as long as $\hat{p}(\cX^{(i)}, \cZ\cup \{j\}) < \delta \cdot \hat{p}_*$. Again, once we have the maximum size set $\cZ$ that satisfies the inequality of the primal test estimate, we use the non-estimate primal score to check if $p(\cX^{(i)}, \cZ) < \text{critical-value}$, where critical value is the rejection critical value calculated for the primal test in Appendix~\ref{appendix:tests}, and only keep $\cZ$ if it passes. For each of the $\cX^{(i)}$ proxy variable sets $\{\cX^{(1)}, \cX^{(2)}, ...\cX^{(K)}\}$, we find $K$ potential sets $\cZ^{(j)}$ for a maximum of $K^2$ total candidate proxy variable set pairs. We repeat this one more iteration, where the only change is that instead of initializing $Z$ to just $[p_Z]$ as we did in the first iteration when constructing the maximal $X$, we loop over all candidate $\cZ^{(j)}$ sets.

The result is a list of potential candidate pairs $\{(\cX^{(i)}, \cZ^{(i)})\}_i$ for a specific $(D,Y)$ estimate that minimally-violate the dual and the primal violation tests. Ideally, this list of candidates would be performed on a separate dataset from the one used to re-run the bias-detection method on each of the pairs.

\subappendix{Dual and primal violation score estimates}\label{proxyscore_est}

Both the primal and the dual tests (see Appendix \ref{appendix:test-primal} and \ref{appendix:test-dual}, respectively) are checking whether a solution $\eta$ to a linear system of equations:
\begin{align*}
    \Sigma \eta = v
\end{align*}
exists, where $\Sigma$ is an $n\times m$ matrix and $v$ is an $m$-dimensional vector. If a solution exists, then one such solution is the minimum norm solution $\Sigma^+ v$ and it would then satisfy:
\begin{align*}
    \Sigma \Sigma^+ v = v \Leftrightarrow (\Sigma \Sigma^+ - I) v = 0
\end{align*}
Thus, as an indication for the existence of a solution, we can use the $\ell_{\infty}$ norm of this vector\footnote{Recall $\norm{x}_{\infty} = \text{max}_i{\|x_i\|}$.}:
\begin{align*}
    s = \|(\Sigma \Sigma^+ - I) v\|_{\infty}
\end{align*}
A linear system that admits a solution will have a zero $\ell_{\infty}$ norm $s$. However, given access to $n$ samples, we can only approximate $\Sigma$ as $\Sigma_n$ and $v$ as $v_n$. Consider the error $\Sigma_n-\Sigma$, which is roughly  $\E[\epsilon_X \epsilon_Z^\top]$ for both the dual and the primal. Because we do not expect the noise vectors $\epsilon_X$ and $\epsilon_Z$ to have a low rank structure, the error $\Sigma_n-\Sigma$ might be full rank, and thus $\Sigma_n$ might not represent well the subspace spanned by the range of $\Sigma$. For this we introduce a low-rank approximation to remove the noise. For any matrix $A$, let $\text{LRA}(A)$ denote its low-rank approximation where we only keep the singular values and corresponding eigenvectors of $A$ are greater than a critical threshold $\tau$ (similarly to the process of calculating the threshold in the covariance rank test in Appendix~\ref{apendix:test:rank}). In particular, if $A=UDV^\top$ is the SVD of $A$ 
with $D=\text{diag}(\lambda_1,\ldots, \lambda_K)$, then we let:
\begin{align}
    \text{LRA}(A) =~& U \hat{D} V^\top &
    \hat{D} =~& \text{diag}(\lambda_1 \cdot 1\{|\lambda_1| > \tau\}, \ldots, \lambda_K \cdot  1\{|\lambda_K| > \tau\})
\end{align}
Moreover, to remove parts of $v_n$ that span coordinates not spanned by $v$, we also introduce a sparsity regularization where $\text{sparse}(v)=(v_1 \cdot  1\{|v_1| > \kappa_1\}, \ldots, v_K \cdot 1\{|v_K|>\kappa_K\})$ and the threshold $\kappa_i$ is chosen to be a small multiple of the standard error of $\hat{v}_i$.

Let $\hat{\Sigma}=\text{LRA}(\Sigma_n)$ and $\hat{v}=\text{sparse}(v_n)$. Then we use as an approximate score quantity:
\begin{align}
    \hat{s} = \|(\hat{\Sigma}\hat{\Sigma}^+ - I)\, \hat{v}\|_{\infty}
\end{align}
Moreover, for any pair of subsets of columns $\cS \subseteq [n]$ and rows $\cT\subseteq [m]$ (e.g., $\cS = \cX$), we use the approximate score function:
\begin{align}
    \hat{s}(\cS, \cT) = \|(\hat{\Sigma}_{\cS, \cT}\hat{\Sigma}_{\cS, \cT}^+ - I)\, \hat{v}_{\cS}\|_{\infty}
\end{align}
where $\hat{\Sigma}_{\cS,\cT}$ denotes the sub-matrix of $\hat{\Sigma}$ that contains only the rows in $\cS$ and the columns in $\cT$, and $\hat{v}_{\cS}$ denotes the sub-vector of $\hat{v}$ that contains only the entries in $\cT$.

For the case of the primal equation, we can create a proxy score $\hat{p}(\cX, \cZ)$ by invoking the above approach with:
\begin{align*}
    \Sigma_n =~& \E_n\left[\begin{pmatrix}{Z}\\ {D}\end{pmatrix} \begin{pmatrix}{X}\\ {D}\end{pmatrix}^\top\right] &
    v_n =~& \E_n\left[\begin{pmatrix}{Z}\\ {D}\end{pmatrix} Y\right]
\end{align*}
and for the dual we can create a proxy score $\hat{d}(\cX, \cZ)$ where:
\begin{align*}
    \Sigma_n =~& \E_n[XZ^\top] &
    v_n =~& \E_n[X D]
\end{align*}

\subappendix{Pseudocode}
\newpage
\vfill
\begin{algorithm}[H]
\caption{Proxy Variable Selection Algorithm}
\begin{algorithmic}[1]
\State \textbf{Input:} Proxy data $Z$ and $X$, hyperparameter $\delta$, number of iterations $K$
\State \textbf{Output:} Pairs of candidate variable sets for proxies in $X$ and $Z$
\State Calculate baseline approximate primal and dual scores $\hat{p}_* = \hat{p}([p_X],[p_Z]), \hat{d}_* = \hat{d}([p_X],[p_Z])$
\State Initialize CandidateSets\_Z = $\{[p_Z]\}$ \Comment{List to store candidate sets of $Z$}
\State
\State Set CandidateSets\_X = $\emptyset$ \Comment{List to store candidate sets of $X$}
\For{each unique candidate $\cZ^{(i)} \in \text{CandidateSets\_Z}$} 
    \For{$k = 1$ to $K$} \Comment{Find candidate proxy sets of $X$}
        \State $\cX = \emptyset$  \Comment{Start with empty set}
        \State Randomly select new proxy variable $j \in [p_X]$
        \While{$\hat{d}(\cX \cup \{j\}, \cZ^{(i)}) < \delta \cdot \hat{d}_*$}
            \State Set $\cX \leftarrow \cX \cup \{j\}$
            \State Randomly select new proxy variable $j \in [p_X] / \cX$
        \EndWhile
        \If{$d(\cX, \cZ^{(i)}) < $ critical-value} \Comment{Compare true statistic to $\chi^2$ critical value}
            \State Add $\cX$ to CandidateSets\_X
        \EndIf
    \EndFor
\EndFor
\State Set CandidateSetsPairs = $\emptyset$ \Comment{List to store candidate variable pairs}
\State Set CandidateSets\_Z = $\emptyset$
\For{each candidate $\cX^{(i)} \in \text{CandidateSets\_X}$} 
\For{$k = 1$ to $K$} \Comment{Find candidate proxy sets of $Z$}
        \State $\cZ = \emptyset$  \Comment{Start with empty set}
        \State Randomly select feature $j\in [p_Z]$
        \While{$\hat{p}(\cX^{(i)}, \cZ \cup \{j\}) < \delta \cdot \hat{p}_*$}
            \State Set $\cZ \leftarrow \cZ \cup \{j\}$
            \State Randomly select new feature $j \in [p_Z]/\cZ$
        \EndWhile
        \If{$p(\cX^{(i)}, \cZ) < $ critical value} \Comment{Compare true statistic to $\chi^2$ critical value}
    \State Add $(\cX^{(i)}, \cZ)$ to CandidateSetPairs 
    \State Add $\cZ$ to CandidateSets\_Z 
    \EndIf
        
    \EndFor
\EndFor
\State
\State Optionally repeat Lines 6 - 35.

\State \textbf{Output:} Candidate proxy variable pairs $\{(\cX^{(i)}, \cZ^{(i)})\}_i$
\end{algorithmic}
\end{algorithm}
\appendix{Tests}\label{appendix:tests}
For simplicity of notation, we omit the tilde's on the variables and use the short-hand notation in this section as: $Y\equiv \tilde{Y}$, $X\equiv \tilde{X}$, $Z\equiv \tilde{Z}$, $D\equiv \tilde{D}$, where recall $\tilde{V} = V - \E[V | W]$.

\subappendix{Primal violation test}\label{appendix:test-primal}

\begin{figure}[H]
\centering

\begin{tikzpicture}[>={Stealth[round]},shorten >=1pt,node distance=2cm,on grid,initial/.style={}]
  % Nodes
  \node[font=\Large] (D) {$D$};
  \node[below right=of D, font=\Large] (Z) {$Z$};
  \node[above right=of Z, font=\Large] (M) {$M$};
  \node[below right=of M, font=\Large] (X) {$X$};
  \node[above right=of X, font=\Large] (Y) {$Y$};

  % Edges
  \draw[->] (D) -- (M) node[midway, above] {};
  \draw[->] (M) -- (Y) node[midway, above] {};
  \draw[->] (D) edge[bend left] node[pos=.5, above, sloped] {} (Y)  ;
  \draw[->, dashed] (D) -- (Z) node[midway, above] {};
  \draw[->] (M) -- (Z) node[midway, above] {};
  \draw[->] (M) -- (X) node[midway, above] {};
  \draw[->, dashed] (X) -- (Y) node[midway, above] {};
  \draw[<->, dashed] (X) edge[bend left=65] node[pos=.9,xshift=-4mm, yshift=-5mm, below, font=\small] {Dual violation} (D);
  \draw[<->, dashed] (Z) edge[bend right=105] node[pos=.7,,yshift=-5mm, below, font=\small] {Primal violation} (Y) ;
\end{tikzpicture}
\caption{Potential causal graph causing the violations of the primal and the dual. The violation of the primal is caused by some $Z' \rightarrow M_Z \rightarrow Y$, and the violation of the dual is caused by some $X'$ such that $D \rightarrow M_X \rightarrow X'$ }
\label{cg:vio}
\end{figure}
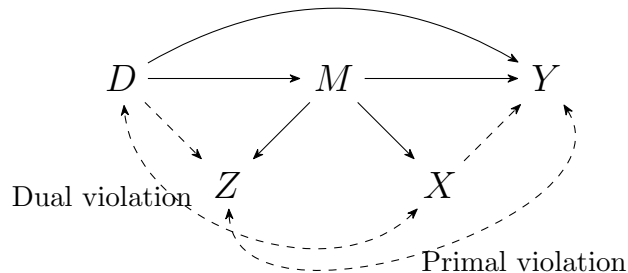

\subsubappendix{Intuition} 
Recall the \eqref{eq:primal} is \begin{align}
        \E\left[({Y} - {X}^\top h - {D}\,\theta) \begin{pmatrix}{Z}\\ {D}\end{pmatrix}\right] =~& 0 
\end{align}
We want to test whether this equation admits a solution $\theta_0, h_*$. This is a linear system of equations:
\begin{align}
    \E\left[\begin{pmatrix}{Z}\\ {D}\end{pmatrix} \begin{pmatrix}{X}\\ {D}\end{pmatrix}^\top \right] \begin{pmatrix}h\\ \theta\end{pmatrix} = \E\left[\begin{pmatrix}{Z}\\ {D}\end{pmatrix} Y\right]
\end{align}
Since all variables correspond to residuals and hence are mean-zero, the above is equivalent to:
\begin{align}
    \cov((Z;D), (X;D)) \begin{pmatrix}h\\ \theta\end{pmatrix} = \cov((Z;D), Y)
\end{align}
Thus, the primal holds if $\text{Cov}((Z;D), Y) \in \text{column-span}(\text{Cov}((Z;D), (X;D))$. This is roughly equivalent to $\text{Cov}(Z,Y) \in \text{column-span}(\text{Cov}(Z, X))$. Intuitively, this means that all the information flowing between $Y$ and $Z$ must go through the same mediator $M$ that influences $X$ and $Z$. This would not be true if either of two (equivalent) assumptions were violated: (1) there was some proxy variable $Z_i$ which has an alternate path to $Y$ not through $X$'s mediator $M$, i.e., either $\exists  Z_i$ that has a direct influence on $Y$ or $\exists  Z_i, M'$ s.t. $Z_i \rightarrow M' \rightarrow Y$ and $M' \cancel{\rightarrow} X$; (2) the proxy variables $X$ have no direct relationship with the mediator $M$. The former potential violation is visualized in Figure \ref{cg:vio}.

If we view the primal violation as an issue of the first scenario (i.e., some $Z_i$ has an alternate mediator $M'$), then we can mitigate the violation by identifying and removing such a $Z_i$, which our proxy variable removal algorithm attempts to do (see Appendix \ref{appendix:proxyrm}). If we view the violation from the second scenario (i.e., $Z$'s mediator $M$ has no direct relationship to all proxy variables $X$), we could try adding stronger, more informative proxy variables to $X$ that are likely to be mediated by $M$.

\subsubappendix{Method} 
If the primal equation admits a solution, then the minimum norm solution is one potential solution and is given by:
\begin{align}
    \begin{pmatrix}h_*\\ \theta_0\end{pmatrix} =~& \Sigma_0^+\,\,  \E\left[\begin{pmatrix}{Z}\\ {D}\end{pmatrix} Y\right] &
    \Sigma_0 = \E\left[\begin{pmatrix}{Z}\\ {D}\end{pmatrix} \begin{pmatrix}{X}\\ {D}\end{pmatrix}^\top \right]
\end{align}
where $\Sigma_0^+$ is the pseudo-inverse of $\Sigma_0$. 

Thus, to validate that the primal admits a solution, we need to prove the minimum norm solution $\theta_0, h_*$, as defined by the above equations, also satisfy the primal moment restrictions:
\begin{align}
    \E\left[({Y} - {X}^\top  h_* - {D}\,\theta_0) \begin{pmatrix}{Z}\\ {D}\end{pmatrix}\right] =~& 0 
\end{align}

Equivalently, if $m(V; \theta, h) = ({Y} - {X}^\top h - {D}\,\theta)({Z}; {D})$, with $V=(X, D, Z, Y)$, then we want to test the null hypothesis that $\E[m(V; \theta_0, h_*)] = 0$.

Let $n$ be the number of samples and $p_Z+1$ the dimension of the instruments $({Z}; {D})$, i.e., the dimension of the moment restrictions. For simplicity assume that $n$ is even and that we split the data in half into training and testing samples. Let $\hat{h}, \hat{\theta}$ be estimates of the minimum norm solution using the training half of the data. Crucial to our proof is the assumption that that the minimum norm estimates are asymptotically linear estimators satisfying the following property:\footnote{The notation $o_p(1)$ denotes terms that converge to zero in probability as the sample size $n$ grows to infinity.}

\begin{align} \label{eq:min_norm_linear}
    \sqrt{\frac{n}{2}}  \Sigma_0 \left(\begin{pmatrix}\hat{h}\\ \hat{\theta}\end{pmatrix} - \begin{pmatrix}h_*\\ \theta_0\end{pmatrix}\right) = \sqrt{\frac{n}{2}} \,\, J_0\,\,  \E_{\text{train}}\left[\begin{pmatrix}{Z}\\ {D}\end{pmatrix} (Y - X^\top  h_* - D\theta_0)\right] + o_p(1)
\end{align}
where $J_0$ is some matrix and $\E_{\text{train}}$ is the empirical average over the training samples.

If we estimate the minimum norm solution $\hat{h}, \hat{\theta}$ using the $\ell_2$ regularized adversarial IV method as described in Appendix \ref{app:adviv}, and furthermore assume that $\E[(Z;D)(Z;D)^\top]$ is invertible, the above holds with:
\begin{align*}
    J_0 =~& \E\left[\begin{pmatrix}{Z}\\ {D}\end{pmatrix} \begin{pmatrix}{Z}\\ {D}\end{pmatrix}^\top\right]^{1/2} \tilde{\Sigma}_0 \tilde{\Sigma}_0^{+} \E\left[\begin{pmatrix}{Z}\\ {D}\end{pmatrix} \begin{pmatrix}{Z}\\ {D}\end{pmatrix}^\top\right]^{-1/2} &
    \tilde{\Sigma}_0 = \E\left[\begin{pmatrix}{Z}\\ {D}\end{pmatrix} \begin{pmatrix}{Z}\\ {D}\end{pmatrix}^\top\right]^{-1/2} \Sigma_0
\end{align*}
assuming we choose the $\ell_2$ penalty on the parameters to decay faster than $1/\sqrt{n}$. We provide a proof for this asymptotic linearity property of the adversarial IV method in Appendix \ref{app:iv-linearity}.\footnote{We remark that if we apply a pre-processing step in this test and apply a PCA transformation to the instruments $(Z;D)$, replacing them with their PCA transformation then for the transformed instruments, we will have that $\E[(Z;D)(Z;D)^\top]=I$ and the formula simplifies to $J_0=\Sigma_0 \Sigma_0^+$.}

Consider the empirical average of the moment on the test set using the minimum norm estimates calculated on the training set, i.e.:
\begin{align*}
    \hat{M} = \E_{\text{test}}[m(V;\hat{\theta}, \hat{h})]
\end{align*}
where $\E_{\text{test}}$ is the empirical average over the test samples. Let $M_0 = \E[m(V;\theta_0, h_*)]$, which is zero under the null. By influence function calculus arguments (see e.g. \cite[Theorem~6.1]{newey1994large}) and since the Jacobian of the moment $m$ with respect to the vector $(h;\theta)$ is $\Sigma_0$, the estimate $\hat{M}$ admits the asymptotic linear representation:
% \vscomment{The second follows by the equation above, i.e. the linear representation of $\hat{h}$. The first we cannot expand; it is a whole proof. I cited and expanded a bit. See also Lemma 36 here \url{https://www.dropbox.com/scl/fi/a4t5pqmmauxvfa2j32hya/MSANE328_CS328.pdf?rlkey=bqe5sc1nre4s7ooartpmldmls&e=1&st=wjm7w8dt&dl=0}}
\begin{align*}
    \sqrt{\frac{n}{2}} (\hat{M} - M_0) =& \sqrt{\frac{n}{2}}\,\, \E_{\text{test}}[m(V;\theta_0, h_*) - M_0] - \sqrt{\frac{n}{2}} \Sigma_0  \left(\begin{pmatrix}\hat{h}\\ \hat{\theta}\end{pmatrix} - \begin{pmatrix}h_*\\ \theta_0\end{pmatrix}\right) + o_p(1)\\
    =& \sqrt{\frac{n}{2}}\,\, \E_{\text{test}}[m(V;\theta_0, h_*) - M_0] - \sqrt{\frac{n}{2}} J_0\,\,  \E_{\text{train}}\left[\begin{pmatrix}{Z}\\ {D}\end{pmatrix} (Y - X^\top  h_* - D\theta_0)\right] + o_p(1)\\
\end{align*}
where the second line derives from Equation \eqref{eq:min_norm_linear}. Thus we can derive the asymptotic linear representation of our estimate $\hat{M}$ as:
\begin{align*}
    \sqrt{n} (\hat{M} - M_0) = \sqrt{n} \E_n\left[(m(V;\theta_0, h_*) - M_0) \frac{1\{\text{test}\}}{\Pr(\text{test})} - J_0\,\,\begin{pmatrix}{Z}\\ {D}\end{pmatrix} (Y - X^\top  h_* - D\theta_0) \frac{1\{\text{train}\}}{\Pr(\text{train})}\right] + o_p(1)
\end{align*}
From this, we obtain that the estimate $\hat{M}$ is asymptotically normal:
\begin{align}
    \sqrt{n} (\hat{M} - M_0) \rightarrow_d N(0, A)
\end{align}
with asymptotic variance:
\begin{align}
    A = \E[(m(V;\theta_0, h_*) - M_0)^2 \mid 1\{\text{test}\}] \frac{1}{\Pr(\text{test})} + \E[\Phi\Phi^\top  \mid 1\{\text{train}\}]\frac{1}{\Pr(\text{train})} 
\end{align}
where:
\begin{align}
    \Phi =~& J_0\,\,\begin{pmatrix}{Z}\\ {D}\end{pmatrix} (Y - X^\top  h_* - D\theta_0) \\
    =~& \E\left[\begin{pmatrix}{Z}\\ {D}\end{pmatrix} \begin{pmatrix}{Z}\\ {D}\end{pmatrix}^\top\right]^{1/2} \tilde{\Sigma}_0 \tilde{\Sigma}_0^{+} \E\left[\begin{pmatrix}{Z}\\ {D}\end{pmatrix}^\top\right]^{-1/2} \,\begin{pmatrix}{Z}\\ {D}\end{pmatrix} (Y - X^\top  h_* - D\theta_0)
\end{align}

% \klcomment{It is unclear to me why we want to soft threshold / how $n^{-.2}$ was chosen}\vscomment{Again this part needs way more analytic derivations and basically I would say a theory paper to be written that works this out very carefully :) The MultiDimMediator.ipynb notebook allows you to play with different values of the exponent in the regularization and check coverage performance. $n^{-0.5}$ is too small; especially when the mediator is very low dimensional (e.g. 1-dimensional). $n^{-0.2}$ works well across a variety of mediator sizes. We can point to that notebook in the github if people want to play with this. I expanded as much as I thought was reasonable here.}

We will construct an empirical estimate $\hat{A}$ of the variance $A$. Our first goal is to construct an estimate of the orthogonal projector matrix:
\begin{align}
    P\defeq \tilde{\Sigma}_0 \tilde{\Sigma}_0^{+} = U DD^+ U^\top
\end{align}
where the SVD of $\tilde{\Sigma}_0 = U D V^\top$, with $D=\text{diag}(\lambda_1, \ldots, \lambda_K)$ and $K=p_Z+1$. To estimate $\hat{P}$, we consider the empirical estimate of $\tilde{\Sigma}_n$ of $\tilde{\Sigma}_0$ as $$\tilde{\Sigma}_n=\E_{\text{train}}\left[\begin{pmatrix}{Z}\\ {D}\end{pmatrix} \begin{pmatrix}{Z}\\ {D}\end{pmatrix}^\top \right]^{-1/2} \E_{\text{train}}\left[\begin{pmatrix}{Z}\\ {D}\end{pmatrix} \begin{pmatrix}{X}\\ {D}\end{pmatrix}^\top \right]$$ and its SVD as  $\tilde{\Sigma}_n=\hat{U} \hat{D} \hat{V}^\top $ with $\hat{D}=\text{diag}(\hat{\lambda}_1, \ldots, \hat{\lambda}_K)$. In Appendix \ref{app:soft_thresh}, we show we can estimate $\hat{P}$ by ``soft-thresholding'' the empirical singular values $\hat{\lambda}_i$ such that 
\begin{align}
    \hat{P} = \hat{U} \hat{D}\hat{D}^+ \hat{U}^\top = \hat{U}\, \text{diag}\left(\frac{\hat{\lambda}_1}{\hat{\lambda}_1 + n^{-0.2}}, \ldots, \frac{\hat{\lambda}_K}{\hat{\lambda}_K + n^{-0.2}}\right)\, \hat{U}^\top
\end{align}

We can then estimate the variance $A$ using the estimates $\hat{P}, \hat{\Phi},$ and $\hat{J}$, i.e.,
\begin{align*}
    \frac{\hat{A}}{n} =~& \frac{1}{n_{\text{test}}} \E_{\text{test}}[(m(V;\hat{\theta}, \hat{h}) - \hat{M})^2] + \frac{1}{n_{\text{train}}} \E_{\text{train}}[\hat{\Phi}\hat{\Phi}^\top ]\\
        \hat{\Phi} =~& \hat{J}\,\,\begin{pmatrix}{Z}\\ {D}\end{pmatrix} (Y - X^\top  \hat{h} - D\hat{\theta}) \\
\hat{J} =~& \E_{\text{train}}\left[\begin{pmatrix}{Z}\\ {D}\end{pmatrix} \begin{pmatrix}{Z}\\ {D}\end{pmatrix}^\top\right]^{1/2}\,\, \hat{P} \,\, \E_{\text{train}}\left[\begin{pmatrix}{Z}\\ {D}\end{pmatrix} \begin{pmatrix}{Z}\\ {D}\end{pmatrix}^\top\right]^{-1/2}
 \end{align*}
Under regularity assumptions, the estimated variance $\hat{A}$ will be consistent and the following asymptotic normality assumption will also hold:
\begin{align}
    \sqrt{n}\, \hat{A}^{-1/2}\, (\hat{M} - M_0) \to_d N(0, I)
\end{align}
Since the random vector $\sqrt{n}\, \hat{A}^{-1/2}\, (\hat{M} - M_0)$ asymptotically follows the distribution of $p_Z+1$ independent standard Gaussian random variables, then the squared $\ell_2$-norm of this vector, i.e. $\norm{\sqrt{n}\, \hat{A}^{-1/2}\, (\hat{M} - M_0) }^2_2$, asymptotically follows the distribution of the sum \textit{of the squares} of $p_Z+1$ independent standard Gaussian random variables, which corresponds to the $\chi^2$ distribution with $p_Z+1$ degrees of freedom. Thus we have:
\begin{align}
    n (\hat{M} - M_0)^\top  \hat{A}^{-1} (\hat{M} - M_0) \to_d \chi^2(\text{dof}=p_Z+1)
\end{align}
Under the null hypothesis the primal yields a solution, $M_0=0$ and thus we have that:
\begin{align}
    n \hat{M}^\top  \hat{A}^{-1} \hat{M} \to_d \chi^2(\text{dof}=p_Z+1)
\end{align}
Thus, we can use the test statistic:
\begin{align}
    \hat{T} = n \hat{M}^\top  \hat{A}^{-1} \hat{M} \tag{Primal test statistic}
\end{align}
and compare it with the quantiles of a $\chi^2(\text{dof}=p_Z+1)$ distribution to reject the null. In particular, if we want to erroneously reject the null with probability at most $\alpha$, then we should be rejecting the null if:
\begin{align}
    \hat{T} > z_{1-\alpha}
\end{align}
where $z_{1-\alpha}$ is the $1-\alpha$ percentile of a $\chi^2(\text{dof}=p_Z+1)$ distribution.

\subsubappendix{Soft-thresholding the orthogonal projection matrix $P$}\label{app:soft_thresh}

We expand here on estimating $P\defeq \tilde{\Sigma}_0 \tilde{\Sigma}_0^{+}$ using soft-thresholding of the singular values. We apply this to the primal and dual equations for their respective definitions of $\tilde{\Sigma}_0 = U D V^\top, \tilde{\Sigma}_n = \hat{U} \hat{D} \hat{V}^\top$. Explicitly, for the primal, recall that we define $$ \tilde{\Sigma}_0 = \E\left[\begin{pmatrix}{Z}\\ {D}\end{pmatrix} \begin{pmatrix}{Z}\\ {D}\end{pmatrix}^\top\right]^{-1/2}\E\left[\begin{pmatrix}{Z}\\ {D}\end{pmatrix} \begin{pmatrix}{X}\\ {D}\end{pmatrix}^\top \right]$$ and for the dual we define $$ \tilde{\Sigma}_0 = \E[Z Z^\top]^{-1/2}\E[ZX^\top ]$$

Recall that given the SVD $\tilde{\Sigma}_0 = U D V^\top$, we have $P = U DD^+ U^\top$. Note that $DD^+ = \text{diag}(1\{\lambda_1 \neq 0\}, \ldots, 1\{\lambda_K \neq 0\})$ where we expect only a small number $k<K$ of non-zero singular values $\lambda_i$. In essence, $P$ is the orthogonal projector onto the range of $\tilde{\Sigma}_0$.

We estimate the orthogonal projector as $\hat{P}$ using the empirical estimate and corresponding SVD $\tilde{\Sigma}_n=\hat{U} \hat{D} \hat{V}^\top $ with $\hat{D}=\text{diag}(\hat{\lambda}_1, \ldots, \hat{\lambda}_K)$. Unlike the population singular values $\lambda_i$, the empirical singular values $\hat{\lambda}_i$ are typically all non-zero in practice due to sampling noise. Thus if we simply estimate $P$ as $\hat{P} = \hat{U} \hat{D}\hat{D}^+ \hat{U}^\top$, then $\hat{D}\hat{D}^+ = \vec 1$ (the $K$-dimensional vector of 1's) and we will get the projection matrix $\hat{P} = \hat{U} \hat{U}^\top = I$.

To construct a projection matrix $\hat{P}$ more robust to sampling noise, note by the Wedin's theorem \cite{wedin1972perturbation} (see also [\citen{davis1970rotation,yu2015useful}]), the space spanned by the top $k$ (where recall $k = \text{rank}(\tilde{\Sigma}_0)$) eigenvectors of $\hat{U}$ is approximately the same as the space spanned by the top $k$ eigenvectors of $U$, since $\tilde{\Sigma}_n$ is a consistent estimate of $\tilde{\Sigma}_0$. Thus if we can construct a matrix $\hat{\Lambda}=\text{diag}(v)$, where $v$ converges to $(1, \ldots, 0)$ with a prefix of $k$ ones, then the matrix $\hat{P}=\hat{U} \hat{\Lambda} \hat{U}^\top$ will be a consistent estimate of the matrix $P$. We will accomplish that by applying soft-thresholding to the empirical singular values:
\begin{align*}
\hat{P} =~& \hat{U}\, \text{diag}\left(\frac{\hat{\lambda}_1}{\hat{\lambda}_1 + n^{-0.2}}, \ldots, \frac{\hat{\lambda}_K}{\hat{\lambda}_K + n^{-0.2}}\right)\, \hat{U}^\top\end{align*}

To motivate this choice of soft-thresholding to estimate $\hat{P},$ first note that by Weyl's theorem \cite{weyl1912asymptotische}, the absolute difference between any of the singular values $\hat{\lambda}_i$ of $\tilde{\Sigma}_n$ and the corresponding singular values $\lambda_i$ of $\tilde{\Sigma}$ are upper bounded by the operator norm of the difference $\tilde{\Sigma}_n-\tilde{\Sigma}_0$, i.e.,\footnote{Recall that, for a matrix $A$, the operator norm $\norm{A}_{op}$ is intuitively the largest factor by which $A$ ``stretches'' a vector. This is also the largest absolute singular value of the matrix $A$.}
\begin{align*}
    \forall j \in [K]: |\hat{\lambda}_j - \lambda_j| \leq \|\tilde{\Sigma}_n - \tilde{\Sigma}_0\|_{op}
\end{align*}
Thus, we intuitively want to find an upper bound for $\|\tilde{\Sigma}_n - \tilde{\Sigma}_0\|_{op}$, which will provide insight into how to construct a new matrix $\hat{\Lambda}$ such that we can bound the error on the true singular values $\lambda_i$.

First, note that, for both the dual and the primal, $\tilde{\Sigma}_0$ takes the form $A^{-1/2} B$, with $A=\E[WW^\top]$ and $B=\E[VW^\top]$ for appropriately defined random vectors $W$ and $V$ and $A$ assumed to be invertible (e.g., for the dual, $A = \E[ZZ^\top]$ and $B = \E[XZ^T]$). Accordingly, $\tilde{\Sigma}_n$ takes the form $A_n^{-1/2}B_n$, with $A_n=\E_n[WW^\top]$ and $B_n=\E_n[VW^\top]$. We can thus upper bound the operator norm:
\begin{align}
    \|\tilde{\Sigma}_n - \tilde{\Sigma}_0\|_{op} = \|A_n^{-1/2}B_n - A^{-1/2}B\|_{op} =~& \|(A_n^{-1/2} - A^{-1/2}) B_n + A^{-1/2}(B_n - B)\|_{op}\\
    \leq~& \|A_n^{-1/2} - A^{-1/2}\|_{op} \|B_n\|_{op} + \|A^{-1/2}\|_{op} \|B_n - B\|_{op} \label{app:error_bound}
\end{align}
Furthermore, if we use the fact that for any matrix $M$, $\norm{M}_{op} \leq \norm{M}_F$, and then apply standard concentration inequality arguments (Chernoff-Hoeffding bounds), we have that $\|A_n - A\|_{op}\leq \|A_n - A\|_{F}=O_p(n^{-1/2})$ and, similarly, $\|B_n-B\|_{op}\leq \|B_n-B\|_{F}=O_p(n^{-1/2})$. Then, by the triangle inequality we can derive that $\|B_n\|_{op} \leq \|B\|_{op} + O_p(n^{-1/2})$. Since $\|B\|_{op}$ is some constant independent of $n$, then $ \|B_n\|_{op} = O_p(n^{-1/2})$. Also, note that since $A$ is invertible, we have $\|A^{-1/2}\|_{op}<\infty$ and therefore $\|A^{-1/2}\|_{op}$ is some constant independent of $n$. Thus $\|A^{-1/2}\|_{op}=O(1)$. Thus, applying these derivations to Equation \eqref{app:error_bound}, we get that: 
\begin{align*}
    \|\tilde{\Sigma}_n - \tilde{\Sigma}\|_{op} = O_p\left(\|A_n^{-1/2} - A^{-1/2}\|_{op} (1 + n^{-1/2}) + n^{-1/2}\right)
\end{align*}
We next derive an upper bound on $\|A_n^{-1/2} - A^{-1/2}\|_{op}$. By continuity of the inverse (see e.g. [\citen{stewart1969continuity}] or [\citen{wilkinson2023rounding}]) we have that as long as $\|A^{-1/2}\|_{op} \|A_n^{1/2} - A^{1/2}\|_{op} < 1$ then we have that: 
\begin{align*}
    \|A_n^{-1/2} - A^{-1/2}\|_{op} \leq \frac{\|A^{-1/2}\|_{op} \|A_n^{1/2} - A^{1/2}\|_{op}}{1 - \|A^{-1/2}\|_{op} \|A_n^{1/2} - A^{1/2}\|_{op}} = O\left(\|A_n^{1/2} - A^{1/2}\|_{op}\right)
\end{align*}
Moreover, we have that:\footnote{Consider any eigenvector $x$ of $\sqrt{A_n}-\sqrt{A}$ with eigenvalue $\mu$. Then: 
\begin{align*}
    x^\top (A_n - A) x = x^\top (\sqrt{A_n} - \sqrt{A}) \sqrt{A_n} x + x^\top \sqrt{A} (\sqrt{A_n} - \sqrt{A}) x = \mu x^\top (\sqrt{A_n} + \sqrt{A}) x
\end{align*}
Since the operator norm corresponds to the largest absolute eigenvalue, for symmetric matrices, and since $A, A_n$ are symmetric, we can choose $\mu=\pm \|\sqrt{A_n} - \sqrt{A}\|_{op}$, which yields:
\begin{align*}
    \|\sqrt{A_n} - \sqrt{A}\|_{op} \leq \frac{x^\top (A_n - A) x}{x^\top (\sqrt{A_n} + \sqrt{A})x} \leq \frac{\|A_n - A\|_{op}}{\lambda_{\min}(\sqrt{A_n} + \sqrt{A})}\leq \frac{\|A_n - A\|_{op}}{\lambda_{\min}(\sqrt{A})} = \frac{\|A_n - A\|_{op}}{\sqrt{\lambda_{\min}(A)}}
\end{align*}
where we also used the fact that $\sqrt{A}$ and $\sqrt{A_n}$ are positive semi-definite.
}
\begin{align*}
    \|A_n^{1/2} - A^{1/2}\|_{op} \leq \frac{1}{\sqrt{\lambda_{\min}(A)}}\|A_n-A\|_{op} = O(\|A_n-A\|_{op}) = O_p(n^{-1/2})
\end{align*}
where $\lambda_{\text{min}}(A)$ is the minimum eigenvalue of $A$. Note, then, that $\|A^{-1/2}\|_{op}\|A^{1/2}-A_n^{1/2}\|_{op} = O_p(n^{-1/2})$ will eventually be smaller than $1$. Then we satisfy the aforementioned property $\|A^{-1/2}\|_{op} \|A_n^{1/2} - A^{1/2}\|_{op} < 1$ and can invoke the upper bound $\|A_n^{-1/2} - A^{-1/2}\|_{op}=O(\|A_n^{1/2}-A^{1/2}\|_{op})=O_p(n^{-1/2})$.
Finally we can conclude that:
\begin{align*}
    \forall j\in [K]: |\lambda_j - \hat{\lambda}_j| \leq \|\tilde{\Sigma}_n - \tilde{\Sigma}\|_{op} = O_p(n^{-1/2})
\end{align*}

For any $j \leq k$, the true singular value $\lambda_j \geq 0$ and thus as $n^{-0.2}$ converges to zero, $\hat{\lambda}_j$ will be bounded away from zero (since by the triangle inequality we have that $\hat{\lambda}_j \geq \lambda_j - O_p(n^{-0.5})$). Thus $\hat{\lambda}_j / (\hat{\lambda}_j + n^{-0.2}) \to 1$. For any $j>k$, we have that $\lambda_j = 0$ and therefore we expect that $|\hat{\lambda}_j| = |\hat{\lambda}_j - \lambda_j| = O_p(n^{-0.5})$. Thus eventually the $n^{-0.2}$ term will become the dominating term and we will have that $\hat{\lambda}_j / (\hat{\lambda_j} + n^{-0.2}) \to 0$. Note that any regularization $n^{-\alpha}$ for $\alpha < 0.5$ would satisfy the above properties. Experimentally we found that $\alpha=0.2$ performs well empirically and gave nominal coverage across a wide range of dimensions of the hidden mediator.\footnote{The reader can experiment with different values of $\alpha$ and different dimensions $p_M$ of the hidden mediator using this notebook: \url{https://github.com/syrgkanislab/hidden_mediators/blob/main/MultiDimMediator.ipynb}.}

\subappendix{Dual violation test}\label{appendix:test-dual}
\subsubappendix{Intuition}
Recall the \eqref{eq:dual} is \begin{align}
        \E\left[(D - Z^\top  \gamma)\, X\right] =~& 0 
\end{align}
We want to test whether this equation admits a solution $\gamma_*$. This is a linear system of equations:
\begin{align}
    \E[XZ^\top] \gamma = \E[X D]
\end{align}
Since all variables correspond to residuals and hence are mean-zero, the above is equivalent to:
\begin{align}
    \cov(X, Z) \gamma = \cov(X, D)
\end{align}
Thus, the dual admits a solution if $\cov(X, D) \in \text{column-span}(\cov(X, Z))$. Intuitively, this means that all the information flowing between $D$ and $X$ must go through the same mediator $M$ that influences $X$ and $Z$. This would not be true if either of the following two (equivalent) assumptions were violated: (1) there was some proxy variable $X_i$ which has an alternate path from $D$ not through $Z$'s mediator $M$, i.e., either $\exists X_i$ such that $D$ has a direct influence on $X_i$ or $\exists  X_i, M''$ s.t. $D \rightarrow M'' \rightarrow X_i$ and $M'' \cancel{\rightarrow} Z$; (2) the proxies $Z$ are not correlated with the mediator $M$, i.e., $Z$ has no direct edge from $M$ nor an indirect correlation of $M$ through $D$.

If we view the dual violation as an issue of the first scenario (i.e., some $X_i$ has an alternate mediator $M''$), then we can mitigate the violation by identifying and removing such a $X_i$ via our proxy selection algorithm. If we view the violation from the second scenario (i.e., $X$'s mediator $M$ has no direct relationship to all proxy variables $Z$), we could try adding stronger, more informative proxies to $Z$ that are likely to be directly influenced by $M$.

\subsubappendix{Method} 
The proof is symmetric to that of the primal. 
If the dual equation admits a solution, then the minimum norm solution is one potential solution and is given by:
\begin{align}
    \gamma_* =~& \Sigma_0^+\,\,  \E \left[X D \right] &
    \Sigma_0 = \E\left[X Z^\top \right]
\end{align}
where $\Sigma_0^+$ is the pseudo-inverse. 
Thus, to validate that the dual admits a solution, we need to prove the minimum norm solution $\gamma_*$, as defined by the above equations, also satisfy the dual moment restrictions: \begin{align}
    \E\left[(D - Z^\top  \gamma_*) X\right] =~& 0 
\end{align}
Equivalently, if $m(V; \gamma) = (D - Z^\top  \gamma) X$, with $V=(X, D, Z)$, then we want to test the null hypothesis that $\E[m(V; \gamma_*)] = 0$.

Let $n$ be the number of samples and $p_X$ the dimension of $X$, i.e., the dimension of the dual moment restrictions. For simplicity assume that $n$ is even and that we split the data in half into training and testing samples. Let $\hat{\gamma}$ be the estimate of the minimum norm solution using the training half of the data. Crucial to our proof is the assumption that that the minimum norm estimate is an asymptotically linear estimator satisfying the following property:
\begin{align*}
    \sqrt{\frac{n}{2}}   \Sigma_0 \left(\hat{\gamma} - \gamma_*\right) = \sqrt{\frac{n}{2}}  \,\, J_0\,\,  \E_{\text{train}}\left[X (D - Z^\top  \gamma_*)\right] + o_p(1)
\end{align*}
In a similar line of reasoning as with the primal, we have the above holds when solving for the minimum norm solution $\hat{\gamma}$ using adversarial IV (assuming $\E[XX^\top]$ is invertible) when
\begin{align}
    J_0 =~& \E[XX^\top]^{1/2} \tilde{\Sigma}_0\tilde{\Sigma}_0^+ \E[XX^\top]^{-1/2} &
    \tilde{\Sigma}_0 =~& \E[XX^\top]^{-1/2} \Sigma_0
\end{align}
and the $\ell_2$ penalty on the parameters decays faster than $1/\sqrt{n}$.

Consider the empirical average of the moment on the test set using the minimum norm estimates calculated on the training set, i.e.:
\begin{align*}
    \hat{M} = \E_{\text{test}}[m(V;\hat{\gamma})]
\end{align*}
Let $M_0 = \E[m(V;\gamma_*)]$, which is zero under the null. By similar arguments as with the primal, we have the asymptotic linear representation
\begin{align*}
    \sqrt{n} (\hat{M} - M_0) = \sqrt{n} \E_n\left[(m(V;\gamma_*) - M_0) \frac{1\{\text{test}\}}{\Pr(\text{test})} + J_0\,\,X (D - Z^\top  \gamma_*) \frac{1\{\text{train}\}}{\Pr(\text{train})}\right] + o_p(1)
\end{align*}
From this, we obtain that the estimate $\hat{M}$ is asymptotically normal:
\begin{align}
    \sqrt{n} (\hat{M} - M_0) \rightarrow_d N(0, A)
\end{align}
with variance:
\begin{align}
    A = \E[(m(V;\gamma_*) - M_0)^2 \mid 1\{\text{test}\}] \frac{1}{\Pr(\text{test})} + \E[\Phi\Phi^\top  \mid 1\{\text{train}\}]\frac{1}{\Pr(\text{train})} 
\end{align}
where:
\begin{align}
    \Phi =~& J_0\,\,X (D - Z^\top  \gamma_*) \\ 
    =~& \E[XX^\top]^{1/2} \tilde{\Sigma}_0\tilde{\Sigma}_0^+ \E[XX^\top]^{-1/2} \,X (D - Z^\top  \gamma_*)
\end{align}

We will then construct an empirical estimate $\hat{A}$ of the variance $A$. If we denote the empirical variant of $\tilde{\Sigma}_0$ as $$\tilde{\Sigma}_n=\E_{\text{train}}[XX^\top]^{-1/2}\E_{\text{train}}[X Z^\top]$$ where its SVD is  $\tilde{\Sigma}_n=\hat{U} \hat{D} \hat{V}^\top $ with $\hat{D}=\text{diag}(\hat{\lambda}_1, \ldots, \hat{\lambda}_K)$ then we can estimate the orthogonal projection matrix $ P\defeq \tilde{\Sigma}_0 \tilde{\Sigma}_0^{+}$ similarly to that of the primal by soft thresholding to the singular values (as described in Appendix \ref{app:soft_thresh}), i.e.,
\begin{align*}
\hat{P} =~& \hat{U}\, \text{diag}\left(\frac{\hat{\lambda}_1}{\hat{\lambda}_1 + n^{-0.2}}, \ldots, \frac{\hat{\lambda}_K}{\hat{\lambda}_K + n^{-0.2}}\right)\, \hat{U}^\top\end{align*}
We can then estimate the variance $A$ using the estimates $\hat{P}, \hat{\Phi},$ and $\hat{J}$, i.e.,
\begin{align*}
    \frac{\hat{A}}{n} =~& \frac{1}{n_{\text{test}}} \E_{\text{test}}[(m(V;\hat{\gamma}) - \hat{M})^2] + \frac{1}{n_{\text{train}}} \E_{\text{train}}[\hat{\Phi}\hat{\Phi}^\top ]\\
        \hat{\Phi} =~& \hat{J}\,\,X (D - Z^\top  \hat{\gamma}) \\
\hat{J} =~& \E_{\text{train}}[XX^\top]^{1/2} \tilde{\Sigma}_n \tilde{\Sigma}_n^+\,\, \E_{\text{train}}[XX^\top]^{-1/2}
\end{align*}

Under regularity assumptions, the estimated variance $\hat{A}$ will be consistent and the following asymptotic normality assumption will also hold:
\begin{align}
    \sqrt{n}\, \hat{A}^{-1/2}\, (\hat{M} - M_0) \to_d N(0, I)
\end{align}
Similar to the case of primal, the squared $\ell_2$ norm $\norm{\sqrt{n}\, \hat{A}^{-1/2}\, (\hat{M} - M_0) }^2_2$ asymptotically follows the distribution of the sum \textit{of the squares} of $p_X$ independent standard Gaussian random variables, which corresponds to the $\chi^2$ distribution with $p_X$ degrees of freedom. Under the null hypothesis the dual yields a solution, $M_0=0$ and thus we have that:
\begin{align}
    n \hat{M}^\top  \hat{A}^{-1} \hat{M} \to_d \chi^2(\text{dof}=p_X)
\end{align}
Thus, we can use the test statistic:
\begin{align}
    \hat{T} = n \hat{M}^\top  \hat{A}^{-1} \hat{M} \tag{Dual test statistic}
\end{align}
and compare it with the quantiles of a $\chi^2(\text{dof}=p_X)$ distribution to reject the null. In particular, if we want to erroneously reject the null with probability at most $\alpha$, then we should be rejecting the null if:
\begin{align}
    \hat{T} > z_{1-\alpha}
\end{align}
where $z_{1-\alpha}$ is the $1-\alpha$ percentile of a $\chi^2(\text{dof}=p_X)$ distribution.

\subappendix{Instrument identification strength}
Recall that our method estimates $V = ({D}-\gamma_*^\top {Z})$ where $\gamma_*$ satisfies the \eqref{eq:dual}, and $V$ is now a new instrument in the \eqref{eq:primal} that guarantees Neyman orthogonality to nuisance parameters $h_*, \gamma_*$ when estimating $\theta$. Let $\bar{Y} = {Y} - {X}^\top h_*$, and then the primal equation can be rewritten as the IV moment restriction equation 
 \begin{align}\label{eq:primalIV-app}
     \E[V(\bar{Y} - {D}\,\theta)] = 0 \Leftrightarrow \theta_0 = \frac{\E[V\bar{Y}]}{\E[V{D}]}
 \end{align} 
 where $V$ is the instrument, $\bar{Y}$ is the outcome, and ${D}$ the treatment. If the denominator $\E[V{D}]$ is small, this can cause instability to the estimate and invalidate the resulting confidence interval. This is exactly the type of condition that weak instrument tests check for. Hence, we propose two weak instrument tests with $V$ as the instrument and $D$ as the treatment.

 \subsubappendix{Intuition} 

The denominator $\E[V D]$ is small if $D-\gamma_*^\top Z$ is uncorrelated with $D$. Note that $\gamma_*$ is the solution to the IV moment $\E[X (D - \gamma^\top Z)]=0$, where we are trying to predict $D$ as a function of $Z$ with instrument $X$. By definition, $\gamma_*^\top Z$ will capture the part of $D$ that is predictable from $Z$ through the mediator $M$. If $M$ is very strongly correlated with $D$, then $\gamma_*^\top Z$ would be able to almost perfect predict $D$ and thus the remnant variation $D-\gamma_*^\top Z$ is too small and uncorrelated with $D$. In essence, if the mediator is very correlated with $D$, then we will not be able to strongly identify the direct effect from the mediated effect of $D$ on $Y$.

For instance, consider the case when $D$ is a zero-meaned binary variable and $X, Z, M$ are scalar random variables:
 \begin{align}
    D \sim~& \text{Bernoulli(p)} - p\\
    M =~& a D + \sigma_m  N(0, 1)\\
    Z =~& e M + \sigma_z  N(0, 1)\\
    X =~& f M + \sigma_x  N(0, 1)
 \end{align}
 Then
 \begin{align*}
    \E[M^2] =& \sigma_m^2 + a^2 \E[D^2], &
    \E[XZ] =& f \cdot e \cdot \E[M^2], & 
    \E[DZ] =& a \cdot e \cdot \E[D^2], &
    \E[DX] =& a \cdot f \cdot \E[D^2]
\end{align*}
Hence, the parameter $\gamma_*$ and the denominator that determines the identification strength takes the form:
\begin{align*}
    \gamma_* =& \frac{\E[DX]}{\E[XZ]} = \frac{a}{e} \frac{\E[D^2]}{\E[M^2]} & 
    \E[VD] =~& \E[D^2] - \gamma_* \E[DZ]= \E[D^2] \frac{1}{1 + a^2 \E[D^2] / \sigma_m^2}
 \end{align*}
Thus we see that if the ratio $a/\sigma_m$ (where recall $a$ is the direct effect of $D \rightarrow M$) is very large, then we have weak identification. In other words, weak identification occurs when the mediator is primarily dictated by the treatment $D$ and not other exogenous sources of variation. For instance, if $\sigma_m=0$, then we cannot identify the direct effect. Moreover, we also have weak identification if $\E[D^2]$ is small, i.e. if the treatment is very deterministic.

\subsubappendix{Method}\label{app:weakiv_tests_method}
We consider two test statistics that target the weak identification problem. The first is based on a standard first stage F-test that is typically used in IV regression. However, here we also need to modify this statistic to account for the extra variance that stems from the estimation of the unknown parameter $\gamma_*$ that enters the definition of the instrument $V$.

% \klcomment{I still think this should be $\pi$ because we are referring to the general equation (and thus the general param) from which we are estimating for? }\vscomment{If it was a moment equation, I'd agree. But when you write a regression equation, you typically use the true parameter.}
\paragraph{(a) $V$ strength F-test} - We develop an effective F-test\cite{olea2013robust,andrews2005inference} to check the correlation strength of $V$ with ${D}$. We view our problem as a single instrument, single treatment setting with instrument $V$, treatment $D$, and outcome $Y$. Note, the first-stage IV equation from Equation \eqref{eq:primalIV-app} is ${D} = \pi_0 V + \epsilon$ where $\epsilon$ is the exogenous error term. The null hypothesis of a weak instrument test assumes $\pi$ takes a very small value which induces a bias in the second-stage instrumental variable estimate $\hat{\theta}$ of approximate magnitude at least $\tau$. A frequently used heuristic for $\tau$ is the Nagar bias, which posits that for a true first stage parameter $\pi_0$, an estimate $\hat{\pi}$ with variance $\sigma_{\pi}$ induces a bias on 
the second-stage estimate of approximately $\tau=1/\lambda^2$, where $\lambda^2 = \pi_0^2 / \sigma_\pi^2$ \cite{andrews2005inference}. To check whether we have a weak instrument we use a heteroskedastic robust F-statistic (which is equal to the effective F-statistic when there is only one instrument) from [\citen{stock_2018}] where, given a dataset with $n$ samples, we estimate
 \begin{align}\label{eq:fteststat}
     F = \frac{\hat{\pi}^2}{\hat{\sigma}_{\pi}^2} \tag{F-test statistic}
 \end{align}
 where $\hat{\pi} = \frac{\E_n[D \hat{V}]}{\E_n[{D}^2]}$ (with $\hat{V}=D - Z^\top\hat{\gamma}$) is the OLS estimate and ${\hat{\sigma}_{\pi}^2}$ is an estimate of the variance $\sigma_{\pi}^2$ of $\hat{\pi}$.

First let us assume that the estimate $\hat{\pi}$ satisfies asymptotic normality, i.e. $\sqrt{n}(\hat{\pi} - \pi_0) \to_d N(0, \tilde{\sigma}_\pi^2)$ where $\sigma_\pi^2 = \tilde{\sigma}_\pi^2 / n$, and then we also have that
 \begin{align}
     \frac{\hat{\pi}}{\hat{\sigma}_\pi} \approx N\left(\frac{\pi_0}{\sigma_\pi}, 1\right)  
 \end{align}
Then, under the null hypothesis of a weak instrument, i.e., that the Nagar bias is at most $\tau_*$ for some user-specified level $\tau_*$ (typically $0.1$), we have the $F$-statistic $\hat{\pi}^2 /{\hat{\sigma}_\pi^2}$ follows approximately a non-central $\chi_{1,c}^2$ distribution with $1$ degree of freedom and centrality parameter $c = \lambda^2 \leq 1/\tau_*$. With confidence $\alpha$, we can reject the null if the test statistic $F$ is larger than the $1-\alpha$ percentile of the $\chi_{1,1/\tau_*}^2$ distribution.

To compute the $F$-statistic, we need to estimate the variance ${\hat{\sigma}_{\pi}^2}$. The typical form of the heteroskedasticity robust asymptotic (not sample-size normalized) variance $\tilde{\sigma}_\pi^2$ is $\E[V^2 (D - \pi V)^2] / \E[V^2]^2$.
 However, in our setting, the instrument $V = D - \hat{\gamma}^\top Z$ contains the estimate $\hat{\gamma}$ which introduces extra variance in the estimate $\hat{\pi}$ that we need to account for. Recall $\hat{\gamma}$ is the minimum norm solution to the regularized adversarial IV method (see Appendix \ref{app:adviv}) with instrument $X$, treatment $Z$, and outcome $D$. For the purpose of this test, we will use a heuristic argument in the influence function of the estimate $\hat{\gamma}$:\footnote{A more formally justified correction term to the variance requires further theoretical investigation that is beyond the scope of the current work.} 

 % \klcomment{sorry how was this influence function derived? it seems different from the influence function in the advIV section in $\ref{app:iv-linearity}$, It looks similar to pg 74 of your MSE textbook though.}\vscomment{It is a heuristic analogue of what you would use if $\E[QQ^\top]$ was invertible. Now that it is not, we just add the $\lambda$. In retrospect I don't like that much this argument. I think after you re-run the tests, I think we should just remove this part completely and only use the Neyman orthogonal versions in the next section.}
 \begin{align}
     \sqrt{n} (\hat{\gamma} - \gamma_*) \approx~& \sqrt{n} \E_n[\Phi_\gamma] + o_p(1) &
     \Phi_\gamma =~& (\E[QQ^\top] +  \frac{\alpha}{n} I)^{-1} Q (D - Z^\top \gamma_*)
 \end{align}
 where $Q=\E[ZX^\top]\E[XX^\top]^+ X$ is the projection of $Z$ on $X$ and $\alpha/n$ is the $\ell_2$-penalty level used in the adversarial IV method.
 Using standard influence function calculus, we can derive the approximate asymptotic linear representation of $\hat{\pi}$ as:
 % \klcomment{I'm assuming this was computed similar to that of the primal estimate, ie. Lemma 36 of your notes?}\vscomment{Yes.}
\begin{align}\label{deleteafter}
     \sqrt{n}(\hat{\pi} - \pi_0) &\approx  \sqrt{n}\E_n[\Phi_\pi] + o_p(1) \\
     \Phi_\pi &= \frac{V (D - \pi_0 V)}{\E[V^2]}  + J_0^\top \Phi_\gamma \\
     J_0 &= \E[Z (2 \pi_0 V - D)]
\end{align}
 leading to variance of the estimate $\hat{\pi}$ that is approximately:
 \begin{align}
     \sigma_\pi^2 = \E[\Phi_{\pi}^2] / n
 \end{align}
 Thus, given a dataset of $n$ samples, we can empirically estimate this variance as
 $$\hat{\sigma}_\pi^2 = \E_n[\hat{\Phi}_\pi^2] / n$$
 where 
 \begin{align*}
     \hat{\Phi}_{\pi} =~& \frac{\hat{V}(D - \hat{\pi}\hat{V})}{\E_n[\hat{V}^2]} + \hat{J}^\top \hat{\Phi}_\gamma &
     \hat{J} =~& \E_n[Z (2\hat{\pi} \hat{V} - D)]\\
     \hat{\Phi}_\gamma =~& (\E_n[\hat{Q}\hat{Q}^\top] + \frac{\alpha}{n} I)^{-1} \E[\hat{Q} (D - Z^\top \hat{\gamma})] &
     \hat{Q} =~& \E_n[ZX^\top]\E_n[XX^\top]^+ X
 \end{align*}

 \paragraph{(b) z-test for $\E[{D}V] \geq \epsilon$} An alternative to the $F$-test is to simply test that the denominator $\E[D V]$ in Equation~\eqref{eq:primalIV-app} is sufficiently far away from zero. Let $\pi_0 = \E[D(D - Z^\top {\gamma_*})]$ and $\hat{\pi}=\E_n[D (D - Z^\top\hat{\gamma})]$ and using similar reasoning as in the test in the previous paragraph, we can derive an approximate asymptotic representation of the estimate as:
 \begin{align*}
     \sqrt{n}(\hat{\pi} - \pi_0) \approx~& \sqrt{n}\E_n[\Phi_\pi] + o_p(1) &
      \Phi_\pi =~& D (D - Z^\top \gamma_*) - \pi_0 + J_0 \Phi_\gamma &
      J_0 = - \E[DZ^\top]
 \end{align*}
Hence, we have that:
 \begin{align}
     \sqrt{n} (\hat{\pi} - \pi_0) \to_d N(0, \tilde{\sigma}_{\pi}^2)
 \end{align}
 with $\tilde{\sigma}_\pi^2 = \E[\Phi_\pi^2]$.
 
 Moreover, an estimate of the variance $\tilde{\sigma}_\pi^2$ can be calculated as:
 \begin{align}
     \hat{\tilde{\sigma}}_\pi^2 =~& \E_n[\hat{\Phi}_\pi^2] &
     \hat{\Phi}_\pi =~& D (D - Z^\top \hat{\gamma}) - \hat{\pi} + \hat{J} \hat{\Phi}_\gamma &
     \hat{J} =~& - \E_n[DZ^\top]
 \end{align}
 Hence, we have that $\sqrt{n} \, \hat{\pi}$ is distributed approximately as $N(\pi_0, \hat{\tilde{\sigma}}_\pi^2)$. We can therefore use the test statistic 
 \begin{align}
     \hat{T}=\sqrt{n} \,|\hat{\pi}| = \sqrt{n} \,|\E_n[D (D - Z^\top\hat{\gamma})]| \tag{Z-test statistic}
 \end{align}
 We can reject the null hypothesis of a weak instrument, i.e., that $\pi_0$ is smaller than some constant $\epsilon$, if $\hat{T}$ is larger than the $1-\alpha$ percentile of a folded normal distribution centered at $\epsilon$ and with scale $\hat{\tilde{\sigma}}_\pi$.

\subsubappendix{Alternative Neyman Orthogonal Testing Method}

To avoid the heuristic argument used in the asymptotically linear representation of the regularized estimate $\hat{\gamma}$, we can instead consider a Neyman orthogonal variant of the two tests and their corresponding statistics. The empirical results presented in the paper use the heuristic argument of the previous section. Our code allows the user to specify which testing method variant they want to use, with the default being the method presented in this section.

\paragraph{(a) $V$ strength $F$-test} We can construct a Neyman orthogonal estimate of the parameter $\pi_0 = \frac{\E[D V]}{\E[D^2]}$, as follows. Consider the moment estimate that defines $\pi_0$:
\begin{align}
    \E[(D - Z^\top \gamma_*)\, (D - \pi_0 (D - Z^\top \gamma_*)] = 0
\end{align}
The gradient of the above equation with respect to $\gamma_*$ is $\E[J_0 Z]$ where $J_0 = 2 \pi_0 Z^\top \gamma_* - D$.
Suppose that we identify a parameter $\zeta_*$ that satisfies the linear IV equation with instrument $Z$, treatment $X$ and outcome $J_0$:
\begin{align*}
    \E[(J_0 - X^\top\zeta) Z] = 0
\end{align*}
Assuming that this moment equation admits a solution, the minimum norm solution $\zeta_*$ can be estimated via an $\ell_2$-regularized adversarial IV method, in a manner similar to estimating $\gamma_*$. Then consider the debiased moment equation:
\begin{align}
\E[(D - Z^\top \gamma_*)\, (D - \pi_0 (D - Z^\top \gamma_*))] + \zeta_*^\top \E[X (D - Z^\top \gamma_*)] = 0 
\end{align}
This moment equation is Neyman orthogonal with respect to $\zeta_*$ since $\E[X (D - Z^\top \gamma_*)]$ is the gradient of the debiased moment with respect to $\zeta_*$ and by the definition of $\gamma_*$, we have that $\E[X (D - Z^\top \gamma_*)]=0$. Moreover, the gradient of the debiased moment with respect to $\gamma_*$ is:
\begin{align*}
    \E[J_0 Z] - \E[Z X^\top]\zeta_* = \E[(J_0 - X^\top \zeta_*) Z]
\end{align*}
which is also zero by the definition of $\zeta_*$. 

Thus we can construct a Neyman orthogonal moment estimate for the parameter $\pi_0$ as the solution to:
\begin{align}
    \E_n[(D - Z^\top \hat{\gamma}) (D - \pi (D - Z^\top \hat{\gamma})) + \hat{\zeta}^\top X (D - Z^\top \hat{\gamma})] = 0
\end{align}
leading to the estimate:
\begin{align}
    \hat{\pi} = \frac{\E_n[(D - Z^\top \hat{\gamma}) ( D + X^\top \hat{\zeta})]}{\E_n[(D - Z^\top \hat{\gamma})^2]}
\end{align}
Due to Neyman orthogonality and since the $\ell_2$-regularized IV estimates converge at $o(n^{1/4})$ rates to the minimum norm solutions (as a special case of the results in [\citen{bennett2022inference}] for the case of linear function spaces), the results of [\citen{chernozhukov2017double}] imply that $\hat{\pi}$ is asymptotically linear:
$$\sqrt{n}(\hat{\pi} - \pi_0) = \sqrt{n} \E_n[\Phi_\pi] + o_p(1)$$
\begin{align*}
    \Phi_\pi 
    = \frac{(D - Z^\top {\gamma_*}) (D - \pi_0 (D - Z^\top {\gamma_*})) + {\zeta_*}^\top X (D - Z^\top {\gamma_*}) }
    {\E[(D - Z^\top \gamma_*)^2]}
\end{align*}
Thus, we can derive the asymptotic normality statement:
\begin{align}
    \sqrt{n} (\hat{\pi} - \pi_0) \to_d N(0, \tilde{\sigma}^2_{\pi})
\end{align}
And we can derive an estimate of the variance $\sigma^2_{\pi} = \tilde{\sigma}^2_{\pi} / n$ of the Neyman orthogonal estimate $\hat{\pi}$ as:
\begin{align}
    \hat{\sigma}_\pi^2 =~& \E_n[\hat{\Phi}_\pi^2]/n &
    \hat{\Phi}_\pi =~& \frac{(D - Z^\top \hat{\gamma}) (D - \hat{\pi} (D - Z^\top \hat{\gamma})) + \hat{\zeta}^\top X (D - Z^\top \hat{\gamma}) }{\E_n[(D - Z^\top \hat{\gamma})^2]}
\end{align}
The rest of the $F$-test is the same as described in Appendix \ref{app:weakiv_tests_method}.

\paragraph{(b) $z$-test for $\E[DV]> \epsilon$} We can construct a Neyman orthogonal estimate of the parameter $\pi_0 = \E[D V]$ in a similar way to the $F$-test. Consider the moment estimate that defines $\pi_0$:
\begin{align}
    \E[(D - Z^\top \gamma_*)\, D] - \pi_0 = 0
\end{align}
The gradient with respect to $\gamma_*$ is $\E[J_0 Z]$ where $J_0 = - D$.

Suppose that we identify a parameter $\zeta_*$ that satisfies the linear IV equation with instrument $Z$, treatment $X$ and outcome $J_0$:
\begin{align*}
    \E[(J_0 - X^\top\zeta) Z] =& 0
\end{align*}
Assuming that this moment equation admits a solution\footnote{Existence of a solution $\zeta_*$ is equivalent to the assumption that $\cov(Z, D)$ lies in the range of $\cov(Z, X)$. Note that our main assumption on the existence of a solution to the dual IV problem is equivalent to the assumption that $\cov(X, D)$ lies in the range of $\cov(X, Z)$. The two assumptions are slightly different but similar in nature. In Appendix~\ref{app:dual-existence}, we show existence of a solution $\gamma_*$ to the dual IV problem is implied by the property that $\cov(M,Z)$ has full row rank, and we see a similar pattern follows for $\zeta_*$. For instance, suppose that $\E[Z\mid M, D]$ is linear and independent of $D$, e.g., $\E[Z\mid M, D] = A M$ for some matrix $A$. (Allowing a dependence on $D$ in this argument can also be handled, albeit the exposition becomes more complex and the sufficient assumptions harder to interpret.) Then $\cov(Z, D) = \E[ZD]=A\E[MD]$. Similarly, we have that $\cov(Z, X) = \E[Z X^\top] = \E[\E[Z\mid M] X^\top] = A \E[MX^\top]$. Thus existence of a solution is equivalent to existence of a $\zeta_*$ such that $\cov(Z, X) \zeta_* = \cov(Z, D)$, or equivalently, $A (\E[MD] - \E[MX^\top] \zeta_*) =0$. Thus, existence of a $\zeta_*$ is implied by existence of a solution to the linear system $\E[MX^\top] \zeta_*=\E[MD]$, which is implied by the assumption that $\E[MX^\top]=\cov(M, X)$ has full row rank. Thus the assumptions needed for the existence of $\eta_*$ are very similar to the prerequisites for the dual parameter $\gamma_*$.}, the minimum norm solution $\zeta_*$ can be estimated via an $\ell_2$-regularized adversarial IV method. Then we can consider the debiased moment equation:
\begin{align*}
\E[(D - Z^\top \gamma_*)\, D - \pi_0] + \zeta_*^\top \E[X (D - Z^\top \gamma_*)] = 0
\end{align*}
This moment equation is Neyman orthogonal with respect to $\zeta_*$ since $\E[X (D - Z^\top \gamma_*)]$ is the gradient of the debiased moment with respect to $\zeta_*$ and by the definition of $\gamma_*$, we have that $\E[X (D - Z^\top \gamma_*)]=0$. Moreover, the gradient of the debiased moment with respect to $\gamma_*$ is:
\begin{align*}
    \E[J_0 Z] - \E[Z X^\top]\zeta_* = \E[(J_0 - X^\top \zeta_*) Z]
\end{align*}
which is also zero by the definition of $\zeta_*$. Thus we can construct a Neyman orthogonal moment estimate for the parameter $\pi_0$ as the solution to:
\begin{align*}
    \E_n[(D - Z^\top \hat{\gamma}) D - \pi + \hat{\zeta}^\top X (D - Z^\top \hat{\gamma})] = 0
\end{align*}
leading to the estimate:
\begin{align}
    \hat{\pi} = \E_n[(D - Z^\top \hat{\gamma}) ( D + X^\top \hat{\zeta})]
\end{align}
Due to Neyman orthogonality and since the $\ell_2$-regularized IV estimates converge at $o(n^{1/4})$ rates to the minimum norm solutions, the results of [\citen{chernozhukov2017double}] imply that $\hat{\pi}$ is asymptotically linear:
 $$\sqrt{n}(\hat{\pi} - \pi_0) = \sqrt{n} \E_n[\Phi_\pi] + o_p(1)$$
\begin{align}
    \Phi_\pi = (D - Z^\top {\gamma_*}) D - \pi + {\zeta_*}^\top X (D - Z^\top {\gamma_*})
\end{align}
Thus, we can derive the asymptotic normality statement:
\begin{align}
    \sqrt{n} (\hat{\pi} - \pi_0) \to_d N(0, \tilde{\sigma}^2_{\pi})
\end{align}
where the estimate of the asymptotic variance $\tilde{\sigma}^2_{\pi}=\E[\Phi_\pi^2]$ of the Neyman orthogonal $\hat{\pi}$ is
\begin{align}
    \hat{\tilde{\sigma}}_\pi^2 =~& \E_n[\hat{\Phi}_\pi^2] &
    \hat{\Phi}_\pi =~& (D - Z^\top \hat{\gamma}) D - \hat{\pi} + \hat{\zeta}^\top X (D - Z^\top \hat{\gamma})
\end{align}
Thus we have that the new test statistic:
\begin{align}
     \hat{T}=\sqrt{n} \,|\hat{\pi}| = \sqrt{n} \,|\E_n[(D - Z^\top \hat{\gamma}) ( D + X^\top \hat{\zeta})]|
 \end{align}
approximately follows a folded normal distribution with center $\pi_0$ and scale $\hat{\tilde{\sigma}}_\pi$. To test for a weak instrument, we proceed as before and reject the hypothesis that $\pi_0 \leq \epsilon$ by checking if the $z-$statistic is larger than the $1-\alpha$ percentile of a folded normal distribution with center $\epsilon$ and scale $\hat{\tilde{\sigma}}_\pi$.

\subappendix{Proxy covariance rank test}\label{apendix:test:rank}

\subsubappendix{Intuition}
We want to check if the health proxies $X$ and $Z$ are sufficiently related through the mediator $M$ by estimating the rank of the covariance matrix $\cov(X,Z)$. To do this, 
recall from Lemma~\ref{lem:plr} that the existence of a partially linear bridge function $q$ relies on a partially conditional model $\E[X \mid M] = FM$, with $F$ having full column rank (and recall all variables correspond to residuals after removing $W$). 

Then the covariance of $X$ and $Z$ can be written as:
\begin{align*}
    \cov(X, Z) = \E[XZ^\top] = \E[\E[X\mid M, Z] Z^\top] = \E[\E[X\mid M] Z^\top] = F\E[M Z^\top ] = F\, \cov(M, Z)
\end{align*}
If $\cov(M, Z)$ has full row rank, we show in Appendix~\ref{app:dual-existence} that this is an interpretable sufficient condition for the existence of a solution to the dual IV. Moreover, we also show that the full row rank assumption for $\cov(M,Z)$ is necessary if we want to use the same proxies $Z$ for any possible treatment $D$. Under this assumption, because $\cov(X, Z)$ is the product of two matrices both of rank equal to the dimension of $M$, then the rank of $\cov(X, Z)$ is also the dimension of $M$ (see e.g. \cite[Appendix A.1]{strang2022introduction}). 
Thus under this sufficient (and essentially necessary) condition, the rank of $\cov(X, Z)$ should reveal the dimension of $M$.

In practice, we do not know the true dimension of $M$. Therefore, by identifying the rank of $\cov(X, Z)$, we are uncovering the dimension of $M$ that such a pair of proxies can implicitly control for. If we believe that the hidden mediator has more dimensions than the identified rank of $\cov(X, Z)$, then we should be concerned that this pair of proxies cannot control for all the mediators we are interested in.

\subsubappendix{Method}
We calculate the rank of the covariance matrix by calculating the empirical covariance matrix $\E_n[XZ^\top]$ and estimating a threshold $\sigma_*$ above which we determine the empirical matrix's singular values as statistically significant, and thus we report as the rank to be the number of such statistically significant singular values.

To compute the threshold singular value $\sigma_*$, we use a conservative test based on the error of the singular values between the sample (empirical) covariance matrix  $\E_n[{X}{Z}^\top ]$ and the true population covariance matrix $\E[{X}{Z}^\top ]$. We leverage Weyl's theorem \cite{weyl1912asymptotische} and a high probability bound on the Frobenius norm of the error between these two matrices using a characterization of the asymptotic distribution of the norm as a weighted sum of $\chi^2$ random variables.

In particular, Weyl's inequality for singular values, states that $\forall i \leq \min\{d_X, d_Z\}$
\begin{align}
    |\sigma_i(\E_n[XZ^\top]) - \sigma_i(\E[XZ^\top])| \leq \|\E_n[XZ^\top] - \E[XZ^\top]\|_{op} \leq \|\E_n[XZ^\top] - \E[XZ^\top]\|_{F}
\end{align}
where $\norm{A}_F = \sqrt{\sum_{i,j} A_{i,j}^2}$ is the Frobenius norm, $\norm{A}_{op}$ is the operator norm, and $\sigma_i(A)$ denotes the $i$-th largest singular values of $A$. Thus we can control the error between the singular values of the empirical covariance matrix and the population covariance matrix, uniformly over all singular values, by constructing a high-probability bound on the Frobenius norm of the error $\E_n[XZ^\top] - \E[XZ^\top]$.

Let the Frobenius norm above be rewritten as 
$$\|\E_n[XZ^\top] - \E[XZ^\top]\|_{F} = \sqrt{\sum_{i\leq d_X,j\leq d_Z} \vec f_{ij}^{\,2}}$$ where 
$\vec f$ is a $p=d_X \cdot d_Z$ dimensional vector and $\vec f_{ij} = \E_n[X_{i}Z_{j}] - \E[X_i Z_{j}]$. Note, by the multivariate CLT, the vector $\vec f$ approximately follows a multivariate normal distribution, i.e., $\vec f \to_d N(0,V)$, with covariance $V_{ij, kl}=\cov(X_i Z_j, X_k Z_l) / n$ . We can approximate $V$ with an empirical covariance $\hat{V}$ where $\hat{V}_{ij,kl}=\cov_n(X_i Z_j, X_k Z_l) / n$. Then, we have $\hat{V}^{-1/2} {\vec f} \to_d N(0, I)$. Therefore, the sum of the squares $\sum_{i,j} \vec f_{ij}^{\,2}$ asymptotically follows the weighted sum of $p$ independent $\chi_2(1)$ distributions where the weights are the eigenvalues of $V$ (see e.g. [\citen{imhof1961computing}]), which can be approximated by the eigenvalues of $\hat{V}$.
% \vscomment{I don't understand the last equality you write: whenever I have a mean-zero random vector $W$ and I get $n$ i.i.d. samples of the random vector $W^{(1)},\ldots, W^{(n)}$ and I take $S=\frac{1}{n} \sum_i W^{(i)}$, then by the multivariate CLT, we have that $\sqrt{n} S \to N(0, \tilde{V})$, with $\tilde{V}_{i,j} = \cov(X_i, X_j)$. Thus the covariance of $S$ is approximately $\tilde{V}/n$. Apply this reasoning with $W$ being a $p_X\cdot p_Z$ dimensional vector with $W_{ij} = X_i Z_j - \E[X_i Z_j]$.}

We then calculate the $1 - \alpha$ percentile of this distribution via Monte Carlo simulation where we draw many samples of a $p$-dimensional vector with independent $\chi_2(1)$ entries, calculate the weighted sum for each sample using the eigenvalues of $\hat{V}$, and taking the $1-\alpha$ percentile of this weighted sum across samples. Then the critical value is the square root of that percentile, i.e., $\sqrt{\sum_{ij} \vec f_{ij}^{\, 2}}$. Asymptotically, the Frobenius norm of the error $\E_n[XZ^\top] - \E[XZ^\top]$ will not be larger than this critical value with probability more than $\alpha$. Therefore, the error $|\sigma_i(\E_n[XZ^\top]) - \sigma_i(\E[XZ^\top])|$ for any of the singular values will not be larger than the critical value with probability more than $\alpha$ asymptotically 
 as $n\to \infty$. Thus, this $1-\alpha$ percentile we calculated will be our threshold $\sigma_*$. Any singular value which is below $\sigma_*$ we declare as insignificant. The singular values above $\sigma_*$ we can confidently claim are non-zero (albeit in a conservative manner, given that Weyl's inequality is only an upper bound). The number of such values is a lower bound on the rank of $\cov(X,Z)$ (and thus $M$) that we can be confident about.

\subappendix{Weak instrument confidence interval}\label{appendix:weakiv}
Let $\hat{V}= D - \hat{\gamma}^\top Z$, $V=D-\gamma_*^\top Z$, $\hat{\bar{Y}} = Y - \hat{h}^\top  X$, and $\bar{Y} = Y - h_*^\top X$, where $\gamma_*, h_*$ are the true parameters and $\hat{\gamma}, \hat{h}$ are the estimates (i.e., minimum norm solutions). Consider the implicit bias effect estimate $\hat{\theta}=\E_n[\hat{V} \bar{Y}] / \E_n[\hat{V} D]$ of $\theta_0=\E[{V} \bar{Y}] / \E[{V} D]$. Under a weak instrument regime, $\E[VD]$ and thus $\E_n[\hat{V} D]$ are very close to zero, leading to highly biased estimates and poor coverage of the confidence intervals that are based on asymptotic normality. In particular, the intervals use the asymptotic approximation that:
\begin{align*}
    \sqrt{n} (\hat{\theta} - \theta_0) = \sqrt{n}\left(\frac{\E_n[\hat{V} \hat{\bar{Y}}]}{\E_n[\hat{V} D]} - \theta_0\right) = \sqrt{n}\frac{\E_n[\hat{V} (\hat{\bar{Y}} - D\theta_0)]}{\E_n[\hat{V} D]} \approx \sqrt{n}\frac{\E_n[\hat{V} (\hat{\bar{Y}} - D\theta_0)]}{\E[\hat{V} D]} + o_p(1)
\end{align*}
The last step is a good approximation only if the denominator $\E[V D]$ (and subsequently also $\E[\hat{V} D]$) is sufficiently separated from zero by a gap that is of magnitude larger than the $O(n^{-1/2})$ error we would expect that the empirical variant $\E_n[\hat{V} D]$ would have, as compared to its population analogue.

The test provided in Chapter 13.3 of [\citen{chernozhukov2024applied}] computes a confidence interval that does not invoke such an approximation argument for the denominator. Consider the null hypothesis that the true parameter $\theta_0$ takes some value $\theta$ and since by construction the identifying moment equation is Neyman orthogonal with respect to all nuisance functions, we have from the results of [\citen{chernozhukov2017double}] that:
\begin{align}
    \sqrt{\frac{n}{\text{Var}((\bar{Y} - D\theta) V)}} \E_n[(\hat{\bar{Y}} - D\theta)\hat{V}] \to_d N(0, 1) 
\end{align}
and that the same also holds for the empirical estimate of the variance:
\begin{align}
    \sqrt{\frac{n}{\text{Var}_n((\hat{\bar{Y}} - D\theta) \hat{V})}} \E_n[(\hat{\bar{Y}} - D\theta)\hat{V}] \to_d N(0, 1) 
\end{align}
Thus the square of the left-hand-side is distributed asymptotically as a $\chi^2(1)$ distribution:
\begin{align}
    C(\theta) \defeq \frac{n\, \E_n[(\hat{\bar{Y}} - D\theta)\hat{V}]^2}{\text{Var}_n((\hat{\bar{Y}} - D\theta) \hat{V})} \to_d \chi^2(1)
\end{align}
% \klcomment{I'm confused what $C(\theta_0)$ vs. $C(\theta)$ mean here, similar to above} \vscomment{$\theta_0$ is the true parameter. $\theta$ is any candidate parameter we are currently testing whether it is equal to $\theta_0$. For instance, testing whether $\theta_0=?0$, does not mean that $\theta_0=0$. Simlarly if you want to test any other value $\theta$ other than zero.}

Thus, we have that, for the true parameter, $C(\theta_0)\to_d \chi^2(1)$ and therefore with probability $1-\alpha$, $C(\theta_0)$ should be less than the $(1-\alpha)$ percentile of the $\chi^2(1)$ distribution, denoted $q_{\chi^2(1)}(1-\alpha)$. Thus we can construct an $1-\alpha$ confidence region (CR) for $\theta_0$ as the set of all parameters $\theta$ for which $C(\theta)\leq q_{\chi^2(1)}(1-\alpha)$:
\begin{align}
    \text{CR} = \{\theta\in \mathbb{R}: C(\theta) \leq q_{\chi^2(1)}(1-\alpha)\}
\end{align}

We operationalize this approach by searching over a grid candidate $\theta$ parameters in some pre-determined sufficiently large range of values and accepting a candidate parameter only if $C(\theta)\leq q_{\chi^2(1)}(1-\alpha)$. In the end we return the confidence interval with endpoints that correspond to the smallest and largest accepted values. In our setting, the outcome is typically binary and hence it suffices to consider a grid of $[-1, 1]$.

\subappendix{Influence score analysis}\label{appendix:inf}

 \subsubappendix{Calculating influence scores}
 An influence score is a continuous value $v_i$ assigned to each data point $i$ that denotes how influential that point was in calculating the estimate (e.g., $\hat{\theta}$). Intuitively, the influence score $v_i$ approximates how much the estimate $\theta$ would change if we remove sample $i$ from the training data, i.e. if we denote with $\hat{\theta}_{-i}$ the estimate we would have obtained if we removed the $i$-th sample from the data, then $v_i \approx \hat{\theta} - \hat{\theta}_{-i}$. Influence scores are often used to assess robustness of an effect estimate, where a robust estimate should not drastically change if a small number of highly-influential points are removed\cite{alaa2019validating,cause:finite}.
    
Since our method is based on the empirical analogue of the Neyman orthogonal moment equation:
\begin{align}
    \E[(Y - X^\top h_* - D\theta)\, (D - Z^\top \gamma_*)] = 0
\end{align}
the estimate $\hat{\theta}$ admits an asymptotic linear representation:
\begin{align}
    \sqrt{n} (\hat{\theta} - \theta_0) = \frac{1}{\sqrt{n}} \sum_{i=1}^n \frac{(Y_i - X_i^\top h_* - D_i \theta_0)\,(D_i - Z_i^\top \gamma_*)}{\E[D\,(D - Z^\top \gamma_*)]} + o_p(1)
\end{align}
Thus the influence of each data point can be well approximated (as $n$ goes to infinity) by:
\begin{align}
    \hat{\theta} - \hat{\theta}_{-i} \approx \frac{1}{n} \frac{(Y_i - X_i^\top h_* - D_i \theta_0)\,(D_i - Z_i^\top \gamma_*)}{\E[D\,(D - Z^\top \gamma_*)]}
\end{align}
Thus we offer an approximation of the influence $\hat{\theta} - \hat{\theta}_{-i}$ of each data point by the score
\begin{align}
    v_i \defeq \frac{1}{n} \frac{(Y_i - X_i^\top \hat{h} - D_i \hat{\theta})\,(D_i - Z_i^\top \hat{\gamma})}{\E_n[D\,(D - Z^\top \hat{\gamma})]}
\end{align}

% \vscomment{We want to say that we will use the RHS as the approximate influence $v_i$. What you wrote before didn't have this meaning.}

Intuitively, if $v_i \approx 0$, then point $i$ has negligible effect on the implicit bias estimate. A large positive influence score $v_i$ means patient $i$ pushes the final estimate of $\theta$ to be more positive, and thus removing patient $i$ from the data would result in a decrease of the final estimate by $\approx v_i$. Similarly, a very negative value of $v_i$ pushes $\theta$ to be more negative, and removing patient $i$ would increase the final estimate by $\approx |v_i|$.  

In our analysis we also calculate a more accurate variant of the influence. In particular, fixing all nuisance estimates, our final estimate $\hat{\theta}$ is the result of a 2SLS algorithm with instrument $D - Z^\top\hat{\gamma}$, treatment $D$ and outcome $Y - X^\top \hat{h}$. Exact leave-one-out influence scores have been derived for this algorithm by \cite[pp. 266-68]{belsley2005regression}. Thus, ignoring the influence that removing a sample has on the nuisance estimates, the leave-one-out influence can be computed exactly via the results of [\citen{belsley2005regression}]. Our implementation uses these exact influence scores as a default. Moreover, analogues of several robust statistic diagnostics have been developed for 2SLS regression, such as Cook's distance, leverage scores, hat-values, studentized residuals and dffits (see [\citen{diag2sls}]). Our implementation provides all these quantities for the final stage 2SLS regression.

% \klcomment{Just confirming: the default is different than $v_i$ as above (the sentence makes it sound like this)?}\vscomment{Yes; the default is the exact leave-one-out influence calculation for 2SLS.}
\subsubappendix{Estimating an influential set}
Similar to [\citen{cause:finite}], we can use influence scores to create an influence set whose removal should maximally alter the estimate $\hat{\theta}$ in some pre-specified direction. For instance, we can return the minimal set of patients such that their removal from the dataset will lead to a new estimate $\hat{\theta}_{-\text{inf}}$ and associated $(1-\alpha)$-confidence interval $[\text{lb}_{-\text{inf}}, \text{ub}_{-\text{inf}}]$\footnote{'lb'=lower bound; 'ub'=upper bound} that will contain zero, implying that no statistically significant implicit bias can be detected for the remaining patients that are not in the minimal set.
% \kledit{kledit to myself: just occurred to me: it probably matters if the intersection of people with attribute D and Y are very small, which is i think the case and will create a small inf set size} 

We next describe how to calculate the minimal influence set.\footnote{Given that the influence values are only approximations, after finding an influential set we remove the set and re-run estimation to confirm that the resulting confidence interval contains zero.} Without loss of generality, suppose that $\hat{\theta}>0$ and the confidence interval $[\text{lb}, \text{ub}]$ using all the data (calculated either via the asymptotic normal approximation or via the weak-identification-robust confidence interval procedure in Appendix~\ref{appendix:weakiv}) does not contain zero, i.e. $\text{lb}>0$. First, we sort all $n$ patients based on their influence scores such that $[v_0, v_1, \ldots, v_n]$ orders the patients from highest (most positive) to lowest (most negative) influence score. We only consider points whose scores $v_i\geq 0$, as removing them will push the estimate to be more negative. Suppose the first $k$ sorted influence scores are non-negative. We then find the index $m$ such that $\sum_{i=0}^m v_i > \text{lb}$. Then our candidate minimal influence set are patients $S = \{1,2,\ldots, m\}$, and we hypothesize that re-estimating $\hat{\theta}_{-\text{inf}}$ after removing all points in $S$ will lead to a new confidence interval $[\text{lb}_{-\text{inf}}, \text{ub}_{-\text{inf}}]$ that will contain zero.

\subsubappendix{Interpreting high-influence sets with feature interpretability}\label{appendix:infint}
To investigate what phenotypes characterize these identified high-influence patients, we perform a simple feature interpretability analysis. For each feature in $W, Z, X$, we compare the high-influence set with the rest of the patients not in the high-influence set. For categorical (including binary) features, we use a $\chi^2$ test. For continuous features, we use a Kolmogorov–Smirnov test. We analyze all statistically significant differences between high-influence patients versus the rest of the data. 

\appendix{Proof of main theoretical claims}

\subappendix{Causal assumptions of Figure~\ref{fig:cg}}\label{appendix:cg_ass}
The following are the assumptions encoded in the structural causal model of Figure ~\ref{fig:cg} and are necessary for identifying the controlled direct effect without observing the mediator $M$:
\begin{enumerate}
    \item Conditional ignorability:  $Y(d,m)\ci \{D, M(d)\}\mid W$
    \item Validity of outcome proxy $X$: $X\ci D \mid (M, W)$
    \item Validity of treatment proxy $Z$: $Z \ci (Y, X) \mid (D, M, W)$
    \item Consistency: $\E[Y(d,m) \mid M(d) = m, D=d] = \E[Y(d,m) \mid M = m, D=d]$
\end{enumerate}

\subappendix{Proof of the controlled direct effect as a g-formula}\label{appendix:cdegformula}

Recall the controlled direct effect 
\begin{align*}
\theta =~& \int_{m,w} \E[Y(1, m) - Y(0, m)\mid W=w] \text{ }p(m, w) \text{ }dm\text{ } dw
\end{align*}

Here, we show that if $M$ is known, we can identify $\theta$ using a g-formula. \footnote{In causal inference, a 'g-formula' estimates a causal effect by adjusting for confounding variables.}
\begin{lemma}\label{lemma:gformula}
    The controlled direct effect of Figure \ref{fig:cg} can be calculated by the g-formula
    \begin{align*}
    \theta = \E[\E[Y\mid D=1, M, W] - \E[Y\mid D=0, M, W]]
\end{align*}
\end{lemma}
\begin{proof}
    We assume the causal constraints in \ref{appendix:cg_ass}, and denote $M(d)$ as the potential value (also known as the counterfactual or intervened value) of the mediator $M$ if we intervene and fix $D=d$. Given this,  
    \begin{align*}
    \E[Y(d, m)\mid W=w] =&  \E[Y(d, m)\mid D=d, M(d)=m, W=w] \\  =& \E[Y(d, m)\mid D=d, M=m, W=w] \\=& \E[Y(D, M)\mid D=d, M=m, W=w]\\
=~& \E[Y\mid D=d, M=m, W=w] 
    \end{align*}
Then the controlled direct effect can be identified as:
    \begin{align*}
        \theta =~& \int_{m,w} \E[Y(1, m) - Y(0, m)\mid W=w] \text{ }p(m, w) \text{ }dm\text{ } dw\\
        =~& \int_{m,w} (\E[Y\mid D=1, M=m, W=w] - \E[Y\mid D=0, M=m, W=w]) \text{ }p(m, w) \text{ }dm\text{ } dw\\
        =~& \E[\E[Y\mid D=1, M, W] - \E[Y\mid D=0, M, W]]
    \end{align*}
\end{proof}

\subappendix{Proof of Theorem~\ref{thm:id}}\label{appendix:thm:id}
\subsubappendix{Intuition}
Theorem~\ref{thm:id} identifies how we could estimate the controlled direct $\theta$ when $M$ isn't observed. To account for $M$, we leverage the information of the proxy variables $X,Z$ via a ``bridge function'' $q$. Intuitively, if we can find a $q(D,X,W)$ that approximates the conditional expectation of $Y$ (given $D,M,W$), then we show in our proof below that $q$ also satisfies another equation:
\begin{align}
    \E[Y\mid D, M, W] = \E[q(D, X, W)\mid D, M, W] 
    \\ \implies  \E[Y - q(D, X, W) \mid D, Z, W] = 0 
\end{align}
Importantly, this latter equation is in the form of a Non-Parametric Instrumental Variable (NPIV) problem with confounders $W$, instruments $(Z;D)$, and treatment $(X;D)$. 
Following these properties of $q$, we will see in the below proof that in fact the controlled direct effect can be written as $\theta_0= \E[q(1, X, W) - q(0, X, W)]$.

\subsubappendix{Proofs}
\begin{lemma} An outcome bridge function $q$ which satisfies $ \E[Y\mid D, M, W] = \E[q(D, X, W)\mid D, M, W]$ also satisfies $\E[Y - q(D, X, W) \mid D, Z, W] = 0 $.
    \end{lemma}
\begin{proof}
First note $ \E[Y\mid D, M, W] = \E[q(D, X, W)\mid D, M, W] \leftrightarrow  \E[Y - q(D, X, W) \mid D, M, W] = 0$.
We use the criteria of Figure \ref{fig:cg} (see also \ref{appendix:cg_ass}) that $Z\ci (Y, X) \mid (D, M, W)$ for the following: 
\begin{align*}
    \E[Y - q(D, X, W) \mid D, Z, W] =~& \E[\E[Y - q(D, X, W)\mid D, Z, M, W]\mid D, Z, W] \\
    =~& \E[\E[Y - q(D, X, W)\mid D, M, W]\mid D, Z, W] = 0
\end{align*}
Thus the outcome bridge function is also a solution to the \emph{feasible} non-parametric Instrumental Variable (NPIV) regression problem in Equation~\eqref{eqn:npiv}.
\end{proof}

\begin{lemma}If we identify an outcome bridge function $q$ such that $ \E[Y\mid D, M, W] = \E[q(D, X, W)\mid D, M, W]$, then $ \theta = \E[q(1, X, W) - q(0, X, W)]$.
\end{lemma}
\begin{proof}
    
We see that 
\begin{align*}
    \E[\E[Y\mid D=d, M, W]] =~& \E[\E[q(D, X, W)\mid D=d, M, W]] \\
    =& \E[\E[q(d, X, W)\mid D=d, M, W]]\\
    =~& \E[\E[q(d, X, W)\mid M, W]] \tag{$X \ci D\mid M, W$}\\
    =~& \E[q(d, X, W)]
\end{align*}
Also recall from Lemma \ref{lemma:gformula} that $ \theta = \E[\E[Y\mid D=1, M, W] - \E[Y\mid D=0, M, W]]$. Then the controlled direct effect is identified as:
\begin{align*}
    \theta=& \E[\E[Y\mid D=1, M, W] - \E[Y\mid D=0, M, W]] \\
    =&\E[q(1, X, W) - q(0, X, W)]
\end{align*}

\end{proof}

\subappendix{Proof of Lemma~\ref{lem:plr}: identification under partial linearity}\label{appendix:lemma1}

\subsubappendix{Intuition}
If we assume $ \E[Y\mid D, M, X, W]$ and $\E[X\mid M, W]$ are partially linear, and make a few other modest assumptions, then we can prove the bridge function $q$ (which we saw in \ref{appendix:thm:id} can be used to identify $\theta$) is also partially linear. We also show in the proofs below that the controlled direct effect $\theta$ is simply a coefficient in the partially linear equation of $ \E[Y\mid D, M, X, W]$.
\subsubappendix{Proofs}
\begin{lemma} (Restatement of Lemma \ref{lem:plr})
         Under the assumption of partial linearity in Equation~\eqref{eqn:plr} and the assumptions encoded by the causal graph in Figure \ref{fig:cg}, we have $q(D, X, W) = D\,c + X^\top h + f(W)$ and $h=F^+b + g$.
\end{lemma}
\begin{proof}
By the assumptions of the causal graph we have that
    \begin{align*}
        Y =~& h_Y(M, D, X, W, u_Y)\\
        X =~& h_M(M, W, u_X)
    \end{align*}
    where $h_Y, h_M$ are two arbitrary structural functions and $u_Y,u_X$ are the independent exogenous error variables of the structural causal model that are independent of $D, Z, M, W$. Moreover, by the assumption of partially linear conditional expectation functions, we can write:
    \begin{align*}
        Y =~& D\,c + M^\top b + X^\top g + f_Y(W) + \epsilon_Y, & \E[\epsilon_Y \mid D, M, X, W] = 0\\
        X =~& F\, M + f_X(W) + \epsilon_X, & \E[\epsilon_X\mid M, W] = 0
    \end{align*}
    with $\epsilon_Y = h_Y(M, D, X, W, u_Y) - \E[h_Y(M, D, X, W, u_Y)\mid D, M, X, W]$ and $\epsilon_X = h_M(M, W, u_X) - \E[h_M(M, W, u_X)\mid M, W]$. Note that $\epsilon_X \ci D \mid M, W$ since the only remnant randomness of $\epsilon_X$ is $u_X$ when we condition on $M, W$, and $u_X \ci D$.

    % \vscomment{The above explanation was not accurate. There is a crucial argument that was there in the original argument. It's only the $u_Y, u_X$ that are exogenous noise variables. The $\epsilon_X$ and $\epsilon_Y$ are regression residuals, which is different.}

    We then rearrange the equations such that we no longer rely on observing $M$. Since $F$ has full column rank, we have that the matrix $F^\top  F$ is a $p_M\times p_M$ invertible matrix and the Moore-Penrose pseudo-inverse takes the form $F^+ = (F^\top  F)^{-1} F^\top $. Thus we have
    \begin{align*}
        F^\top  X = F^\top  F\, M + F^\top  f_X(W) + F^\top  \epsilon_X \implies M = F^+ (X - f_X(W) - \epsilon_X)
    \end{align*}
    Note this expression of $M$ only contains observable variables. We can then plug in this expression of $M$ into the regression equation for $Y$ to get:
    \begin{align*}
        Y = (b^\top  F^+ + g^\top ) X + D\, c - b^\top  F^+ f_X(W) + f_Y(W) - b^\top  F^+ \epsilon_X + \epsilon_Y
    \end{align*}
    Let $h^\top  = b^\top  F^+ + g^\top $ and $f(W) = - b^\top  F^+f_X(W) + f_Y(W)$. Then we can write:
    \begin{align*}
        Y = h^\top  X + D\, c + f(W) - b^\top  F^+\epsilon_X + \epsilon_Y
    \end{align*}
    Moreover, the error terms satisfy that:
    \begin{align*}
        \E[\epsilon_Y \mid D, M, W] =~& \E[\E[\epsilon_Y\mid D, M, W, X]\mid D, M, W] = 0\\
        \E[\epsilon_X\mid D, M, W] =~& \E[\epsilon_X \mid M, W] = 0
    \end{align*}
    Thus, we can conclude that 
    \begin{align*}
        \E[Y\mid D, M, W] = \E[h^\top  X + D\, c + f_Y(W)\mid D, M, W]
    \end{align*}
    Thus, the function $q(D, X, W) = D\,c + X^\top  h + f(W)$ satisfies the property of an outcome bridge function as defined in Theorem~\ref{thm:id} and therefore is a valid bridge function. Moreover, by Theorem~\ref{thm:id}, $q(D, X, W)$ also satisfies:
    \begin{align*}
        \E[Y\mid D, Z, W] = \E[h^\top  X + D\, c + f(W)\mid D, Z, W]
    \end{align*}
\end{proof}

\begin{lemma}
     Under the assumption of partial linearity in Equation~\eqref{eqn:plr} and the assumptions encoded by the causal graph in Figure \ref{fig:cg}, we have the controlled direct effect $\theta=c$. 
\end{lemma}

\begin{proof}
    We know from Theorem \ref{thm:id} that $\theta= \E[q(1, X, W) - q(0, X, W)]$. Plugging in for $q$, we have 
\begin{align*}
    \theta =~& \E[q(1,X,W) -q(0,X,W)] \\ 
    =~& (1\cdot c + X^\top h + f(W)) - (0\cdot c + X^\top h + f(W)) \\
    =~& c
\end{align*}
\end{proof}

\paragraph{Equivalence to other definitions of controlled direct effect} Note that the partially linear restriction in Equation~\eqref{eqn:plr}, together with the causal graph assumption, implies that: $\E[Y(d, m)\mid X, W] = d\,c + m^\top b + X^\top g + f_Y(W)$. The claim follows by the following set of equalities:
\begin{align*}
    \E[Y\mid D=d, M=m, X, W] =~& \E[Y(D, M)\mid D=d, M=m, X, W]\\
    =~& \E[Y(d, m)\mid D=d, M(d)=m, X, W] \\
    =~& \E[Y(d, m)\mid X, W] \tag{$Y(d,m) \ci \{D, M(d)\}\mid X, W$}
\end{align*}
Thus we conclude that:
\begin{align*}
    \E[Y(d, m)\mid X, W] = \E[Y\mid D=d, M=m, X, W] = d\,c + m^\top b + X^\top g + f_Y(W)
\end{align*}

Therefore we obtain that 
$$\E[Y(1,m) - Y(0, m)\mid X, W]=c,$$ 
i.e. the direct effect is a constant, independent of $X, W$ and independent of the value of the mediator $m$ that we fix. Under this partially linear restriction, the controlled direct effect $\theta$ that we defined is equal to the constant $c$. Moreover, under this restriction, the constant $c$ is equivalent to other definitions of a controlled direct effect, such as:
\begin{align}
    \theta(m) \defeq \E[Y(1, m) - Y(0, m)] = c
\end{align}
for any fixed value $m$ of the mediator.

\subappendix{Proof of residualization of $W$}\label{app:res-w}
For any variable $V$, define the residual $\tilde{V} = V - \E[V\mid W]$. Under the assumption of linear structural equations, we show that we can solve the NPIV moment restriction equation from Equation \ref{eqn:npiv} after removing the effect of confounders $W$ from all variables. 

\begin{proof}
    Note that the original conditional moment restrictions imply that (by marginalizing these constraints over $X, D$) yields
\begin{align*}
    \E[Y - D\, \theta - X^\top  h \mid W] = \E[Y\mid W] - \E[D\mid W]\, \theta - \E[Y\mid W]^\top  h = 0
\end{align*}
Subtracting these constraints from the original moment constraints, we see
\begin{align*}
    \E[\tilde{Y} - \tilde{D}\,\theta - \tilde{X}^\top h\mid D, Z, W] = 0
\end{align*}
Since $\tilde{Z}, \tilde{D}$ is measurable with respect to $D, Z, W$, 
\begin{align*}
    \E[\tilde{Y} - \tilde{D}\,\theta - \tilde{X}^\top h\mid \tilde{D}, \tilde{Z}] = 0
\end{align*}
\end{proof}

\subappendix{Interpretable sufficient conditions for the existence of a dual solution}\label{app:dual-existence}

We now provide a more primitive and intuitive condition for the existence of a solution $\gamma_*$ to the dual IV problem. For simplicity of notation, we omit the tilde's on the variables and use the short-hand notation in this section as: $Y\equiv \tilde{Y}$, $X\equiv \tilde{X}$, $Z\equiv \tilde{Z}$, $D\equiv \tilde{D}$, where recall $\tilde{V} = V - \E[V | W]$.

The dual IV can be phrased as the question of the existence of a solution to the linear system:
\begin{align}
    \Sigma \gamma =~& v &
    \Sigma \defeq~& \E[XZ^\top] &
    v \defeq~& \E[X D]
\end{align}
Under the partially linear conditional expectation function assumption, we have that:
\begin{align*}
    v\defeq \E[X D] = \E[\E[X\mid M, W, D]D] = \E[\E[X\mid M, W] D] = F \E[MD] = F \cov(M, D)
\end{align*}
Similarly:
\begin{align*}
    \Sigma \defeq \E[XZ^\top]= F\E[MZ^\top] = F\cov(M, Z)
\end{align*}
Thus existence of a solution $\gamma$ to the original system is equivalent to:
\begin{align}
    F \left(\cov(M, Z) \gamma - \cov(M, D)\right) = 0
\end{align}
Since $F$ has full column rank, the latter is equivalent to:
\begin{align}
    \cov(M, Z) \gamma = \cov(M, D)
\end{align}
If $\text{Cov}(M, Z)$ has full row rank, then we have an underdetermined system\footnote{We are assuming $p_Z > p_M$.} and the above system always admits a solution, since $\text{Cov}(M, Z)$ contains $p_M$ (the dimension of $M$) linearly independent columns and hence its columns can span any $p_M$ dimensional vector. Moreover, if we want to strengthen the assumption to say that we want the proxies \kledit{$Z$} to satisfy the assumption for any correlation pattern of the treatment $D$ with the mediator, then we are asking the above system to have a solution for any vector $\cov(M, D)$, which is equivalent to the assumption that $\cov(M, Z)$ has full row rank. Thus if we want to use the same proxies $Z$ to control for the mediator for many candidate treatments $D$ (as we are in our empirical application), then such a full row-rank condition is basically both sufficient and necessary.
\begin{corollary}
    If $\cov(M, Z)$ has full row rank, then there exists a solution $\gamma_*$ to the dual IV, under the partially linear assumption in Lemma~\ref{lem:plr}.
\end{corollary}

\textbf{Example 1 } Consider the case where $M$ is a scalar (e.g. binary or continuous). In this case, we have that $\cov(M, Z)$ is a row vector and $\cov(M, D)$ is a scalar. Thus, the above holds so long as $\exists \, Z_j$ such that $\cov(M, Z_j)$ is non-zero, i.e. some variable of the treatment proxy $Z$ is correlated with the mediator. We see that re-interpreting solving the dual IV using the aforementioned linear system helps reveal what information the proxy $Z$ has to carry about the hidden mediator.

\textbf{Example 2 } Consider the case where the mediator is a categorical variable with $K+1$ categories and let $M$ be the $K$-dimensional one-hot encoding of the mediator. Then note that row $t$ of $\cov(M,Z)$ is a scalar multiple of the vector $\alpha_t\triangleq \E[Z\mid M=t]$. Thus, we need that the vectors $\alpha_t$ are linearly independent (i.e. none of them can be expressed as a linear combination of the rest). In other words, different values of the mediator cause linearly independent expected patterns in the treatment proxy vector. %\klcomment{Note to self: this example doesn't make sense to me. return to later.}

\subappendix{Proof of point identification from Theorem \ref{thm2}}\label{app:unique-id}
For simplicity of notation, we omit the tilde's on the variables $Y, X, Z, D$ and use the short-hand notation in this section as: $Y\equiv \tilde{Y}$, $X\equiv \tilde{X}$, $Z\equiv \tilde{Z}$, $D\equiv \tilde{D}$.

\subsubappendix{Intuition}
We prove that any $h_*, \gamma_*$ that satisfy the \eqref{eq:primal} and the \eqref{eq:dual}, respectively, will lead to the unique identification of the controlled direct effect $\theta_0$ when solving for the moment equation $\E[(Y - X^\top h_* - D \theta) (D - \gamma_*^\top Z)] = 0$.  This is important because it simplifies the parameter search space yet guarantees we still recover the controlled direct effect $\theta$ that we care about. 

\subsubappendix{Proofs}

Let $h_0, \theta_0$ be the true parameters that satisfy the \ref{eq:primal}
\begin{align*}
    \E\left[(Y - X^\top h - D\theta) \begin{pmatrix}Z\\D\end{pmatrix}\right] = 0
\end{align*}
Let $h_1, \theta_1$ any other solution to \ref{eq:primal}. Both of these are also solutions to the modified NPIV moment equation
\begin{align}\label{appendix:eq:npiv2}
\E[(Y - X^\top h - D \theta) (D - \gamma^\top Z)] = 0
\end{align}
for any $\gamma$, since this moment is a linear combination of the original moments. Let $\gamma_*$ be any solution to the dual IV moment, i.e., $\E[X(D-\gamma_*^\top Z)]=0$, with $\E[D(D - \gamma_*^\top Z)]\neq 0$. Then note that the modified NPIV equation is invariant to $h$, i.e., 
\begin{align*}
\E[(Y - X^\top h - D \theta) (D - \gamma_*^\top Z)] = \E[(Y - D \theta) (D - \gamma_*^\top Z)]
\end{align*}
which implies that any solution to the modified NPIV equation takes the form (since $\E[D(D - \gamma_*^\top Z)]\neq 0$):
\begin{align}
    \E[(Y - D \theta) (D - \gamma_*^\top Z)] = 0 \implies \theta = \frac{\E[Y (D - \gamma_*^\top Z)]}{\E[D(D - \gamma_*^\top Z)]}
\end{align}
which is invariant to $h$. Thus, $\theta_1 = \theta_0$.

\subappendix{Proof of Neyman orthogonality and Asymptotic Normality}\label{app:neyman}

\subsubappendix{Proofs}
\begin{proof}
    
We will show that the final moment equation:
\begin{align}
\E[(\tilde{Y} - \tilde{X}^\top h - \tilde{D} \theta) (\tilde{D} - \gamma^\top \tilde{Z})] = 0
\end{align}
is Neyman orthogonal (as defined in [\citen{chernozhukov2017double}]) with respect to nuisance parameters $h$, $\gamma$ and the nuisance functions that estimate the residual variables $\tilde{Y}, \tilde{X}, \tilde{D}, \tilde{Z}$.

\begin{itemize}
    \item \textit{Neyman orthogonality with respect to $h$:} Consider any solution to the dual equation $\gamma_*$. The orthogonality with respect to $h$ is verified since the derivative with respect to $h$ is zero:
\begin{align}
    \partial_h \E[(\tilde{Y} - \tilde{X}^\top h - \tilde{D} \theta) (\tilde{D} - \gamma_*^\top \tilde{Z})] =~& \E[\tilde{X}(\tilde{D} - \gamma_*^\top \tilde{Z})] = 0
\end{align}
where we use the fact that $\gamma_*$, by definition, must satisfy $\E[\tilde{X}(\tilde{D} - \gamma^\top \tilde{Z})] = 0$.

    \item \textit{Neyman orthogonality with respect to $\gamma$:} Consider any solution $h_*, \theta_0$ to the primal equation and modified NPIV problem. We can prove Neyman orthogonality with respect to $\gamma$ using the fact that $h_*, \theta_0$ also solve the primal equation moment restriction: 
\begin{align}
    \partial_\gamma \E[(\tilde{Y} - \tilde{X}^\top h_* - \tilde{D} \theta_0) (\tilde{D} - \gamma^\top Z)]= -\E[(\tilde{Y} - \tilde{X}^\top h_* - \tilde{D} \theta_0)\tilde{Z}^\top ] = 0
\end{align}

    \item \textit{Neyman orthogonality with respect to the residual functions:} Finally, we show Neyman orthogonality with respect to the conditional expectation functions that appear in the residuals $\tilde{Y}, \tilde{X}, \tilde{D}, \tilde{Z}$. First note, that for the correct residuals,
    \begin{align}
            \E[\tilde{V}\mid W] = \E[V - \E[V \mid W] \mid W] = 0 & & & \forall \, V \in [Y, Z, X, D]\\
        \E[\tilde{Y} - \tilde{X}^\top h - \tilde{D}\theta\mid W] = 0
        \end{align}
The directional derivative with respect to $q_Y(\cdot)=\E[Y\mid W=\cdot]$ in any direction $\nu_Y$ (which is any function that lies in the space of differences of $q_Y$) is:
\begin{align*}
    \partial_{q_Y}[\nu_Y] \defeq~& \partial_{\tau} \E[(Y - q_Y(W) - \tau \nu_Y(W) - \tilde{X}^\top h_* - \tilde{D} \theta_0) (\tilde{D} - \gamma_*^\top Z)]\mid_{\tau=0}\\
    =~& -\E\left[\nu_Y(W)\,(\tilde{D} - \gamma_*^\top \tilde{Z})\right] = 0
\end{align*}
Similarly, the directional derivative with respect to $q_D(\cdot)=\E[D\mid W=\cdot]$ in direction $\nu_D$ is:
\begin{align*}
    \partial_{q_D}[\nu_D] \defeq~& \partial_{\tau} \E[(\tilde{Y} - \tilde{X}^\top h_* - {D} \theta_0 + q_D(W) + \tau \theta_0 \nu_D(W)) (\tilde{D} - q_D(W) - \tau \nu_D(W) - \gamma_*^\top Z)]\mid_{\tau=0}\\
    =~& \E\left[-\nu_D(W)\left(\tilde{Y} - \tilde{X}^\top h - \tilde{D}\theta_0\right) + \nu_D(W) \theta_0(\tilde{D} - \gamma_*^\top \tilde{Z})\right] = 0
\end{align*}
Similarly, the directional derivatives with respect to $q_X=\E[X\mid W=\cdot]$ and $q_Z=\E[Z\mid W=\cdot]$ can be verified to be zero.

\end{itemize}

\end{proof}

\subsubappendix{Asymptotic Normality}
We first define some notation. Let $U$ be the vector of random variables $(W, X, Z, D, Y)$ and one instance as $u=(w, x, z, d, y)$. Denote the nuisance parameter vector as $\eta = (h, \gamma, q_X, q_Z, q_D, q_Y)$, with $q_V(\cdot)=\E[V\mid W=\cdot]$ and $q_{0,V}(W)=E[V\mid W]$ for $V\in [Y, Z, X, D]$. Denote the function $\nu_V = \hat{q}_V- q_{0,V}$. For any function $\nu$ over an input space $W$, denote with $\|\nu\| = \sqrt{\E[\nu(W)^2]}$, the $L^2$ norm, and for any parameter vector $x\in \R^p$, we denote the squared $\ell_2-$norm with $\|x\|=\sqrt{\sum_{i=1}^p x_i^2}$.

Our estimator is based on a moment restriction that is Neyman orthogonal and linear in the target parameter $\theta$:
\begin{align}
    \psi(u; \theta, \eta) = \psi^{a}(u; \eta) \theta + \psi^{b}(u; \eta)
\end{align}
where
\begin{align}
    \psi^{a}(u;\eta) =~& - (D - q_D(W))\,(D - q_D(W) - \gamma^\top (Z - q_Z(W))) \\
    \psi^{b}(u;\eta)=~& (Y - q_Y(W) - (X - q_X(W))^\top h)\, \,(D - q_D(W) - \gamma^\top (Z - q_Z(W)))
\end{align}
We want to prove such an estimator for $\theta$ is both asymptotically linear and asymptotically normal. 
To do so, we apply \cite[Theorem~3.1]{chernozhukov2016doubleArxiv} and show all the necessary regularity conditions for asymptotic behavior is satisfied.
\begin{theorem}\label{thm:normality}
    Suppose that all random variables $W, X, Z, D, Y$ are almost surely and absolutely bounded by a constant, and that the nuisance estimates are consistent:
    \begin{align}
        \|\hat{h}-h_*\| = o_p(1) & & \|\hat{\gamma}-\gamma_*\| = o_p(1) & &  \|\hat{q}_V - q_{0,V}\| = o_p(1)
    \end{align}
    Furthermore, suppose that the nuisance estimates satisfy the rate conditions: 
    \begin{align}
        \sqrt{n}\left( \|\nu_D\|^2 + \|\nu_D\|\, \|\nu_Z\| + \|\nu_Y\|\, \|\nu_D\| + \|\nu_Y\|\, \|\nu_Z\| + \|\nu_D\|\, \|\nu_X\| + \|\nu_Z\|\, \|\nu_X\|\right) = o_p(1)
    \end{align}
    and:
    \begin{align}
        \sqrt{n} (\hat{\gamma} - \gamma_*) \E[\tilde{Z}\tilde{X}^\top] (\hat{h}-h_*) = o_p(1)
    \end{align}
    Then the parameter estimate $\hat{\theta}$ satisfies:
    \begin{align}\label{app:asym_est1}
        \sqrt{n} (\hat{\theta}-\theta_0) = \frac{1}{\sqrt{n}} \sum_{i=1}^n \frac{(\tilde{Y}_i - \tilde{X}_i^\top h_* - \tilde{D}_i \theta_0)\, (\tilde{D}_i - \gamma_*^\top \tilde{Z}_i)}{\E[\tilde{D}\, (\tilde{D}-\gamma_*^\top \tilde{Z})]} + o_p(1) \to_d N\left(0, \sigma^2\right) 
    \end{align}
    with 
    \begin{align}\label{app:asym_est2}
    \sigma^2 = \frac{\E[(\tilde{Y} - \tilde{X}^\top h_* - \tilde{D} \theta_0)^2\, (\tilde{D} - \gamma_*^\top \tilde{Z})^2]}{\E[\tilde{D}\, (\tilde{D}-\gamma_*^\top \tilde{Z})]^2}.
    \end{align}
    Moreover,
     $(1-\alpha)$-confidence intervals can be constructed as:
    \begin{align}\label{app:asym_est3}
        \Pr\left(\theta_0 \in [\hat{\theta} \pm \Phi^{-1}(1-\alpha/2)\,\sqrt{\hat{\sigma}^2/n}]\right) \to 1-\alpha
    \end{align}
    where $\Phi$ is the CDF of the standard normal distribution and \footnote{We denote $\hat{\tilde{V}} = V - \hat{q}_V(W)$, where $\hat{q}_V$ is the estimate of the regression function $q_V(\cdot)=\E[V\mid W=\cdot]$.}:
    \begin{align}\label{app:asym_est4}
        \hat{\sigma}^2 = \frac{\E_n[(\hat{\tilde{Y}} - \hat{\tilde{X}}^\top \hat{h} - \hat{\tilde{D}} \hat{\theta})^2\, (\hat{\tilde{D}} - \hat{\gamma}^\top \hat{\tilde{Z}})^2]}{\E_n[\hat{\tilde{D}}\, (\hat{\tilde{D}}-\hat{\gamma}^\top \hat{\tilde{Z}})]^2}
    \end{align}
\end{theorem}
\begin{proof}
    We need to verify the regularity conditions of \cite[Theorem~3.1]{chernozhukov2016doubleArxiv}. We already have satisfied three conditions: (1) that the moment must satisfy Neyman orthogonality; (2) the nuisance functions must be consistent, i.e.:
    \begin{align}
        \|\hat{\eta} - \eta_0\| = o_p(1) 
    \end{align}
    and (3) the true Jacobian of the moment with respect to the parameter $\theta$ is invertible, which is satisfied as the true Jacobian $\E[\phi^{a}(U; \eta_0)] = \E[\tilde{D} (\tilde{D} - \gamma_*^\top \tilde{Z})]$ is assumed to be non-zero.
    Next, note that since all the variables are uniformly and absolutely bounded, the function $\psi^{\alpha}$ is Lipschitz in the nuisance functions, i.e.:
    % \klcomment{Why are we only showing Lipschitz for $\psi^a$ and not $\psi$? also I deleted the $\psi^2$?}\vscomment{That's what the theorem requires. Lipschitz of $\psi$ is implied by Mean-squared continuity}
    \begin{align}
        |\E[\psi^{a}(U; \eta)] - \E[\psi^{a}(U; \eta_0)]| \leq C\, \|\eta - \eta_0\|
    \end{align}
    for some large enough constant $C$, where $\|\eta-\eta_0\|^2 = \sum_{V\in [X, Z, D, Y]} \|q_V - q_{0,V}\|^2 + \|h - h_0\|^2 + \|\gamma - \gamma_0\|^2$. Moreover, the moments $\psi$ and $\psi^a$ both satisfy the mean-squared continuity condition:
    \begin{align}
        \E[\|\psi(U; \theta_0, \eta) - \phi(U;\theta_0, \eta_0)\|^2] \leq~& C \|\eta - \eta_0\|^2\\
        \E\left[\left(\psi^{a}(U; \eta) - \psi^{a}(U; \eta_0)\right)^2\right] \leq~& C\, \|\eta - \eta_0\|^2
    \end{align}
    
    Next, we need to satisfy that the Hessian of the moment with respect to the nuisance functions in the directions of their errors converges to zero faster than $n^{-1/2}$. That is,
    \begin{align}
        \sqrt{n}\left(\partial_{\tau}^2 \E[\psi(U; \theta_0, \eta_0 + \tau (\hat{\eta}-\eta_0))]\mid_{\tau=\bar{\tau}}\right) = o_p(1)
    \end{align}
    Let $\nu_V(W) = \hat{q}_V(W) - q_{0,V}(W)$. The Hessian can be decomposed as a sum of separate terms. First, we have the terms related to the second order derivatives with respect to the nuisance functions $q_V$, for $V\in [Z, X, D, Y]$:
    \begin{align*}
        A_1 =~& 2 \E[\nu_D(W)^2] - 2 \E[\nu_D(W)\, \gamma_*^\top \nu_Z(W)] = O\left(\|\nu_D\|^2 + \|\nu_D\|\, \|\nu_Z\|\right)\\
        A_2 =~& 2\E[\nu_Y(W) \nu_D(W)] - 2\E[\nu_Y(W) \gamma_*^\top\nu_Z(W)] - 2\E[\nu_D(W)\, h_*^\top \nu_X(W)] + 2\E[\gamma_*^\top \nu_Z(W)\, h_*^\top \nu_X(W)]\\
        =~& O\left(\|\nu_Y\|\, \|\nu_D\| + \|\nu_Y\|\, \|\nu_Z\| + \|\nu_D\|\, \|\nu_X\| + \|\nu_Z\|\, \|\nu_X\|\right)
    \end{align*}
    Finally, we have the term related to the second order derivative with respect to the vector $(h, \gamma)$:
    \begin{align*}
        A_3 = 2 (\hat{\gamma}-\gamma_*)^\top \E[\tilde{X} \tilde{Z}^\top] (\hat{h}-h_*) 
    \end{align*}
    Thus we need that:
    \begin{align}
        \sqrt{n} (A_1 + A_2 + A_3) = o_p(1)
    \end{align}
    This condition is satisfied under our assumptions.
    Finally, we note that our estimation algorithm does not invoke sample splitting when estimating the nuisance parameters $h, \gamma$. However, these parameters are low dimensional and hence the corresponding spaces satisfy the Donsker condition. Therefore, sample splitting can be avoided for these particular nuisance parameters. 
    Given that the regularity conditions are all met, we can invoke the conclusion of \cite[Theorem~3.1]{chernozhukov2016doubleArxiv}] which states that the estimate $\hat{\theta}$ satisfies the asymptotic properties stated in Equations \eqref{app:asym_est1}-\eqref{app:asym_est4}.
\end{proof}

\newpage
\vfill
\subappendix{Estimation and confidence interval construction algorithm}\label{app:algo}

\begin{algorithm}[H]
\caption{Estimation Algorithm}
\begin{algorithmic}[1]
    \State \textbf{Input: } samples of $W, D, Z, X, Y$
    \State \textbf{Output: }{controlled direct effect estimate $\hat{\theta}$ and standard error $\hat{\sigma}$}
    \State Residualize $W$ from $(D, Z, X, Y)$ using LASSO regularization and cross-fitting to select $\lambda$:
    \begin{align}
        \tilde{D} =~& D - \beta_D^\top W & \beta_D = \arg\min_{\beta} \E_n[(D - \beta_D^\top W)^2] + \lambda \|\beta_D\|_1\\
        \forall i\in [p_z]: \tilde{Z}_i =~& Z_i - \beta_{Z_i}^\top W & \beta_{Z_i} = \arg\min_{\beta} \E_n[(Z_i - \beta_{Z_i}^\top W)^2] + \lambda \|\beta_{Z_i}\|_1\\
        \forall i\in [p_x]: \tilde{X}_i =~& X_i - \beta_{X_i}^\top W & \beta_{X_i} = \arg\min_{\beta} \E_n[(X_i - \beta_{X_i}^\top W)^2] + \lambda \|\beta_{X_i}\|_1\\
        \tilde{Y} =~& Y - \beta_Y^\top W & \beta_Y = \arg\min_{\beta} \E_n[(Y - \beta_Y^\top W)^2] + \lambda \|\beta_Y\|_1
    \end{align}

    \State Calculate estimates $\hat{h}, \hat{\theta}_{pre}$ of the minimum norm solution $(h_*, \theta_0)$ to the primal moment equation
    \begin{align}
    \E\left[(\tilde{Y} - \tilde{X}^\top h - \tilde{D}\theta) \begin{pmatrix} \tilde{Z}\\\tilde{D}\end{pmatrix}\right]=0
    \end{align}
    using an linear regularized adversarial IV solver (see Appendix~\ref{app:adviv}) with instruments $(\tilde{Z};\tilde{D})$, treatments $(\tilde{X};\tilde{D})$, outcome $\tilde{Y}$ and penalty parameter $\alpha\sim n^{0.3}$ (equiv. penalty level $\lambda \sim n^{-0.7}$).
    
    \State Calculate estimate $\hat{\gamma}$ of the minimum norm solution ${\gamma_*}$ to the dual moment equation 
    \begin{align*}
        E[(\tilde{D} - \gamma^\top \tilde{Z})\tilde{X}] = 0 
    \end{align*}
    using an linear regularized adversarial IV solver (see Appendix~\ref{app:adviv}) with instruments $\tilde{X}$, treatments $\tilde{Z}$ and outcome $\tilde{D}$ and $\alpha\sim n^{0.3}$ (equiv. penalty level $\lambda \sim n^{-0.7}$).
    
    \State Calculate final estimate $\hat{\theta}$ of the solution $\theta_0$ to the Neyman orthogonal moment equation
    $$\E[(\tilde{Y} - \tilde{X}^\top  \hat{h} - \tilde{D}\theta)\,(\tilde{D} - \tilde{Z}^\top  \hat{\gamma})] = 0,$$ and corresponding standard error of estimate
    as:
    \begin{align}
        \hat{\theta} =~& \frac{\E_n[(\tilde{Y} - \hat{h}^\top  \tilde{X})\, (\tilde{D} - \hat{\gamma}^\top  \tilde{Z})]}{\E_n[\tilde{D} (\tilde{D} - \hat{\gamma}^\top \tilde{Z})]} &
        \hat{\sigma} = \frac{1}{\sqrt{n}} \sqrt{\frac{\E_n[(\tilde{Y} - \hat{h}^\top  \tilde{X} - \tilde{D} \hat{\theta})^2 (\tilde{D} - \hat{\gamma}^\top  \tilde{Z})^2]}{\E_n[\tilde{D} (\tilde{D} - \hat{\gamma}^\top  \tilde{Z})]^2}}
    \end{align}
\end{algorithmic}
\end{algorithm}

\subappendix{Linear Regularized Adversarial IV}
\subsubappendix{Procedure}\label{app:adviv}
We define in detail a generic procedure for estimating the minimum norm solution $\theta_0$ to a linear instrumental variable (IV) regression problem of the form:
$$\E[(Y - \theta^\top  X) Z]=0$$
for some outcome $Y$, instrument vector $Z$, and treatment vector $X$, where the covariance matrix $\E[ZX^\top ]$ is ill-posed, i.e., has a zero singular value. (Note that the optimal solution $\hat{\theta}$ to this IV problem is a general parameter and thus separate from the implicit bias effect estimate.) Our procedure is based on the $\ell_2$-regularized adversarial IV method that optimizes the empirical criterion defined in [\citen{adviv}] that solves:
\begin{equation}
    \min_{\theta} \max_{\beta} \E_n[2 (Y - \theta^\top  X) Z^\top  \beta - \beta^\top  ZZ^\top  \beta] + \lambda \|\theta\|^2
\end{equation}
The optimal solution to the inner optimization problem takes the form:
\begin{align}
    \beta_* = \E_n[ZZ^\top ]^+ \E_n[Z (Y - \theta^\top  X)] 
\end{align}
and thus, using the optimal solution, we get:
\begin{align}
    \max_{\beta} \E_n[2 (Y - \theta^\top  X) Z^\top  \beta - \beta^\top  ZZ^\top  \beta] = \E_n[(Y - \theta^\top  X) Z^\top ]\,\E_n[ZZ^\top ]^+ \E_n[Z(Y - \theta^\top  X) Z]
\end{align}
Thus $\hat{\theta}$ is defined as the minimizer of:
\begin{align}
    \min_{\theta} \E_n[(Y - \theta^\top  X) Z^\top ]\,\E_n[ZZ^\top ]^+ \E_n[Z(Y - \theta^\top  X)] + \lambda \|\theta\|^2
\end{align}
We can define the random variable:
\begin{align}
    \hat{Q} = \E_n[XZ^\top ]\E_n[ZZ^\top ]^+ Z
\end{align}
Note that $\hat{B} = \E_n[ZZ^\top ]^+ \E_n[ZX^\top ]$ is the minimum norm solution to the OLS problem of predicting $X$ linearly as a function of $Z$. 
Hence, $\hat{Q}=\hat{B}^\top  Z$ is a best linear prediction of $X$ as a function of $Z$. Notably, $\hat{B}$ satisfies the first order conditions to the OLS problem, i.e. 
\begin{align}
    \E_n[Z (X - \hat{B}^\top  Z)^\top ] = 0 \implies \hat{B}^\top  \E_n[Z (X - \hat{B}^\top  Z)^\top ] = 0 \implies \E_n[\hat{Q} X^\top ] = \E_n[\hat{Q}\hat{Q}^\top ]
\end{align}
Then the solution $\hat{\theta}$ to the regularized adversarial IV problem is the solution to the equation:
\begin{align}\label{app:adiv_theta}
   \E_n[\hat{Q}\, (Y - X^\top  \theta)] + \lambda \theta = 0 \Leftrightarrow \E_n[\hat{Q} Y] - (\E_n[\hat{Q}X^\top ] + \lambda I) \theta = 0 
\end{align}
which takes the closed form:
\begin{align}
    \hat{\theta} = (\E_n[\hat{Q}X^\top ] + \lambda I)^+ \E_n[\hat{Q}Y] = (\E_n[\hat{Q}\hat{Q}^\top ] + \lambda I)^+ \E_n[\hat{Q}Y]
\end{align}
Thus in this setting, the regularized adversarial IV solution is equivalent to a regularized two-stage-least-squares procedure where we use the empirical minimum norm solution for the first stage and we use a $\ell_2$-regularized second stage with penalty level $\lambda$.

\begin{algorithm}[H]
\caption{Regularized Adversarial IV Procedure: $\text{AdvIV}(Z, X, Y; \alpha)$}\label{alg:adviv}
\begin{algorithmic}[1]
    \State \textbf{Input: } $n$ samples of the variables $(Z, X, Y)$, with instrument vector $Z$, treatment vector $X$, outcome $Y$; $\ell_2$ penalty hyperparameter $\alpha$ 
    \State \textbf{Output: }{Estimate of the minimum norm solution $\hat{\theta}$ to the linear IV problem defined via the moment restriction equations: $\E[(Y - \theta^\top  X) Z]=0$}
    \State Calculate the OLS solution of the first stage, giving the minimum mean-squared-projection $\hat{Q}$ of $X$ on $Z$:
    \begin{align}
        \hat{Q} =~& \hat{B}^\top  Z &
        \hat{B} =~& \E_n[ZZ^\top ]^+ \E_n[ZX^\top ]
    \end{align}
    \State Calculate the regularized second stage solution:
    \begin{align}
        \hat{\theta} = \left(\E_n[\hat{Q}X^\top ] + \frac{\alpha}{n} I\right)^+ \E_n[\hat{Q}Y] = \left(\E_n[\hat{Q}\hat{Q}^\top ]^{-1} + \frac{\alpha}{n} I\right)^+ \E_n[\hat{Q}Y]
    \end{align}
\end{algorithmic}
\end{algorithm}

Following very similar arguments as the ones presented in  \cite[Theorem~4]{bennett2023source} (see Appendix~\ref{app:rates}), we can argue that $\hat{\theta}$ converges to the minimum norm solution $\theta_0$, as long as $\lambda=\omega(1/n)$ and the covariance matrix $\E[ZZ^\top]$ is full rank, as well as several other regularity assumptions. Moreover, the convergence of $\|\hat{\theta}-\theta_0\|$ follows a rate of $\sqrt{n^{-1}/\lambda + \lambda}$, while the convergence in terms of the weak norm 
$$\|\hat{\theta}-\theta_0\|_{w}\defeq \|\E[ZX^\top](\hat{\theta}-\theta_0)\|$$
should follow a rate of $\sqrt{n^{-1} + \lambda^2}$. In particular, in Appendix~\ref{app:rates} we prove the following theorem:

\begin{theorem}[Rates]\label{thm:rates}
Assume that $\E[ZZ^\top]$ is full rank. Let $\Sigma_0 = \E[ZX^\top]$ and $\tilde{\Sigma}_0 = \E[ZZ^\top]^{-1/2}\E[ZX^\top]$ and $\theta_0$ be the minimum norm solution $\theta_0 = \tilde{\Sigma}_0^+ \E[ZY]$ to the IV problem $\E[ZZ^\top]^{-1/2}\E[Z(Y - \theta^\top  X)]=0$. Assume that $X, Z, Y$ are uniformly and absolutely bounded. Given $n$ samples of $(Z,X,Y)$, let $\hat{\theta}$ be the solution to the adversarial IV method described in Algorithm~\ref{alg:adviv}, with $\lambda=\alpha/n$ chosen such that $\lambda=o(1)$ and $n\lambda \to \infty$. Then $\hat{\theta}$ satisfies:
\begin{align}
    \|\hat{\theta}- \theta_0\|_w^2 
    \leq~& O_p\left(\lambda^2 + \frac{1}{n}(1+\|\theta_0\|^2)\right) \\
    \|\hat{\theta}- \theta_0\|^2 \leq~& O_p\left(\lambda + \frac{1}{\lambda n} (1+\|\theta_0\|^2)\right)
\end{align}
\end{theorem}
In essence, this theorem states that $\hat{\theta}$ will eventually converge, in probability, to the true $\theta_0$, i.e., that the estimate is consistent.

We will use the regularized IV algorithm to estimate both of our nuisance parameters $h_*, \gamma_*$, with the corresponding definitions of an instrument, treatment and outcome and the corresponding definitions of convergence metrics (note, then, that the parameter $\theta$ in this section then refers to these nuisance parameters). Recall under Theorem~\ref{thm:normality}, to prove asymptotic normality of our target parameter estimate $\hat{\theta}$ (i.e., the implicit bias effect) in the context of our main algorithm, we need the product of the errors of the primal estimate $\hat{h}$ and the dual estimate $\hat{\gamma}$, both obtained by the adversarial IV method, to satisfy the convergence property:
\begin{align}
    \sqrt{n} (\hat{h} - h_0)^\top\E\left[X Z^\top\right] (\hat{\gamma} - \gamma_*) = o_p(1)
\end{align}
Note that a sufficient condition is that:
\begin{align*}
    \sqrt{n} \|\hat{h}-h_*\|\, \|\gamma_*- \hat{\gamma}\|_w = o_p(1)
\end{align*}
It suffices to show that $n \|\hat{h}-h_*\|^2\, \|\gamma_*- \hat{\gamma}\|^2_w = o_p(1)$. Applying the results from Theorem \ref{thm:rates}, we have that $n \|\hat{h}-h_*\|^2\, \|\gamma_*- \hat{\gamma}\|^2_w$ is of order:
\begin{align*}
n \left(\frac{1}{n\lambda} + \lambda\right)\, \left(\frac{1}{n} + \lambda^2\right) =~& \left(\frac{1}{n\lambda} + \lambda\right)\, \left(1 + n \lambda^2\right)
=~ \frac{1}{n\lambda} + 2\lambda + n \lambda^3
\end{align*}
Thus if we choose $\lambda = o(n^{-1/3})$ and $\lambda=\omega(n^{-1})$, then $\frac{1}{n\lambda} + 2\lambda + n \lambda^3$ will be $o(1)$, and we have from Theorem~\ref{thm:normality} that the primal and dual IV estimates are asymptotically normal. In our experiments we chose $\alpha\sim n^{0.3}\implies \lambda \sim 1/n^{0.7}$, which satisfies both properties. 

\subsubappendix{Convergence Rate of Estimate}\label{app:rates}
% \klcomment{Were these proofs not proved in the original paper? these seem like fundamental properties of the convergence of the advIV method that might have been proved already}\vscomment{Not for the version we use here. The original paper was making other regularity assumptions that don't apply for the exact version we use here, which are fine because we have linearity. The original paper was not assuming linearity. So these proofs need to be re-done. For instance, to be able to invoke the original rates we would need to impose a hard constraint on the $\ell_2$-norm of the estimates, which would make the implementation harder and not closed form. Moreover, the theorem in the paper requires a closedness assumption i.e. that $\E[X\mid Z]$ be a linear function of $Z$, which we don't require here. Here we only need the IV function to be linear. So this is essentially a new theorem. It just follows the exact same proof pattern as the proof in that paper.}

In this section we prove Theorem \ref{app:rates} of the convergence of the regularized adversarial estimate $\hat{\theta}$ to the true estimate $\theta_0$, with respect to an upper bound as a function of the penalty parameter $\lambda$.

First, let $\theta_0$ be the minimum norm solution to the un-regularized minimization problem:
\begin{align}
    \min_{\theta} \|\E[ZZ^\top ]^{-1/2}\E[(Y - \theta^\top  X) Z]\|^2
\end{align}
This minimization problem is assumed to have an optimal value of zero, or, equivalently, $\theta_0$ is the minimum norm solution to the linear system $\E[ZZ^\top]^{-1/2} \E[ZX^\top] \theta = \E[ZZ^\top]^{-1/2}\E[ZY]$, or $\tilde{\Sigma}_0 \theta = \E[ZZ^\top]^{-1/2}\E[ZY]$. Since $\E[ZZ^\top]$ is invertible,
the minimum norm solution $\theta_0$ also satisfies $\E[ZY] = \E[ZX^\top]\theta_0$.

Then, we define a parameter $\theta_{\lambda}$ as a solution to the \textit{population} regularized adversarial criteria for a general parameter $\lambda$. We can then find the error of $\theta_{\lambda}$ from the true $\theta_0$, both in terms of strong and weak norm (i.e., where the error is multiplied by $\Sigma_0$).

To define $\theta_{\lambda}$, first consider the \textit{population} adversarial criterion:
\begin{equation}
    \min_{\theta} \max_{\beta} \E[2 (Y - \theta^\top  X) Z^\top  \beta - 2\beta^\top  ZZ^\top  \beta] + \lambda \|\theta\|^2
\end{equation}
By similar argument to the empirical adversarial case, we can solve for the optimal $\beta$,
\begin{align}
    \beta_* = \frac{1}{2}\E[ZZ^\top]^{-1} \E[Z (Y - \theta^\top  X)] 
\end{align}
and plugging in, the inner optimal solution takes the form
\begin{align*}
\max_{\beta} \E[2 (Y - \theta^\top  X) Z^\top  \beta - 2\beta^\top  ZZ^\top  \beta] + \lambda \|\theta\|^2 =~&
    \frac{1}{2}\E[(Y - \theta^\top  X) Z^\top ]\,\E[ZZ^\top ]^{-1} \E[Z(Y - \theta^\top  X) Z] \\ =~& \frac{1}{2}\|\E[ZZ^\top ]^{-1/2}\E[(Y - \theta^\top  X) Z]\|^2
\end{align*}
We can furthermore rewrite the population criterion in terms of $\theta_0$ where we see that:
\begin{align}\label{eq:regl_iv}
    \theta_\lambda =~& \argmin_{\theta} \frac{1}{2}\|\E[ZZ^\top ]^{-1/2}\E[(ZY - ZX^\top  \theta)]\|^2 + \lambda \|\theta\|^2 \\
    =~& \argmin_{\theta} \frac{1}{2}\|\E[ZZ^\top]^{-1/2} \E[ZX^\top] (\theta_0 - \theta)\|^2 + \lambda \|\theta\|^2 \\ =~& \argmin_{\theta} \frac{1}{2}\|\tilde{\Sigma}_0(\theta - \theta_0)\|^2 + \lambda \|\theta\|^2 \defeq L_{\lambda}(\theta)
\end{align}

We will first bound the error of our empirical regularized estimate $\hat{\theta}$ with respect to this population regularized estimate $\theta_\lambda$. Then we will combine it with a bound on the error between $\theta_\lambda$ and the minimum norm solution $\theta_0$ to conclude the proof.

First, note that $\theta_{\lambda}$ minimizes the loss $L_{\lambda}(\theta)$. Moreover, we show that the loss is strongly convex, which implies a bound on the distance between the true minimizer $\theta_{\lambda}$ and any other $\theta$ (including $\hat{\theta}$). In particular, we can invoke strong convexity of this loss function to upper bound the error of our estimate $\hat{\theta}$ with respect to $\theta_\lambda$, as measured by a combination of a strong and a weak norm.
\begin{lemma}[Strong Convexity]\label{lem:strong-convexity}
    The loss function $L_{\lambda}(\theta) = \|\tilde{\Sigma}_0(\theta - \theta_0)\|^2 + 2\lambda \|\theta\|^2$ is strongly convex with:
    \begin{align}
        \partial_{\theta}^2 L(\theta) = 2 (\tilde{\Sigma}_0^\top \tilde{\Sigma}_0 + 2\lambda I)
    \end{align}
    Since $\theta_\lambda$ optimizes the objective, we have:
    \begin{align}
        L_{\lambda}(\theta) - L_{\lambda}(\theta_{\lambda}) \geq (\theta - \theta_{\lambda})^\top (\tilde{\Sigma}_0^\top \tilde{\Sigma}_0 + 2\lambda I) (\theta - \theta_{\lambda}) = \|\tilde{\Sigma}_0(\theta- \theta_\lambda)\|^2 + 2\lambda \|\theta - \theta_\lambda\|^2
    \end{align}
\end{lemma}

Applying Lemma~\ref{lem:strong-convexity} with $\theta=\hat{\theta}$, we have that:
\begin{align}\label{eq:convex_bound},
    \|\tilde{\Sigma}_0(\hat{\theta}- \theta_\lambda)\|^2 + 2\lambda \|\hat{\theta} - \theta_\lambda\|^2 \leq \|\tilde{\Sigma}_0 (\hat{\theta}-\theta_0)\|^2 - \|\tilde{\Sigma}_0 (\theta_\lambda - \theta_0)\|^2 + 2\lambda (\|\hat{\theta}\|^2 - \|\theta_\lambda\|^2)
\end{align}
Recall the weak norm error is defined as $L(\theta)=\|\tilde{\Sigma}_0 (\theta-\theta_0)\|$. Thus we need to upper bound the difference on the right hand side between the weak norm error $L(\theta_{\lambda})$ of the population adversarial estimate $\theta_{\lambda}$ 
and the error $L(\hat{\lambda}$), evaluated at the empirical estimate $\hat{\theta}$. If we can prove such a bound exists, then we have the empirical adversarial estimate $\hat{\theta}$ satisfies an oracle inequality with respect to the weak norm error. The following lemma states that a bound exists as long as $n$ is sufficiently large. We defer the proof to Appendix~\ref{app:proof-oracle-ineq}.

\begin{lemma}[Oracle inequality]\label{lem:oracle-ineq}
    Let $\theta_0$ be the minimum norm solution to the linear system $\E[ZZ^\top]^{-1/2} \E[Z(Y-\theta^\top X)]=0$, $\theta_{\lambda}$ the population $\ell_2$-regularized adversarial estimate, and $\hat{\theta}$ the empirical $\ell_2$-regularized adversarial estimate with appropriate choice of $\lambda$. Then the adversarial estimate satisfies the oracle inequality as long as $n\geq C \log(p_X \cdot p_Z/\delta)$ for some large enough universal constant $C$ with probability $1-\delta$:
    \begin{multline*}
        \|\tilde{\Sigma}_0(\hat{\theta}-\theta_0)\|^2 - \|\tilde{\Sigma}_0(\theta_\lambda - \theta_0)\|^2 \leq 7 \|\tilde{\Sigma}_0(\theta_\lambda - \theta_0)\|^2 + 2\lambda (\|\theta_\lambda\|^2 - \|\hat{\theta}\|^2)\\
        + O\left(\sqrt{\frac{\log(p_X\cdot p_Z/\delta)}{n}} \|\tilde{\Sigma}_0(\hat{\theta}-\theta_\lambda)\| (1 + \|\hat{\theta}-\theta_\lambda\| + \|\theta_0\|) + \frac{\log(p_X\cdot p_Z / \delta)}{n} (1 + \|\theta_0\|)^2\right) 
    \end{multline*}
\end{lemma}

If we apply Lemma~\ref{lem:oracle-ineq} to the right hand side of Equation \eqref{eq:convex_bound}, we have:
\begin{multline*}
    \|\tilde{\Sigma}_0(\hat{\theta}- \theta_\lambda)\|^2 + 2\lambda \|\hat{\theta} - \theta_\lambda\|^2 \leq 7 \|\tilde{\Sigma}_0(\theta_\lambda-\theta_0)\|^2 + O\left(\sqrt{\frac{\log(p_X\cdot p_Z/\delta)}{n}} \|\tilde{\Sigma}_0(\hat{\theta}-\theta_\lambda)\| (1 + \|\hat{\theta}-\theta_\lambda\| + \|\theta_0\|)\right)\\
    + O\left(\frac{\log(p_X\cdot p_Z / \delta)}{n} (1 + \|\theta_0\|)^2\right)
\end{multline*}
Applying the AM-GM inequality to the second term of the right hand side, we get:\footnote{We use that $O(a\cdot b) \leq \frac{1}{2}a^2 + O(b^2)$ for any $a,b>0$, and $O(a\cdot b)\leq C\cdot a \cdot b$ for some constant $C$. Applying the AM-GM inequality $C \cdot a \cdot b = a \cdot (C\cdot b)\leq \frac{1}{2}a^2 + \frac{1}{2} C^2 b^2= \frac{1}{2}a^2 +O(b^2)$. Then the result follows be invoking this property for $a=\|\tilde{\Sigma}_0 (\hat{\theta}-\theta_\lambda)\|$ and $b = \sqrt{\frac{\log(p_X\cdot p_Z/\delta)}{n}}  (1 + \|\hat{\theta}-\theta_\lambda\| + \|\theta_0\|)$. Moreover, $(a + b + c)^2 = O(a^2 + b^2 + c^2)$.}
\begin{align*}
    \frac{1}{2}\|\tilde{\Sigma}_0(\hat{\theta}- \theta_\lambda)\|^2 + 2\lambda \|\hat{\theta} - \theta_\lambda\|^2 \leq 7 \|\tilde{\Sigma}_0(\theta_\lambda-\theta_0)\|^2 + C\left(\frac{\log(p_X\cdot p_Z/\delta)}{n} (1 + \|\hat{\theta}-\theta_\lambda\|^2 + \|\theta_0\|^2)\right) 
\end{align*}
for some constant $C$. For $n\geq \frac{2}{3\lambda} C \log(p_X\cdot p_Z/\delta)$ (which will eventually hold if $\lambda$ is chosen such that $n\lambda \to \infty$), we have that $C\frac{\log(p_X\cdot p_Z/\delta)}{n} \|\hat{\theta}-\theta_\lambda\|^2\leq \frac{3}{2}\lambda \|\hat{\theta}-\theta_\lambda\|^2$, hence we can bring this part from the right hand side to the left hand side and simplify the above as:
\begin{align}\label{eq:intmd-bound}
    \frac{1}{2}\|\tilde{\Sigma}_0(\hat{\theta}- \theta_\lambda)\|^2 + \frac{\lambda}{2} \|\hat{\theta} - \theta_\lambda\|^2 \leq 7 \|\tilde{\Sigma}_0(\theta_\lambda-\theta_0)\|^2 + C\left(\frac{\log(p_X\cdot p_Z/\delta)}{n} (1 + \|\theta_0\|^2)\right) 
\end{align}

Thus we have derived a bound on the error between $\hat{\theta}$ and $\theta_\lambda$ as a function of the erorr between $\theta_\lambda$ and $\theta_0$. We next bound the error of the regularized population adversarial solution $\theta_{\lambda}$ with respect to $\theta_0$.

\begin{lemma}[Bias of $\theta_{\lambda}$]\label{lem:bias}
Let $\theta_0$ be the minimum norm solution to the linear system $\E[ZZ^\top]^{-1/2} \E[Z(Y-\theta^\top X)]=0$. Let $\tilde{\Sigma}_0=UDV^\top$ by the SVD of $\tilde{\Sigma}_0$, with $D=\text{diag}(\sigma_1, \ldots, \sigma_K, 0, \ldots, 0)$ and $K$ the rank of $\tilde{\Sigma}_0$. Let $a_0 = V^\top \theta_0$. Then the population solution $\theta_{\lambda}$ using penalty parameter $\lambda$ satisfies that:
\begin{align}
\|{\theta}_{\lambda} - \theta_0\|^2 \leq~& 4\lambda^2 \sum_{i=1}^K \frac{a_{0,i}^2}{\sigma_i^4} = O(\lambda^2) &
\|\tilde{\Sigma}_0({\theta}_{\lambda} - \theta_0)\|^2 \leq~& 4\lambda^2 \sum_{i=1}^K \frac{a_{0,i}^2}{\sigma_i^2} = O(\lambda^2)
\end{align}
\end{lemma}
% \kledit{Recall the minimum norm solution $\theta_0$ is also solves the regularized problem Equation \eqref{eq:regl_iv}}. 
\begin{proof}
    Let $a=V^\top \theta$ and $a_0=V^\top\theta_0$, and note $\tilde{\Sigma}_0^\top \tilde{\Sigma}_0=V D^2 V^\top$ and $\|\theta\|=\|V^\top\theta\|=\|a\|$. Then we rewrite the regularized problem defining $\theta_{\lambda}$ from Equation \eqref{eq:regl_iv} as:
    \begin{align}
        \min_{a} (a-a_0)^\top D^2 (a-a_0) + 2\lambda \|a\|^2 = \sum_{i=1}^{p_X} \sigma_i^2 (a_i - a_{0,i})^2 + 2\lambda a_i^2
    \end{align}
    Considering the first order condition optimi $a_i$, we can write the optimal solution as:
    \begin{align}
        a_{\lambda,i} = \frac{\sigma_i^2}{\sigma_i^2 + 2\lambda} a_{i,0}
    \end{align}
    Moreover, note that for $i > K$,  $\sigma_i=0$ and the minimum norm solution $\theta_0$ has zero inner product with the eigenvectors $V_{\cdot, i}$. Thus, we have that:
    \begin{align}
        \|\theta_\lambda - \theta_0\|^2 = \|a_\lambda - a_0\|^2 = \sum_{i=1}^K a_{i,0}^2 \left(1 - \frac{\sigma_i^2}{\sigma_i^2 + 2\lambda}\right)^2 = \sum_{i=1}^K a_{i,0}^2 \frac{4\lambda^2}{(\sigma_i^2 + 2\lambda)^2} \leq 4\lambda^2  \sum_{i=1}^K \frac{a_{i,0}^2}{\sigma_i^4}
    \end{align}
    Then the weak norm error $\|\tilde{\Sigma}_0 (\theta_\lambda -\theta_0)\|$ can be written as:
    \begin{align}
        \|\tilde{\Sigma}_0({\theta}_{\lambda} - \theta_0)\|^2 = \sum_{i=1}^K \sigma_i^2 (a_{\lambda,i} - a_{0,i})^2 \leq 4\lambda^2 \sum_{i=1}^K \frac{a_{i,0}^2}{\sigma_i^2}
    \end{align}
\end{proof}

We can now combine all these derived bounds to get the desired error rates.

\paragraph{Weak norm rate} For the weak norm convergence rate (i.e., $\|\E[ZX^\top](\hat{\theta} - \theta_0\|$), we can apply the triangle inequality and Equation~\eqref{eq:intmd-bound} to get:
\begin{align*}
    \|\tilde{\Sigma}_0(\hat{\theta}- \theta_0)\|^2 \leq~& \|\tilde{\Sigma}_0(\hat{\theta} - \theta_\lambda)\|^2 + \|\tilde{\Sigma}_0(\theta_\lambda - \theta_0)\|^2 \leq 2\|\tilde{\Sigma}_0(\hat{\theta} - \theta_\lambda)\|^2 + 2\|\tilde{\Sigma}_0(\theta_\lambda - \theta_0)\|^2 \\
    \leq~& 
    O\left(\|\tilde{\Sigma}_0(\theta _0- \theta_\lambda)\|^2 + \frac{\log(p_X\cdot p_Z/\delta)}{n}(1 + \|\theta_0\|^2)\right)
\end{align*}
Subsequently, by Lemma~\ref{lem:bias}:
\begin{align*}
     \|\tilde{\Sigma}_0(\hat{\theta}- \theta_0)\|^2 \leq~& O\left(\lambda^2 + \frac{\log(p_X\cdot p_Z/\delta)}{n}(1 + \|\theta_0\|^2)\right) 
\end{align*}
Recall $\tilde{\Sigma}_0 = \E[ZZ^\top]^{-1/2} \E[ZX^\top]$. Let $K\Lambda K^\top$ be the eigendecomposition of $\E[ZZ^\top]$ and for any $x\in R^{p_Z}$, let $a_x= K^\top x$. Then we see that
\begin{align*}
    \|\E[ZZ^\top]^{-1/2} x\|^2 =~& x^\top K \Lambda^{-1} K^\top x = a_x^\top \Lambda^{-1} a_x = \sum_{i=1}^{p_Z} \frac{1}{\lambda_i} a_{x,i}^2 \geq \frac{1}{\lambda_{\max}} \sum_{i=1}^{p_Z} a_{x, i}^2\\
    =~& \frac{1}{\lambda_{\max}} a_x^\top a_x = \frac{1}{\lambda_{\max}} x^\top K K^\top x = \frac{1}{\lambda_{\max}} x^\top x = \frac{1}{\lambda_{\max}} \|x\|^2
\end{align*}
where $\lambda_{\max}$ is the maximum eigenvalue of $\E[ZZ^\top]$ (which is non-zero).
Thus we get that for any vector $\nu$:
\begin{align*}
    \|\tilde{\Sigma}_0\nu\|^2 =  \| \E[ZZ^\top]^{-1/2} \E[ZX^\top]\nu\|^2 \geq \frac{1}{\lambda_{\max}} \|\E[ZX^\top]\nu\|^2 = \frac{1}{\lambda_{\max}} \|\nu\|_w^2
\end{align*}
from which we deduce that:
\begin{align*}
    \|\hat{\theta}-\theta_0\|_w^2 = \|\E[ZX^\top] (\hat{\theta}- \theta_0)\|^2 \leq ~& \lambda_{\max} \|\tilde{\Sigma}_0(\hat{\theta}-\theta_0)\|^2 \\ =~&  O\left(\lambda^2 + \frac{\log(p_X\cdot p_Z/\delta)}{n}(1+\|\theta_0\|^2)\right) 
\end{align*}

\paragraph{Strong norm rate} For the $\ell_2$-norm convergence rate, we can invoke the triangle inequality and Equation~\eqref{eq:intmd-bound}
\begin{align*}
    \|\hat{\theta}-\theta_0\|^2 \leq~& \|\hat{\theta}-\theta_\lambda\|^2 + \|\theta_\lambda-\theta_0\|^2 \leq 2\|\hat{\theta}-\theta_\lambda\|^2 + 2\|\theta_\lambda-\theta_0\|^2\\
    \leq~& O\left(\frac{1}{\lambda} \|\tilde{\Sigma}_0(\theta_\lambda - \theta_0)^2\|^2 + \|\theta_\lambda - \theta_0\|^2 + \frac{\log(p_X\cdot p_Z/\delta)}{\lambda n}(1 + \|\theta_0\|^2)\right)
\end{align*}
Subsequently, by Lemma~\ref{lem:bias}:
\begin{align*}
    \|\hat{\theta}-\theta_0\|^2 \leq~& O\left(\lambda + \lambda^2 + \frac{\log(p_X\cdot p_Z/\delta)}{\lambda n} (1 + \|\theta_0\|^2)\right)\\
    =~& O\left(\lambda + \frac{\log(p_X\cdot p_Z/\delta)}{\lambda n} (1+\|\theta_0\|^2)\right)
\end{align*}

Thus, overall we can conclude the rates of the estimator:
\begin{align*}
    \|\hat{\theta}- \theta_0\|_w^2 
    \leq~& O_p\left(\lambda^2 + \frac{1}{n}(1+\|\theta_0\|^2)\right) \\
    \|\hat{\theta}- \theta_0\|^2 \leq~& O_p\left(\lambda + \frac{1}{\lambda n} (1+\|\theta_0\|^2)\right)
\end{align*}

\subsubappendix{Proof of Lemma~\ref{lem:oracle-ineq}}\label{app:proof-oracle-ineq}

\begin{proof}
    Let the optimal solution to the inner population optimization in the adversarial criterion when $\theta=\hat{\theta}$ be $\hat{\beta} = \frac{1}{2}\E[ZZ^\top]^{-1} \E[ZX^\top] (\theta_0 - \hat{\theta}) = \frac{1}{2}\E[ZZ^\top]^{-1/2}\tilde{\Sigma}_0 (\theta_0-\hat{\theta})$. Since $\theta_0$ also satisfies $\E[ZY] = \E[ZX^\top]\theta_0$, we see that:
    \begin{align*}
        A\defeq \frac{1}{2} \|\tilde{\Sigma}_0 (\theta_0 - \hat{\theta})\|^2 =~& 2\hat{\beta}^\top \E[ZX^\top] (\theta_0 - \hat{\theta}) - 2\hat{\beta}^\top \E[ZZ^\top] \hat{\beta}\\
        =~& 2\hat{\beta}^\top (\E[ZY] - \E[ZX^\top]\hat{\theta}) - 2\hat{\beta}^\top \E[ZZ^\top] \hat{\beta}
    \end{align*}
    Since $X,Z,Y$ are bounded, by a Chernoff bound and a union bound, we have that with probability $1-\delta$:
    \begin{align*}
        \|\E[ZX^\top] - \E_n[ZX^\top]\|_{\infty},\, \|\E[ZY] - \E_n[ZY]\|_{\infty},\, \|\E[ZZ^\top] - \E_n[ZZ^\top]\|_{\infty} \leq O\left(\sqrt{\frac{\log(p_X\cdot p_Z / \delta)}{n}}\right)
    \end{align*}
    Thus, we can replace the population covariances with the empirical covariances (leaving one population covariance $\hat{\beta}^\top \E[ZZ^\top] \hat{\beta}$ intact, which will serve its purpose later on in the proof):
    \begin{align*}
    A \leq~& 
    2\hat{\beta}^\top (\E_n[ZY] - \E_n[ZX^\top]\hat{\theta}) - \hat{\beta}^\top \E_n[ZZ^\top] \hat{\beta} - \hat{\beta}^\top \E[ZZ^\top] \hat{\beta} + \epsilon(n)\\
    \leq~& \max_{\beta} 2\beta^\top \E_n[Z(Y- X^\top\hat{\theta})] - \beta^\top \E_n[ZZ^\top] \beta - \hat{\beta}^\top \E[ZZ^\top]\hat{\beta} + \epsilon(n)
    \end{align*}
    where:
    \begin{align*}
        \epsilon(n) =~& O\left(\sqrt{\frac{\log(p_X\cdot p_Z / \delta)}{n}} (\|\hat{\beta}\| + \|\hat{\beta}\|^2 + \|\hat{\beta}\|\,\|\hat{\theta}\|)\right)\\
        \leq~& C\left(\sqrt{\frac{\log(p_X\cdot p_Z / \delta)}{n}} (\|\hat{\beta}\| (1 + \|\hat{\theta}\|) + \hat{\beta}^\top\E[ZZ^\top]\hat{\beta})\right)
    \end{align*}
    for some constant $C$. 
    
    For $n\geq C^2 \log(p_X\cdot p_Z/\delta)$, we have that $C\sqrt{\frac{\log(p_X\cdot p_Z / \delta)}{n}}\hat{\beta}^\top\E[ZZ^\top]\hat{\beta}$ (which will be a positive scalar) will be smaller than the absolute value of the negative term in the optimization criterion $-\hat{\beta}^\top\E[ZZ^\top]\hat{\beta}$. Thus these two terms can be ignored as they sum up to something negative. Thus, we can derive:
    \begin{align*}
        A \leq~& \max_{\beta} 2\beta^\top \E_n[Z(Y- X^\top\hat{\theta})] - \beta^\top \E_n[ZZ^\top] \beta + \zeta(n)
    \end{align*}
    with:
    \begin{align*}
    \zeta(n) = O\left(\sqrt{\frac{\log(p_X\cdot p_Z / \delta)}{n}} \|\hat{\beta}\| (1 + \|\theta_0\| + \|\hat{\theta}-\theta_\lambda\|) \right)
    \end{align*}
    where we also invoked the triangle inequality $\|\hat{\theta}\|\leq \|\theta_\lambda\| + \|\hat{\theta}-\theta_\lambda\|$ and the fact that by the bias Lemma~\ref{lem:bias} $\|\theta_\lambda\|=O(\|\theta_0\|)$.
    Moreover, note that
    \begin{align}
        \|\hat{\beta}\| =~& \norm{\frac{1}{2}\E[ZZ^\top]^{-1/2}\tilde{\Sigma}_0 (\theta_0-\hat{\theta})} \\ =~& O\left(\|\tilde{\Sigma}_0(\theta_0-\hat{\theta})\|\right) \\ =~& O(\sqrt{A})
    \end{align} so we can rewrite 
    \begin{align*}
        \zeta(n) = O\left(\sqrt{\frac{\log(p_X\cdot p_Z / \delta)}{n}} \sqrt{A} (1 + \|\theta_0\| + \|\hat{\theta}-\theta_\lambda\|)\right)
    \end{align*}
    Since $\hat{\theta}$ optimizes the regularized empirical criterion $L_{n,\lambda}(\theta)\defeq \max_{\beta} 2\beta^\top \E_n[Z(Y-X^\top\theta)] - \beta^\top\E_n[ZZ^\top] + \lambda \norm{\theta}^2$ over all parameters $\theta$, it means that $L_{n,\lambda}(\hat{\theta})\leq L_{n,\lambda}(\theta_{\lambda})$. Thus, we can derive:
    \begin{align*}
        A \leq~& \max_{\beta} 2\beta^\top \E_n[Z(Y- X^\top\theta_\lambda)] - \beta^\top \E_n[ZZ^\top] \beta + \zeta(n) + \lambda (\|\theta_\lambda\|^2 - \|\hat{\theta}\|^2)\\
        \leq~& \max_{\beta} 2\beta^\top \E[Z(Y- X^\top\theta_\lambda)] - \beta^\top \E[ZZ^\top] \beta + \delta(n, \beta) + \zeta(n) + r(n)
    \end{align*}
    with:
    \begin{align*}
        r(n) =~& \lambda (\|\theta_\lambda\|^2 - \|\hat{\theta}\|^2)\\
        \delta(n,\beta) =~& O\left(\sqrt{\frac{\log(p_X\cdot p_Z / \delta)}{n}} (\|\beta\| (1 + \|\theta_\lambda\|)+\|\beta\|^2)\right)\\
        \leq~& C\left(\sqrt{\frac{\log(p_X\cdot p_Z / \delta)}{n}} (\sqrt{\beta^\top\E[ZZ^\top]\beta} (1 + \|\theta_\lambda\|)+ \beta^\top\E[ZZ^\top]\beta)\right)
    \end{align*}
    for some constant $C$. For $n\geq 4 C^2 \log(p_X\cdot p_Z / \delta)$ and since by the Lemma \ref{lem:bias} $\|\theta_{\lambda}\|=O(\|\theta_0\|)$:
    \begin{align*}
        \delta(n,\beta) 
        \leq~& \frac{1}{2} \beta^\top\E[ZZ^\top]\beta +  O\left(\sqrt{\frac{\log(p_X\cdot p_Z / \delta)}{n}} (\sqrt{\beta^\top\E[ZZ^\top]\beta} (1 + \|\theta_0\|)\right)
    \end{align*}
    Applying the AM-GM inequality to the second term on the right hand side, we can further bound:
    \begin{align*}
        \delta(n,\beta)\leq~& \frac{3}{4} \beta^\top\E[ZZ^\top]\beta +  O\left(\frac{\log(p_X\cdot p_Z / \delta)}{n} (1 + \|\theta_0\|)^2 \right)
    \end{align*}
    Letting $\kappa(n) = O\left(\frac{\log(p_X\cdot p_Z / \delta)}{n} (1 + \|\theta_0\|)^2 \right)$, we can bound $A$ as:
    \begin{align*}
    A \leq~& \max_{\beta} 2\beta^\top \E[Z(Y- X^\top\theta_\lambda)] - \frac{1}{4} \beta^\top \E[ZZ^\top] \beta + \zeta(n) + \kappa(n) + r(n)
    \end{align*}
    Solving for the optimization problem, we derive an optimal solution of:
    \begin{align*}
        \max_{\beta} 2\beta^\top \E[Z(Y- X^\top\theta_\lambda)] - \frac{1}{4} \beta^\top \E[ZZ^\top] \beta = 4 \|\tilde{\Sigma}_0(\theta_0 - \theta_\lambda)\|^2
    \end{align*}
    where we used that $\theta_0$ satisfies $\E[ZY]=\E[ZX^\top]\theta_0$. 

    Overall, we have derived that:
    \begin{align*}
        \|\tilde{\Sigma}_0 (\theta_0 - \hat{\theta})\|^2 \leq 8 \|\tilde{\Sigma}_0(\theta_0 - \theta_\lambda)\|^2 + 2\zeta(n) + 2\kappa(n) + 2r(n)
    \end{align*}
\end{proof}

\subsubappendix{Asymptotic Linearity of Estimate}\label{app:iv-linearity}

We prove that the error of the estimate $\hat{\theta}$, when projected onto the range of the covariance $\Sigma_0$, i.e. the quantity $\Sigma_0 (\hat{\theta}-\theta_0)$, satisfies an asymptotic linearity property. Note that this does not necessarily imply that $\hat{\theta}-\theta_0$ is asymptotically linear, since $\Sigma_0$ is not necessarily invertible (in our application, it is most definitely rank-deficient).

\begin{theorem}
    Assume that $\E[ZZ^\top]$ is full rank. Let $\Sigma_0 = \E[ZX^\top]$, $\tilde{\Sigma}_0 = \E[ZZ^\top]^{-1/2}\E[ZX^\top]$, and $\theta_0$ the minimum norm solution (i.e., $\theta_0 = \tilde{\Sigma}_0^+ \E[ZY]$) to the IV problem $\E[ZZ^\top]^{-1/2}\E[Z(Y - \theta^\top  X)]=0$. Assume that $X, Z, Y$ are uniformly and absolutely bounded. Given $n$ samples of $(Z,X,Y)$, let $\hat{\theta}$ be the solution to the adversarial IV method described in Algorithm~\ref{alg:adviv}, with $\lambda=\alpha/n=o(n^{-1/2})$ and $\lambda=\omega(n^{-1})$. Then $\hat{\theta}$ satisfies the following asymptotic linearity property:
    \begin{align}
        \sqrt{n} \Sigma_0 (\hat{\theta}-\theta_0) = \sqrt{n} \E[ZZ^\top]^{1/2}\, \tilde{P}\, \E[ZZ^\top]^{-1/2} \E_n\left[Z(Y - X^\top  \theta_0)\right] + o_p(1)
    \end{align}
    where $\tilde{P}=\tilde{\Sigma}_0 \tilde{\Sigma}_0^+$ is the orthogonal projector onto the range of $\tilde{\Sigma}_0$.
\end{theorem}

\begin{proof}
Recall from Appendix \ref{app:adviv} that $\hat{\theta}$ solves the empirical moment equation 
\begin{align}
 \E_n[\hat{Q}\,(Y - X^\top  \theta)] + \lambda \theta = 0  
\end{align}
where
\begin{align}
    \hat{Q} = \hat{B}^\top Z = (\E_n[ZZ^\top ]^+ \E_n[ZX^\top ])^\top Z = \E_n[XZ^\top] \E_n[ZZ^\top ]^+ Z
\end{align}

Then, using the fact that $\Sigma_0 \theta_0 = E[ZY]$, we have that 
\begin{align}
    \Sigma_0 (\hat{\theta} - \theta_0) = \Sigma_0 \hat{\theta} - \Sigma_0 \theta_0 = \Sigma_0 \hat{\theta} - \E[Z Y] = \E[Z(X^\top  \hat{\theta} - Y)]
\end{align}
Thus for any matrix $B$ and hence for the projection matrix $\hat{B}$ (of $Z$ onto $X$):
\begin{align}
    \hat{B}^\top  \Sigma_0 (\hat{\theta}-\theta_0) 
    =~& \E[\hat{B}^\top Z(X^\top  \hat{\theta} - Y)] \\
    =~&\E[\hat{Q}(X^\top  \hat{\theta} - Y)]\\
    =~& \E[\hat{Q}(X^\top  \hat{\theta} - Y)] - \E_n[\hat{Q}(X^\top  \hat{\theta} - Y)] - \lambda \hat{\theta}
\end{align}
since $\E_n[\hat{Q}(X^\top  \hat{\theta} - Y)] + \lambda\hat{\theta}$ = 0.
If we choose $\lambda = \alpha / n = o_p(n^{-1/2})$, then we have $\sqrt{n}\lambda = o_p(1)$. By Theorem~\ref{thm:rates}, and since by our assumption $\lambda = o(1)$ and $n\lambda = \omega(1) \to \infty$, we know that $\|\hat{\theta}-\theta_0\| = o_p(1)$ and therefore $\|\hat{\theta}\|\leq \|\theta_0\| + \|\hat{\theta}-\theta_0\|=O_p(\|\theta_0\|)=O_p(1)$. Thus, we have that $\sqrt{n} \lambda \|\hat{\theta}\| = o_p(1)$. Thus:
\begin{align}
    \sqrt{n} \hat{B}^\top  \Sigma_0 (\hat{\theta}-\theta_0) 
    =~& \sqrt{n} \E[\hat{Q}(X^\top  \hat{\theta} - Y)] - \sqrt{n}\E_n[\hat{Q}(X^\top  \hat{\theta} - Y)] + \sqrt{n}\lambda \hat{\theta}\\
    =~& \sqrt{n} \E[\hat{Q}(X^\top  \hat{\theta} - Y)] - \sqrt{n}\E_n[\hat{Q}(X^\top  \hat{\theta} - Y)] + o_p(1)\\
    =~& \sqrt{n} (\E-\E_n)\left[\hat{Q}(X^\top  \hat{\theta} - Y)\right] + o_p(1)\\
    =~& \sqrt{n} (\E-\E_n)\left[\hat{Q}(X^\top  \hat{\theta} - Y)\right] + \\ ~& (\sqrt{n} (\E-\E_n)\left[\hat{Q}X^\top  \theta_0\right] - \sqrt{n} (\E-\E_n)\left[\hat{Q}X^\top  \theta_0\right]) + o_p(1)\\
    =~& \sqrt{n} (\E-\E_n)\left[\hat{Q}(X^\top  \theta_0 - Y)\right] + \sqrt{n} (\E-\E_n)\left[\hat{Q}X^\top \right](\hat{\theta} - \theta_0) + o_p(1)
\end{align}
Note that the term $\sqrt{n}(\E-\E_n)[\hat{Q} X^\top ] =\hat{B}^\top \sqrt{n}(\E-\E_n)[ZX^\top ]=O_p(1)$. Moreover,
since the regularized adversarial IV estimate converges to the minimum norm solution, we have $\|\hat{\theta}-\theta_0\|=o_p(1)$. Thus we have $\sqrt{n} (\E-\E_n)[\hat{Q}X^\top](\hat{\theta}-\theta_0)=o_p(1)$. This leads to:
\begin{align*}
    \sqrt{n} \hat{B}^\top  \Sigma_0 (\hat{\theta}-\theta_0) 
    =~& \sqrt{n} (\E-\E_n)\left[\hat{Q}(X^\top  \theta_0 - Y)\right] + o_p(1)\\
    =~& \sqrt{n} \hat{B}^\top (\E-\E_n)\left[Z(X^\top  \theta_0 - Y)\right] + o_p(1)
\end{align*}
Moreover, since $\E[ZZ^\top]$ is full rank, then by the continuity of the inverse and the law of large numbers, $\hat{B}=\E_n[XZ^\top] \E_n[ZZ^\top]^{-1}$ converges in probability to the true projection matrix $B=\E[XZ^\top ] \E[ZZ^\top ]^{-1}$. Using the fact that $\E[Z(Y-\theta_0^\top X)] = 0$,  we can also conclude that
\begin{align*}
    \sqrt{n} B^\top  \Sigma_0 (\hat{\theta}-\theta_0) 
    =~& \sqrt{n} B^\top  (\E-\E_n)\left[Z(X^\top  \theta_0 - Y)\right] + o_p(1)\\
    =~& \sqrt{n} B^\top  (-\E_n)\left[Z(X^\top  \theta_0 - Y)\right] + o_p(1)\\
    =~& \sqrt{n} B^\top  \E_n\left[Z(Y - X^\top  \theta_0)\right] + o_p(1)
\end{align*}
Since $\Sigma_0^\top \E[ZZ^\top ]^{-1}= B^\top$, we derive that:
\begin{align*}
    \sqrt{n} \Sigma_0^\top  \E[ZZ^\top ]^{-1} \Sigma_0 (\hat{\theta}-\theta_0) = \sqrt{n} \Sigma_0^\top  \E[ZZ^\top]^{-1} \E_n[Z(Y - X^\top  \theta_0)] + o_p(1)
\end{align*}
Let $\tilde{\Sigma}_0 = \E[ZZ^\top ]^{-1/2}\Sigma_0$. Then we can re-write the above as:
\begin{align*}
    \sqrt{n} \tilde{\Sigma}_0^\top  \tilde{\Sigma}_0 (\hat{\theta}-\theta_0) = \sqrt{n} \tilde{\Sigma_0}^\top \E[ZZ^\top]^{-1/2}  \E_n[Z(Y - X^\top  \theta_0)] + o_p(1)
\end{align*}
Let the SVD of $\tilde{\Sigma}_0=UDV^\top $, and then $\tilde{\Sigma}_0^\top  \tilde{\Sigma}_0=VD^2V^\top $ and $(\tilde{\Sigma}_0^\top )^+=UD^{+}V^\top$. Thus $(\tilde{\Sigma}_0^\top )^+ \tilde{\Sigma}_0^\top  \tilde{\Sigma}_0= U D^+V^\top  VD^2V^\top  = UD^+D^2 V^\top =UDV^\top =\tilde{\Sigma}_0$. Thus left-multiplying the above equation with $(\tilde{\Sigma}_0^\top )^+$, we get:
\begin{align*}
    \sqrt{n} \tilde{\Sigma}_0 (\hat{\theta}-\theta_0) = \sqrt{n} (\tilde{\Sigma}_0^\top )^+\tilde{\Sigma}_0^\top  \E[ZZ^\top]^{-1/2} \E_n[Z(Y - X^\top  \theta_0)] + o_p(1)
\end{align*}
Moreover, $(\tilde{\Sigma}_0^\top )^+\tilde{\Sigma}_0^\top  = U D^+ V^\top  V D U^\top =U D^+ D U^\top  = \tilde{\Sigma}_0 \tilde{\Sigma}_0^+$. Thus we can equivalently write:
\begin{align*}
    \sqrt{n} \tilde{\Sigma}_0 (\hat{\theta}-\theta_0) = \sqrt{n} \tilde{\Sigma}_0 \tilde{\Sigma}_0^+ \E[ZZ^\top]^{-1/2} \E_n[Z(Y - X^\top  \theta_0)] + o_p(1)
\end{align*}
Since $\tilde{\Sigma}_0=\E[ZZ^\top]^{-1/2} \Sigma_0$, we can left-multiply both sides by $\E[ZZ^\top ]^{1/2}$ to get that:
\begin{align*}
    \sqrt{n} \Sigma_0 (\hat{\theta}-\theta_0) 
    =~& \sqrt{n} \E[ZZ^\top ]^{1/2} \tilde{\Sigma}_0\, \tilde{\Sigma}_0^+  \E[ZZ^\top ]^{-1/2}\E_n\left[Z(Y - X^\top  \theta_0)\right] + o_p(1)
\end{align*}
\end{proof}

\appendix{Data}\label{appendix:data}
All UK Biobank data can be explored in their accessible online search engine: \url{https://biobank.ndph.ox.ac.uk/showcase/search.cgi}. 

For preprocessing our data, we compress all variables over time to a single feature. (Although the UK Biobank data is primarily static, there are a few follow-up analyses on a small subset of the original population, and thus some variables are collected multiple times.) We treat all categorical variables as multiple-categorical variables, i.e., a patient could check off multiple categories for a single feature. For $W, Z,$ and $X$, categorical features are multi-hot encoded and zero-meaned (i.e., scaled to -0.5 or 0.5), and continuous features are standardized. 

We additionally ran into moderate levels of missingness for a few features, a common problem in structured healthcare data. We left categorical data unchanged (as a value of zero in categorical data indicates both missing and unobserved) as we did not find high levels of missingness in these variables. For continuous data, we mean-imputed missing values. 

\subappendix{List of variables for $W,Z,X$ }\label{appendix:wzx}
\subsubappendix{$W$ variables}
For the sociodemographic confounders $W$, we used 34 features, which, after multi-hot encoding, resulted in 171 total binary and continuous variables that are listed in Table \ref{table:w}. Note that we include all possible patient attributes $D$ in $W$ except the current attribute we are analyzing for. 

\begin{table}[H]
\caption{$W$ features}\label{table:w}
\begin{tabular}{@{}llc@{}}
\toprule
\textbf{Feature} & \textbf{Variable Type} & \textbf{Cardinality} \\ \midrule
Age when attended assessment centre & Continuous & 1 \\
Alcohol intake frequency. & Categorical & 7 \\
Attendance/disability/mobility allowance & Categorical & 6 \\
Average total household income before tax & Categorical & 7 \\
Current employment status & Categorical & 9 \\
Distance between home and job workplace & Continuous & 1 \\
Duration of moderate activity & Continuous & 1 \\
Frequency of travelling from home to job workplace & Continuous & 1 \\
Gas or solid-fuel cooking/heating & Categorical & 6 \\
Hand grip strength (left) & Continuous & 1 \\
Hand grip strength (right) & Continuous & 1 \\
Heating type(s) in home & Categorical & 9 \\
How are people in household related to participant & Categorical & 9 \\
Job involves heavy manual or physical work & Categorical & 6 \\
Job involves mainly walking or standing & Categorical & 6 \\
Job involves night shift work & Categorical & 6 \\
Job involves shift work & Categorical & 6 \\
Length of time at current address & Continuous & 1 \\
Length of working week for main job & Continuous & 1 \\
Number in household & Continuous & 1 \\
Number of vehicles in household & Categorical & 7 \\
Own or rent accommodation lived in & Categorical & 8 \\
Private healthcare & Categorical & 6 \\
Education level & Categorical & 8 \\
Race & Categorical & 2 \\
Sex & Categorical & 1 \\
Smoking status & Categorical & 4 \\
Standing height & Continuous & 1 \\
Time employed in main current job & Continuous & 1 \\
Transport type for commuting to job workplace & Categorical & 6 \\
Type of accommodation lived in & Categorical & 7 \\
Types of physical activity in last 4 weeks & Categorical & 7 \\
UK Biobank assessment centre & Categorical & 26 \\
Weight & Continuous & 1 \\ \bottomrule
\end{tabular}
\end{table}
\newpage
\subsubappendix{$X$ variables}

For the outcome proxies $X$, we used 65 continuous variables representing several different biomarkers. These listed in Table \ref{table:x}.
% \usepackage{booktabs}
\begin{table}[H]
\centering
\caption{$X$ features}\label{table:x}
\begin{tabular}{@{}llc@{}}
\toprule
\textbf{Feature} & \textbf{Variable Type} & \textbf{Cardinality} \\ \midrule
Alanine aminotransferase & Continuous & 1 \\
Albumin & Continuous & 1 \\
Alkaline phosphatase & Continuous & 1 \\
Apolipoprotein A & Continuous & 1 \\
Apolipoprotein B & Continuous & 1 \\
Aspartate aminotransferase & Continuous & 1 \\
Basophill count & Continuous & 1 \\
Basophill percentage & Continuous & 1 \\
C-reactive protein & Continuous & 1 \\
Calcium & Continuous & 1 \\
Cholesterol & Continuous & 1 \\
Creatinine & Continuous & 1 \\
Cystatin C & Continuous & 1 \\
Diastolic blood pressure, automated reading & Continuous & 1 \\
Diastolic blood pressure, manual reading & Continuous & 1 \\
Direct bilirubin & Continuous & 1 \\
Eosinophill count & Continuous & 1 \\
Eosinophill percentage & Continuous & 1 \\
Gamma glutamyltransferase & Continuous & 1 \\
Glucose & Continuous & 1 \\
Glycated haemoglobin (HbA1c) & Continuous & 1 \\
HDL cholesterol & Continuous & 1 \\
Haematocrit percentage & Continuous & 1 \\
Haemoglobin concentration & Continuous & 1 \\
High light scatter reticulocyte count & Continuous & 1 \\
High light scatter reticulocyte percentage & Continuous & 1 \\
IGF-1 & Continuous & 1 \\
Immature reticulocyte fraction & Continuous & 1 \\
LDL direct & Continuous & 1 \\
 \bottomrule
\end{tabular}
\end{table}

\begin{table}[H]
\centering
\caption{$X$ features, continued}
\begin{tabular}{@{}llc@{}}
\toprule
\textbf{Feature} & \textbf{Variable Type} & \textbf{Cardinality} \\ \midrule
Lipoprotein A & Continuous & 1 \\
Lymphocyte count & Continuous & 1 \\
Lymphocyte percentage & Continuous & 1 \\
Mean corpuscular haemoglobin & Continuous & 1 \\
Mean corpuscular haemoglobin concentration & Continuous & 1 \\
Mean corpuscular volume & Continuous & 1 \\
Mean platelet (thrombocyte) volume & Continuous & 1 \\
Mean reticulocyte volume & Continuous & 1 \\
Mean sphered cell volume & Continuous & 1 \\
Monocyte count & Continuous & 1 \\
Monocyte percentage & Continuous & 1 \\
Neutrophill count & Continuous & 1 \\
Neutrophill percentage & Continuous & 1 \\
Nucleated red blood cell count & Continuous & 1 \\
Nucleated red blood cell percentage & Continuous & 1 \\
Phosphate & Continuous & 1 \\
Platelet count & Continuous & 1 \\
Platelet crit & Continuous & 1 \\
Platelet distribution width & Continuous & 1 \\
Pulse rate (during blood-pressure measurement) & Continuous & 1 \\
Pulse rate, automated reading & Continuous & 1 \\
Red blood cell (erythrocyte) count & Continuous & 1 \\
Red blood cell (erythrocyte) distribution width & Continuous & 1 \\
Reticulocyte count & Continuous & 1 \\
Reticulocyte percentage & Continuous & 1 \\
Rheumatoid factor & Continuous & 1 \\
SHBG & Continuous & 1 \\
Systolic blood pressure, automated reading & Continuous & 1 \\
Systolic blood pressure, manual reading & Continuous & 1 \\
Total bilirubin & Continuous & 1 \\
Total protein & Continuous & 1 \\
Triglycerides & Continuous & 1 \\
Urate & Continuous & 1 \\
Urea & Continuous & 1 \\
Vitamin D & Continuous & 1 \\
White blood cell (leukocyte) count & Continuous & 1 \\ \bottomrule
\end{tabular}
\end{table}
\newpage
\subsubappendix{$Z$ variables}

For the proxies $Z$, we used 67 features, which, after multi-hot encoding, equals 196 total binary and continuous variables, and are listed in Table \ref{table:z}.

\begin{table}[H]
\caption{$Z$ features}\label{table:z}
\resizebox{.8\textwidth}{!}{%
\begin{tabular}{@{}llc@{}}
\toprule
\textbf{Feature} & \textbf{Variable Type} & \textbf{Cardinality} \\ \midrule
Back pain for 3+ months & Categorical & 1 \\
Bipolar and major depression status & Categorical & 6 \\
Bipolar disorder status & Categorical & 2 \\
Chest pain due to walking ceases when standing still & Categorical & 1 \\
Chest pain or discomfort & Categorical & 1 \\
Chest pain or discomfort walking normally & Categorical & 2 \\
Chest pain or discomfort when walking uphill or hurrying & Categorical & 2 \\
Daytime dozing / sleeping (narcolepsy) & Categorical & 4 \\
Ever depressed for a whole week & Categorical & 1 \\
Ever highly irritable/argumentative for 2 days & Categorical & 1 \\
Ever manic/hyper for 2 days & Categorical & 1 \\
Ever unenthusiastic/disinterested for a whole week & Categorical & 1 \\
Facial pains for 3+ months & Categorical & 1 \\
Fed-up feelings & Categorical & 1 \\
Frequency of depressed mood in last 2 weeks & Categorical & 4 \\
Frequency of tenseness / restlessness in last 2 weeks & Categorical & 4 \\
Frequency of tiredness / lethargy in last 2 weeks & Categorical & 4 \\
Frequency of unenthusiasm / disinterest in last 2 weeks & Categorical & 4 \\
General pain for 3+ months & Categorical & 1 \\
Getting up in morning & Categorical & 4 \\
Guilty feelings & Categorical & 1 \\
Headaches for 3+ months & Categorical & 1 \\
Hip pain for 3+ months & Categorical & 1 \\
Illness, injury, bereavement, stress in last 2 years & Categorical & 7 \\
Irritability & Categorical & 1 \\
Knee pain for 3+ months & Categorical & 1 \\
Length of longest manic/irritable episode & Categorical & 3 \\
Loneliness, isolation & Categorical & 1 \\
Longest period of depression & Continuous & 1 \\
Longest period of unenthusiasm / disinterest & Continuous & 1 \\
Manic/hyper symptoms & Categorical & 6 \\
Miserableness & Categorical & 1 \\
Mood swings & Categorical & 1 \\
Morning/evening person (chronotype) & Categorical & 4 \\
Nap during day & Categorical & 3 \\
Neck/shoulder pain for 3+ months & Categorical & 1 \\
Nervous feelings & Categorical & 1 \\
Neuroticism score & Continuous & 1 \\
Number of depression episodes & Continuous & 1 \\
Number of unenthusiastic/disinterested episodes & Continuous & 1 \\
Pain type(s) experienced in last month & Categorical & 9 \\
Probable recurrent major depression (moderate) & Categorical & 1 \\
Probable recurrent major depression (severe) & Categorical & 1 \\
Risk taking & Categorical & 1 \\
Seen a psychiatrist for nerves, anxiety, tension or depression & Categorical & 1 \\
Seen doctor (GP) for nerves, anxiety, tension or depression & Categorical & 1 \\
Sensitivity / hurt feelings & Categorical & 1 \\
Severity of manic/irritable episodes & Categorical & 2 \\
Single episode of probable major depression & Categorical & 1 \\
Sleep duration & Continuous & 1 \\
Sleeplessness / insomnia & Categorical & 3 \\
Snoring & Categorical & 1 \\
Stomach/abdominal pain for 3+ months & Categorical & 1 \\
Suffer from 'nerves' & Categorical & 1 \\
Tense / 'highly strung' & Categorical & 1 \\
Worrier / anxious feelings & Categorical & 1 \\
Worry too long after embarrassment & Categorical & 1 \\ \bottomrule
\end{tabular}%
}
\end{table}
\subappendix{List of full $D,Y$ pairs}\label{appendix:alldy}
We use eight $D$ attributes total, which were detailed in Table \ref{table:dy}: Race=Asian, Race=Black, Gender=Female, On disability allowance, Low income, No post-secondary education, Clinically considered Obese (BMI $>$ 30), and Not on private health insurance.\footnote{Note that the UK offers universal healthcare, and thus public vs. private insurance is less of a proxy for income as in the US.} 

In addition, we use thirteen $Y$ medical diagnoses, where we use ICD10 codes as proxies and where the binary label of a diagnosis is aggregated across all of a patient's UK synced hospital records to represent if diagnosis $Y$ was every given at any point.\footnote{Note that, unlike the US, ICD10 codes in the UK are not the primary source for billing information, and thus is likely a stronger proxy for diagnoses than using ICD10 codes in the US.} In addition to the seven diagnoses $Y$ mentioned in Table \ref{table:dy} (osteoarthritis, rheumatoid arthritis, chronic kidney disease, complications during labor, heart disease, depression, and melanoma), we also look at the diagnoses listed in Table \ref{table:y2}: back pain, fibromyalgia, migraines, endometriosis, and inflammatory bowel disease. 

In total, we look at 102 pairs (all $D$ attributes $\times$ all $Y$ attributes, except $D=$Female and $Y \in$ [endometriosis, compilations during labor]). 

\begin{table}[H]
\centering
\resizebox{.5\textwidth}{!}{%
\begin{tabular}{@{}llc@{}}
\toprule
\textbf{} & {} & \textbf{\begin{tabular}[c]{@{}c@{}}Prevalence in \\ UK Biobank \\ (n=502411)\end{tabular}} \\ \midrule
\textbf{Medical diagnosis $Y$} & Migraine & 1.7\% \\
                               & Inflammatory bowel disease & 1.5\% \\
                               & Back pain & 6.4\% \\
                               & Endometriosis & 2.0\% \\
                               & Fibromyalgia & 0.7\% \\ \bottomrule
\end{tabular}%
}
\caption{}
\label{table:y2}
\end{table}

\appendix{Additional results}
\subappendix{Semi-synthetic experiments}\label{appendix:semisynthetic_results}
\subsubappendix{Hyperparameters}
Recall our data generation process uses the linear structural equations:
\begin{align}
    M_\gen =~& a D_{\gen} + \hat{\epsilon}_M\\
    Y_{\gen} =& \frac{b}{K} \sum_{i=1}^K M_{i,\gen} + \theta D_{\gen} + g X_{0, \gen} + \hat{f}_Y(W_{\gen}) + \sigma_Y \hat{\epsilon}_Y
\end{align}
For each experiment, we select $a=1$, $b=1$, $g=0$, and $\sigma_Y=1.$ For the implicit bias effect $\theta_0$ we try $\theta_0 \in [-0.5, -0.1, 0.1, 0.5]$. We fit the propensity model $\E[D|W]$ with a logistic regression model using semi-cross fitting where the penalty parameter $\lambda$ is first chosen from cross-validation to minimize the out-of-sample error. To estimate $G$, $F$, and Cov($M$) = $\Sigma = \text{diagonal}(\sigma_1, ..., \sigma_K)$, we compute the SVD of Cov$(\tilde{X}_{\text{train}}, \tilde{Z}_{\text{train}})$ where $\tilde{V}_{\text{train}}$ denotes 50\% of the data from a train-test split. In addition, we keep only the non-zero singular values of Cov$(\tilde{X}_{\text{train}}, \tilde{Z}_{\text{train}})$, using our proposed covariance rank test from Appendix \ref{apendix:test:rank}, to generate Cov($M$). During inference, we use the data from the test split to estimate $\epsilon_X, \epsilon_Z, \epsilon_Y \sim \sigma_Y F_n(\tilde{Y}_{\text{test}})$ and generate new samples given the linear equations. 

All $V_{\gen}$ for $V\in[X,Z,Y]$ are sampled as continuous. To test on semi-synthetic data that is more realistic of mixed-type medical data, we convert the sampled data to binary by thresholding all features which are binary in the real data. Specifically, since only $Z$ contains binary data, if $b_Z$  are the indices of all real binary $Z$ features, we set  $Z_{\text{gen,binary}}[:,b_Z] =  Z_{\gen}[:,b_Z] > 0$.

We fit our structural equations once and then run our method over $K=100$ randomly generated semi-synthetic datasets of size $n=$50,000. 

For baselines, in addition to the two OLS models mentioned in Section \ref{main:semisyn}, we also consider adversarial IV estimation using the \eqref{eq:primal} where $(\tilde{D}_{\gen};\tilde{Z}_{\gen})$ are the instruments and $(\tilde{D}_{\gen};\tilde{X}_{\gen})$ are the treatments (note this is naturally run in our method when we estimate $h_*$; our method just "throws" out the computed $\theta$). We also ran LASSO regression using cross validation to choose the penalty parameter $\lambda$ but found very similar results to OLS and thus excluded for brevity. 

\subsubappendix{Evaluation metrics}
We report the following evaluation metrics:
\begin{itemize}
    \item Average point estimate $\bar{\theta} = \frac{1}{K} \sum_i \theta_i$ over the $K=100$ iterations
    \item Confidence interval (CI) = $\bar{\theta}\pm1.96\sigma$, where $\sigma$ is the standard deviation over all estimates $\theta_i$
        \item Average CI = $\frac{1}{K}  \sum_i 1.96\sigma_i$
\item Coverage, or the percentage of the $K$ iterations $\theta_i\pm1.96\sigma_i$ that contains the true $\theta_0$
    \item RMSE (root mean squared error) = $\sqrt{\frac{1}{K} \sum_i (\theta_i - \theta_0)^2}$
    \item Bias = $ \abs{\frac{1}{K} \sum_i (\theta_i - \theta_0)}$
    \item Average success rate of our five tests proposed in Section  \ref{main:tests}
\end{itemize}

\subsubappendix{Results}
We report results for the mixed-type synthetic data, i.e., continuous and binary. In Table \ref{table:synthbase}, we test $\theta_0 \in [-0.5, -0.1, 0.1, 0.5]$ and compare our method to the baselines. In Table \ref{table:synthmetrics} we report the aforementioned evaluation metrics of our method over different $\theta_0$.
\begin{table}[H]
\centering
\resizebox{.7\textwidth}{!}{%
{\begin{tabular}{@{}ccccc@{}}\toprule
 \textbf{$\theta_0$} & \textbf{Our method} & \textbf{OLS($D, W, M, X$)} & \textbf{OLS($D, W, Z, X$)} & \textbf{AdvIV($(\tilde{D};\tilde{Z}), (\tilde{D};\tilde{X})$}\\ \colrule
  -0.5 & -0.46 $\pm$ 0.003  & -0.5 $\pm$ 0.01  & 0.38 $\pm$ 0.01 & -0.39 $\pm$ 0.002  \\
 -0.1 & -0.07 $\pm$ 0.003  & -0.1 $\pm$ 0.01  & 0.78 $\pm$ 0.01 & 0.00 $\pm$ 0.002 \\ 
  0.1 & 0.13 $\pm$ 0.003  & 0.1 $\pm$ 0.01  & 0.98 $\pm$ 0.01 & 0.19 $\pm$ 0.002 \\   
  0.5 & 0.53 $\pm$ 0.003  & 0.5 $\pm$ 0.01  & 1.38 $\pm$ 0.01  & 0.58 $\pm$ 0.002
\\ \botrule
\end{tabular}}
}
\caption{Semi-synthetic data: our method versus baselines, reporting $\bar{\theta}\pm1.96\sigma$ over $K=100$ iterations for mixed-type data, i.e., continuous and binary.}
\label{table:synthbase}
\end{table}

% \usepackage{booktabs}
\begin{table}[ht]
\centering
\resizebox{1.1\textwidth}{!}{%
\begin{tabular}{@{}ccccccccccc@{}}
\toprule
\textbf{$\theta_0$} & \textbf{$\bar{\theta}\pm1.96\sigma$} & \textbf{$\pm1.96\sigma_i$} & \textbf{Coverage} & \textbf{RMSE} & \textbf{Bias} & \multicolumn{1}{c}{\textbf{\begin{tabular}[c]{@{}c@{}}(1) \% Passing\\ Primal \end{tabular}}} & \multicolumn{1}{c}{\textbf{\begin{tabular}[c]{@{}c@{}}(2) \% Passing\\ Dual \end{tabular}}} & \multicolumn{1}{c}{\textbf{\begin{tabular}[c]{@{}c@{}}(3) \% Passing\\ $\E[\tilde{D}V]\neq 0$ \end{tabular}}} & \multicolumn{1}{c}{\textbf{\begin{tabular}[c]{@{}c@{}}(4) \% Passing\\ $V$ strength F-test \end{tabular}}} & \multicolumn{1}{c}{\textbf{\begin{tabular}[c]{@{}c@{}}(5) \% Passing\\ Cov($\tilde{X},\tilde{Z})$ rank test \end{tabular}}} \\ \midrule
-0.5 & -0.46 $\pm$ 0.0 & 0.07 & 0.78 & 0.05 & 0.04 & 91\%  &94\%  & 100\% & 100\% & 100\%\\
-0.1 & -0.07 $\pm$ 0.0 &0.07 & 0.80 & 0.05 & 0.03 & 93\%  &94\%  & 100\% & 100\% & 100\% \\
0.1 & 0.13 $\pm$ 0.0 & 0.07 & 0.82 & 0.05 & 0.03 & 95\%  & 94\%  & 100\% & 100\% & 100\% \\
0.5 & 0.53 $\pm$ 0.0 & 0.07 & 0.85 & 0.05 & 0.03 & 95\%  & 94\%  & 100\% & 100\% & 100\% \\ \bottomrule
\end{tabular}
}
\caption{Semi-synthetic data: evaluation metrics on our method over $K=100$ iterations for mixed-type data, i.e., continuous and binary.}
\label{table:synthmetrics}
\end{table}

\subappendix{UK Biobank effect estimates using all $X,Z$ proxy data}\label{appendix:7metrics_allXZ}
In Table \ref{table:allxz} we show the results of our method on 20 $(D,Y)$ pairs using all of the $X,Z$ as the proxies for $M$. We show the effect estimate $\hat{\theta}$, 95\% confidence interval, and the results from the first four tests we proposed. Note we exclude the covariance rank test; since we are using the same $X,Z$ for all $(D,Y)$ pairs, we only need to compute the test once and note that it passes with four statistically-significant non-zero singular values. 

We see that, while the identification tests pass, both the dual and the primal violation tests fail for all $(D,Y)$ pairs. This invalidates any estimates $\hat{\theta}$. % Please add the following required packages to your document preamble:
\begin{table}[H]
\centering
\resizebox{.9\textwidth}{!}{%
\begin{tabular}{@{}cccccc@{}}
\toprule
\multicolumn{1}{c}{\textbf{$(D,Y)$}} & \multicolumn{1}{c}{\textbf{$\theta\pm95$\% CI}} & \multicolumn{1}{c}{\textbf{\begin{tabular}[c]{@{}c@{}}(1) Primal \\ statistic $<$ critical  \end{tabular}}} & \multicolumn{1}{c}{\textbf{\begin{tabular}[c]{@{}c@{}}(2) Dual \\ statistic $<$ critical  \end{tabular}}} & \multicolumn{1}{c}{\textbf{\begin{tabular}[c]{@{}c@{}}(3) $\E[\tilde{D}V] \neq 0$\\ statistic $>$ critical  \end{tabular}}} & \multicolumn{1}{c}{\textbf{\begin{tabular}[c]{@{}c@{}}(4) $V$ strength F-test \\ statistic $>$ critical  \end{tabular}}} \\ \midrule
Black, Osteoarthritis & 0.18$\pm$0.24 & 422.8$\nless$230.7 & 144.1$\nless$84.8 & 7.0\textgreater{}0.3 & 37.8\textgreater{}23.1 \\
Black, Heart disease & -0.03$\pm$0.13 & 243.4$\nless$230.7 & 144.1$\nless$84.8 & 7.0\textgreater{}0.3 & 37.8\textgreater{}23.1 \\
Black, Depression & 0.48$\pm$0.2 & 417.9$\nless$230.7 & 144.1$\nless$84.8 & 7.0\textgreater{}0.3 & 37.8\textgreater{}23.1 \\
Black, Back pain & -0.43$\pm$0.13 & 344.5$\nless$230.7 & 144.1$\nless$84.8 & 7.0\textgreater{}0.3 & 37.8\textgreater{}23.1 \\
Black, Rh. Arthritis & 0.01$\pm$0.05 & 246.1$\nless$230.7 & 144.1$\nless$84.8 & 7.0\textgreater{}0.3 & 37.8\textgreater{}23.1 \\
Female, Osteoarthritis & -0.14$\pm$0.31 & 440.0$\nless$230.7 & 136.2$\nless$84.8 & 9.5\textgreater{}1.1 & 33.4\textgreater{}23.1 \\
Female, Heart disease & 0.4$\pm$0.18 & 248.9$\nless$230.7 & 136.2$\nless$84.8 & 9.5\textgreater{}1.1 & 33.4\textgreater{}23.1 \\
Female, Depression & 0.27$\pm$0.27 & 411.1$\nless$230.7 & 136.2$\nless$84.8 & 9.5\textgreater{}1.1 & 33.4\textgreater{}23.1 \\
Female, Back pain & 0.66$\pm$0.17 & 361.5$\nless$230.7 & 136.2$\nless$84.8 & 9.5\textgreater{}1.1 & 33.4\textgreater{}23.1 \\
Female, Rh. Arthritis & -0.04$\pm$0.06 & 248.1$\nless$230.7 & 136.2$\nless$84.8 & 9.5\textgreater{}1.1 & 33.4\textgreater{}23.1 \\
Obese, Osteoarthritis & -0.31$\pm$0.05 & 407.9$\nless$230.7 & 180.4$\nless$84.8 & 64.9\textgreater{}1.3 & 131.2\textgreater{}23.1 \\
Obese, Heart disease & -0.07$\pm$0.03 & 241.0$\nless$230.7 & 180.4$\nless$84.8 & 64.9\textgreater{}1.3 & 131.2\textgreater{}23.1 \\
Obese, Depression & -0.06$\pm$0.04 & 403.7$\nless$230.7 & 180.4$\nless$84.8 & 64.9\textgreater{}1.3 & 131.2\textgreater{}23.1 \\
Obese, Back pain & -0.09$\pm$0.02 & 354.9$\nless$230.7 & 180.4$\nless$84.8 & 64.9\textgreater{}1.3 & 131.2\textgreater{}23.1 \\
Obese, Rh. Arthritis & -0.04$\pm$0.01 & 249.4$\nless$230.7 & 180.4$\nless$84.8 & 64.9\textgreater{}1.3 & 131.2\textgreater{}23.1 \\
Asian, Osteoarthritis & 0.04$\pm$0.1 & 438.9$\nless$230.7 & 157.1$\nless$84.8 & 10.5\textgreater{}0.3 & 40.4\textgreater{}23.1 \\
Asian, Heart disease & -0.03$\pm$0.06 & 243.5$\nless$230.7 & 157.1$\nless$84.8 & 10.5\textgreater{}0.3 & 40.4\textgreater{}23.1 \\
Asian, Depression & 0.11$\pm$0.09 & 424.3$\nless$230.7 & 157.1$\nless$84.8 & 10.5\textgreater{}0.3 & 40.4\textgreater{}23.1 \\
Asian, Back pain & 0.0$\pm$0.05 & 324.3$\nless$230.7 & 157.1$\nless$84.8 & 10.5\textgreater{}0.3 & 40.4\textgreater{}23.1 \\
Asian, Rh. Arthritis & -0.01$\pm$0.02 & 254.1$\nless$230.7 & 157.1$\nless$84.8 & 10.5\textgreater{}0.3 & 40.4\textgreater{}23.1 \\ \bottomrule
\end{tabular}
}
\caption{Results using all $X,Z$ proxies (no proxy selection algorithm). Note that the primal and dual tests fail, as the statistic is never $<$ the critical value, as is needed to pass.}
\label{table:allxz}
\end{table}

\subappendix{Discovery of invalid proxies and selection of non-violating proxy subsets}\label{appendix:invalidproxies}
As see in the results in Table \ref{table:allxz}, using all $X,Z$ proxies leads to the violation of the primal and dual tests, regardless of the $(D,Y)$ pair. Given the covariance rank test is still passing, we hypothesized this is due to the existence of some subset of $Z$ proxies (say, $Z'$) that have a causal path to $Y$ that isn't through $M$, and a subset of $X$ proxies (say, $X'$) that have a causal path to $Y$ that isn't $M$. We provide an example of the hypothesized causal graph of this scenario in Figure \ref{cg:vio}, where we remove confounders $W$ for clarity. 

Given this hypothesis, we altered our data in two ways that proved integral to retrieving statistically significant estimates. First, we removed all $Z$ proxy binary features indicating a survey response as 'Do not know' or 'Prefer not to answer' (e.g., $Z_j$ representing the binary variable 'Back pain in last 3+ months = Do not know' would be removed). While similar ambiguous or non-responses can contain very real phenotypes in certain cases\cite{montagni2019don}, we observed they had little association with other $X$ proxies, $D$, or $Y$. Second, we developed a proxy selection algorithm (see the full description of this algorithm in Appendix \ref{appendix:proxyrm}) that finds an admissible set of proxy features $\{(\cX^{(i)}, \cZ^{(i)})\}_i$ as subsets of the original proxies such that we exclude any proxy features that the algorithm finds to be in violation of the assumptions. 

For the proxy variable selection algorithm hyperparameters, we set $K=150$, $\delta = 0.1$, and we run the algorithm for two iterations. 68 of the 102 $(D,Y)$ pairs find no admissible proxy sets.\footnote{Increasing $K$ or varying $\delta$ might produce an admissible set in this case, but we did not explore these further.} 
We found 34 $(D,Y)$ pairs that find one or more candidate proxy pairs $\{(\cX^{(i)}, \cZ^{(i)})\}_i$.

\subsubappendix{Interpretability into the proxy selection algorithm}
We next investigate which proxies were selected as admissible and if these proxies share any meaningful biological significance. We run several interpretability analysis:
\begin{itemize}
    \item For a given $(D,Y)$ pair, are the proxy candidates $\{(\cX^{(i)}, \cZ^{(i)})\}_i$ similar for all candidates $i$?
    \item Are the proxies biologically relevant to $D$ or $Y$?
    \item Is there a pattern which proxies are selected with respect to $D$ or $Y$?
\end{itemize}

First, we look for any consistency in the selected candidates for a given $(D,Y)$ pair that have multiple candidates selected. We find most, such as $D=$Female, $Y=$Heart disease, to have almost perfect consistency in $X$ and $Z$ proxies selected. Other pairs, however, such as $D=$Black, $Y=$Chronic kidney disease, had high variance in the proxy subsets chosen and warrants further investigation.

Next we examined if the selected proxies are biologically reasonable. We can sort $X$ (or $Z$) features based on how often the feature is selected into $\cX^{(i)} (or \cZ^{(i)}$) across all candidates $i$ for a given ($D,Y$) pair. Although $D=$Black, $Y=$Chronic kidney disease had high variance in candidate features selected, a qualitative analysis showed the features commonly selected were biologically reasonable. Two of the most select biomarker features of $X$, aspartate aminotransferase (AST) and alanine aminotransferase (ALT), are enzymes found commonly in healthy livers and are often biomarkers used to check for kidney disease. We also probed $Z$ proxies for biological significance but find most $(D,Y)$ pairs are consistent in which proxies are selected versus removed. For example, $Z$ proxies related to a patient's self-reported bipolar disorder status or daily lethargy consistently were selected into the $Z$ proxy subset, while reports of neck, shoulder, or chest pain were rarely selected. We give three examples of the most common proxy features selected in Table \ref{table:featselected}.
% \usepackage{booktabs}
\begin{table}[H]
\centering
\resizebox{1.1\textwidth}{!}{%
\begin{tabular}{@{}lllllll@{}}
\toprule
\textbf{($D,Y$)} & \textbf{Top 10 Z proxies selected} & \textbf{Top 10 X proxies selected} &  &  &  &  \\ \midrule
Disability Insurance, Rh. arthritis & \begin{tabular}[c]{@{}l@{}}Bipolar disorder status=Bipolar Type I (Mania),\\  Bipolar and major depression status=Single Probable major depression episode,\\  Manic/hyper symptoms=I needed less sleep than usual,\\  Manic/hyper symptoms=I was more creative or had more ideas than usual,\\  Manic/hyper symptoms=All of the above,\\  Frequency of depressed mood in last 2 weeks=Nearly every day,\\  Length of longest manic/irritable episode=A week or more,\\  Length of longest manic/irritable episode=Less than a week,\\  Illness, injury, bereavement, stress in last 2 years=Death of a spouse or partner,\\  Bipolar and major depression status=Probable Recurrent major depression (severe)\end{tabular} & \begin{tabular}[c]{@{}l@{}}White blood cell (leukocyte) count,\\ Albumin,\\  Aspartate aminotransferase,\\  Pulse rate,\\  Basophill percentage,\\  Immature reticulocyte fraction,\\  High light scatter reticulocyte percentage,\\ Monocyte percentage,\\  Nucleated red blood cell percentage,\\ Vitamin D\end{tabular} &  &  &  &  \\ \hline
Female, Heart disease & \begin{tabular}[c]{@{}l@{}}Frequency of tiredness / lethargy in last 2 weeks=Not at all,\\  Frequency of unenthusiasm / disinterest in last 2 weeks=Not at all,\\  Bipolar and major depression status=Bipolar II Disorder,\\  Bipolar and major depression status=Probable Recurrent major depression (severe),\\  Bipolar and major depression status=Probable Recurrent major depression (moderate),\\  Bipolar and major depression status=Single Probable major depression episode,\\  Loneliness, isolation=Yes,\\  Guilty feelings=Yes,\\  Frequency of depressed mood in last 2 weeks=Not at all,\\  Frequency of depressed mood in last 2 weeks=Several days\end{tabular} & \begin{tabular}[c]{@{}l@{}}Red blood cell (erythrocyte) count,\\ LDL direct,\\ Direct bilirubin,\\  Mean platelet (thrombocyte) volume, \\ Monocyte count,\\  C-reactive protein, \\ Gamma glutamyltransferase,\\  Mean sphered cell volume,\\ Aspartate aminotransferase,\\ Lipoprotein A\end{tabular} &  &  &  &  \\ \hline
Black, Chronic kidney disease & \begin{tabular}[c]{@{}l@{}}Bipolar and major depression status=Bipolar II Disorder,\\  Bipolar disorder status=Bipolar Type II (Hypomania),\\  Daytime dozing / sleeping (narcolepsy)=All of the time,\\  Bipolar and major depression status=Bipolar I Disorder,\\  Bipolar disorder status=Bipolar Type I (Mania),\\  Illness, injury, bereavement, stress in last 2 years=Death of a spouse or partner,\\  Manic/hyper symptoms=I needed less sleep than usual,\\  Chest pain or discomfort walking normally=Unable to walk on the level,\\  Single episode of probable major depression=Yes,\\ \end{tabular} & \begin{tabular}[c]{@{}l@{}}Total protein,\\  Pulse rate,\\ Aspartate aminotransferase,\\ Alanine aminotransferase, \\ SHBG,\\  Rheumatoid factor,\\  Urate,\\ Monocyte count,\\ IGF-1\end{tabular} &  &  &  &  \\ \bottomrule
\end{tabular}
}
\caption{Top 10 proxies selected for three $(D,Y)$ pairs that generate multiple proxy set candidates. $X$ and $Z$ features are sorted based on how often they were selected by each $\cX^{(i)}$ or $\cZ^{(i)}$, respectively, where $\cX^{(i)}$ is a candidate of admissible $X$ proxy features for a given $(D,Y)$ pair.}
\label{table:featselected}
\end{table}

\begin{figure}[ht]
\centerline{\includegraphics[width=\textwidth]{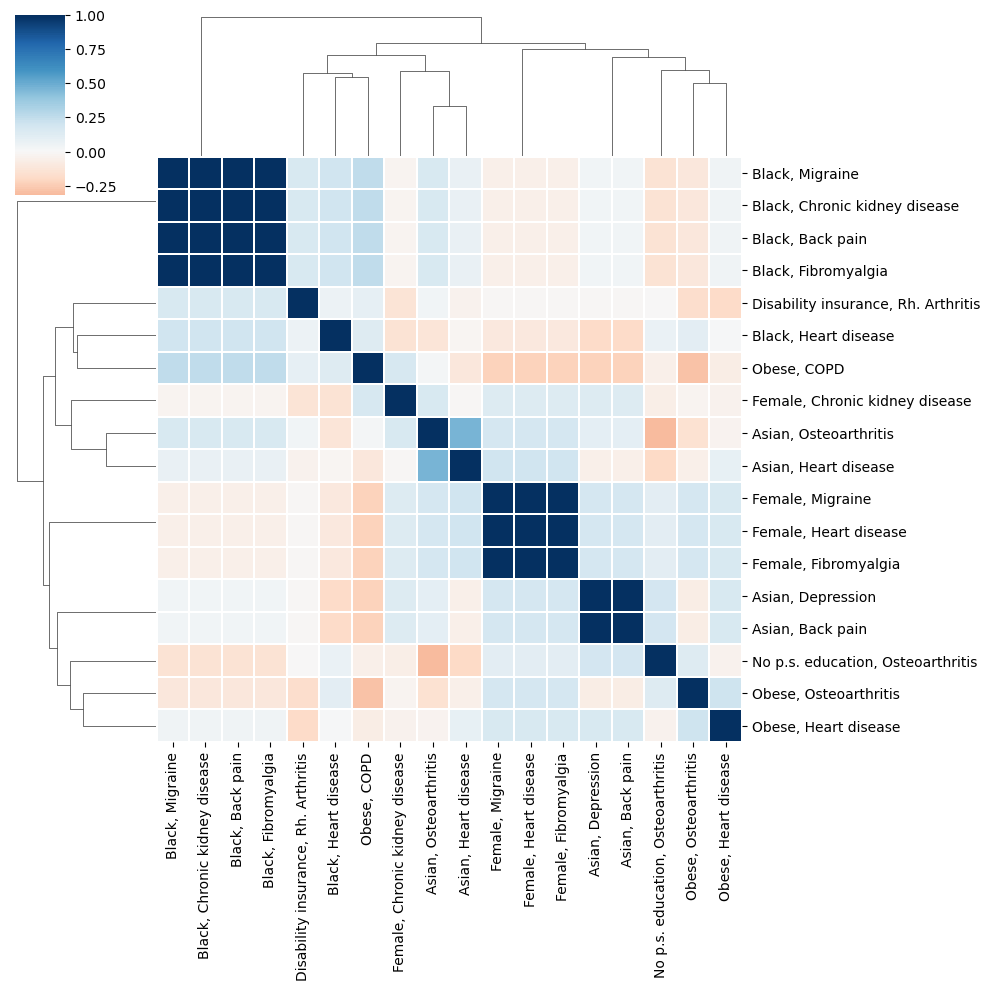}}
\caption{Correlation clustermap comparing proxy $X$ features selected by the proxy removal algorithm across $(D,Y)$ pairs that (1) yielded admissible $\{(\cX^{(i)}, \cZ^{(i)})\}_i$ candidates and (2) whose median point estimate $\mid \theta \mid > 0.05$. COPD = Chronic obstructive pulmonary disease. p.s. = post-secondary. }
\label{fig:heatx}
\end{figure}

\begin{figure}[ht]
\centerline{\includegraphics[width=\textwidth]{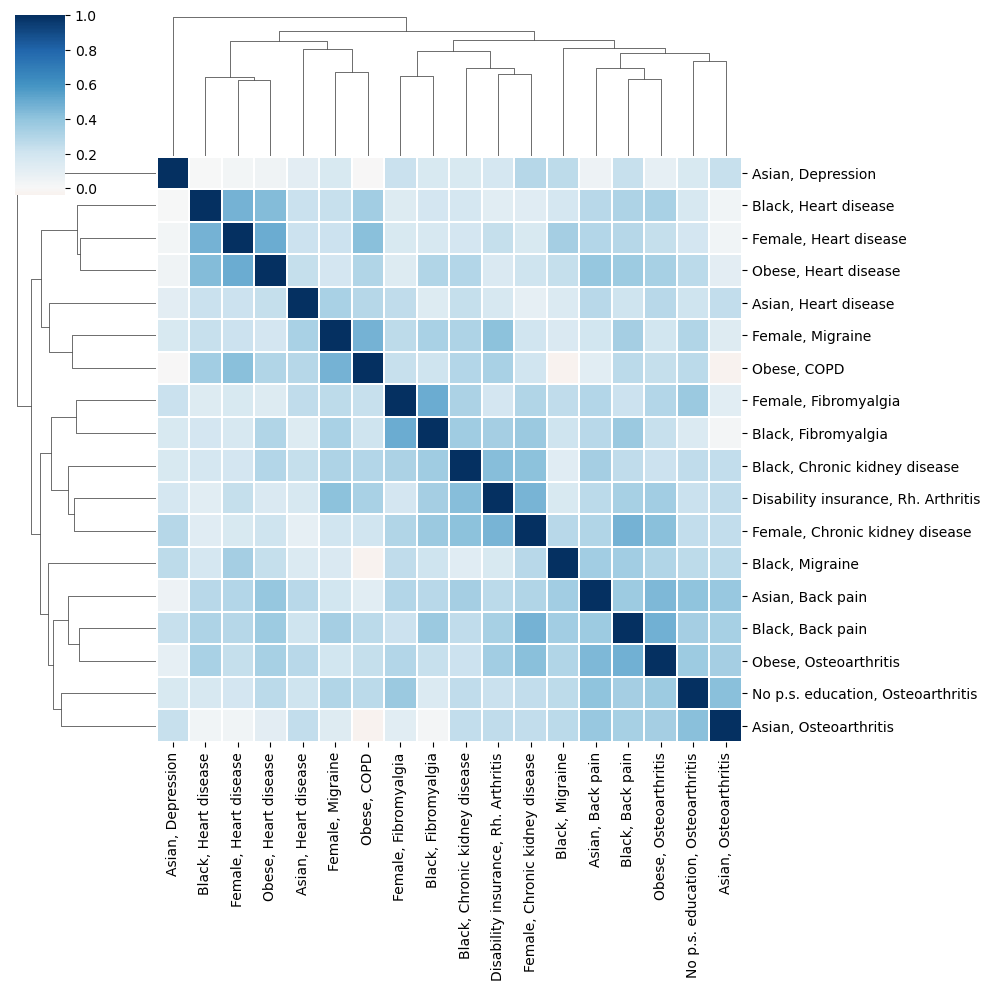}}
\caption{Correlation clustermap comparing proxy $Z$ features selected by the proxy removal algorithm across $(D,Y)$ pairs that (1) yielded admissible $\{(\cX^{(i)}, \cZ^{(i)})\}_i$ candidates and (2) whose median point estimate $\mid \theta \mid > 0.05$. COPD = Chronic obstructive pulmonary disease. p.s. = post-secondary.}
\label{fig:heatz}
\end{figure}

To select a single proxy feature set given a list of candidates for a $(D,Y)$ pair, we compute the final point estimate for all candidates $i$ in $\{(\cX^{(i)}, \cZ^{(i)})\}_i$ and we choose the median estimate and its corresponding proxy features. We then compare the proxies selected across $(D,Y)$ pairs by taking the correlation of the binary matrix indicating if a feature was selected as admissible for each $(D,Y)$ pair or not. To only consider cases where the proxies contributed to a non-zero bias effect being detected, we filter the $(D,Y)$ pairs if $\mid \theta \mid > 0.05$. 

In Tables \ref{fig:heatx} and \ref{fig:heatz} we show the correlation clustermap for $X$ and $Z$ features, respectively. We observe a strong similarity of $X$ proxies selected based on the attribute $D$. For example, the admissible proxy subsets $S_X$ for $D=$Black $X$ are almost identical. We believe this finding supports our hypothesis as depicted in Figure \ref{cg:vio} that there exists a violating path $D \rightarrow X'$, as our algorithm appears to remove the same violating $X'$ proxies for a specific attribute $D$. For selected proxies $Z$, there are less evident patterns of features selected across $D$ or $Y$, although the clustermap slighly indicates $Z$ proxies are selected according to $Y$ diagnoses. Instead, it appears that most $(D,Y)$ pairs have a non-trivial positive correlation with each other. This could indicate there is a subset $Z'$ of $Z$ features that have little association with any of the $X$ proxies available in the UK Biobank. This violation could be remedied by having a richer set of biomarker proxies $X$.   

In Table \ref{table:infxzdim} we list the number of $X,Z$ proxies chosen for each $(D,Y)$ pair. 

\subappendix{All valid estimates after running our proxy selection algorithm}\label{appendix:7metrics_proxyrm}
After running the proxy selection algorithm, we found 34 $(D,Y)$ pairs that produced admissible proxy subsets $\{(\cX^{(i)}, \cZ^{(i)})\}_i$ which pass all five of our proposed tests and thus produce valid effect estimates for $\theta$. In Table \ref{table:proxyrm_alldy}, we show the results for the remaining 28 (the other six were presented in Table \ref{table:proxyrm}). We highlight strong bias estimates for ($D$=Female, $Y$=Migraine) with $\theta$=-0.14; ($D$=Female, $Y$=Chronic kidney disease) with $\theta$=-0.09; ($D$=Female, $Y$=Fibromyalgia) with $\theta$=-0.08; ($D$=No post-secondary education, $Y$=Osteoarthritis) with $\theta$=0.06; ($D$=Black, $Y$=Back pain) with $\theta$=0.16; ($D$=Black, $Y$=Fibromyalgia) with $\theta$=0.06; ($D$=Black, $Y$=Migraine) with $\theta$=0.08; ($D$=Obese, $Y$=Heart disease) with $\theta$=-0.06; ($D$=Obese, $Y$=COPD) with $\theta$=-0.08; ($D$=Asian, $Y$=Heart disease) with $\theta$=0.09; ($D$=Asian, $Y$=Depression) with $\theta$=0.32; and ($D$=Asian, $Y$=Back pain) with $\theta$=0.09.

% \usepackage{booktabs}
\begin{table}[H]
\resizebox{1.1
\textwidth}{!}{%
\begin{tabular}{@{}lcccccc@{}}
\toprule
\multicolumn{1}{c}{\textbf{$(D,Y)$}} & \multicolumn{1}{c}{\textbf{$\theta\pm95$\% CI}} & \multicolumn{1}{c}{\textbf{\begin{tabular}[c]{@{}c@{}}(1) Primal \\ statistic $<$ critical  \end{tabular}}} & \multicolumn{1}{c}{\textbf{\begin{tabular}[c]{@{}c@{}}(2) Dual \\ statistic $<$ critical  \end{tabular}}} & \multicolumn{1}{c}{\textbf{\begin{tabular}[c]{@{}c@{}}(3) $\E[\tilde{D}V] \neq 0$\\ statistic $>$ critical  \end{tabular}}} & \multicolumn{1}{c}{\textbf{\begin{tabular}[c]{@{}c@{}}(4) $V$ strength F-test \\ statistic $>$ critical  \end{tabular}}} & \multicolumn{1}{c}{\textbf{\begin{tabular}[c]{@{}c@{}}(5) Cov($\tilde{X}, \tilde{Z}$) \\ rank  \end{tabular}}} \\ \midrule
Low income, Endometriosis & 0.0$\pm$0.0 & 59.2\textless{}76.8 & 28.1\textless{}31.4 & 64.6\textgreater{}0.5 & 1947.8\textgreater{}23.1 & 2 \\
Disability insr., Back pain & 0.03$\pm$0.01 & 91.1\textless{}97.4 & 31.6\textless{}40.1 & 27.6\textgreater{}0.3 & 402.6\textgreater{}23.1 & 4 \\
Disability insr., Complications during labor & 0.0$\pm$0.0 & 63.3\textless{}66.3 & 15.9\textless{}25.0 & 20.7\textgreater{}0.4 & 739.5\textgreater{}23.1 & 2 \\
Not on private insr., Chronic kidney disease & 0.0$\pm$0.0 & 19.3\textless{}33.9 & 16.4\textless{}22.4 & 59.1\textgreater{}0.3 & 82373.3\textgreater{}23.1 & 2 \\
Not on private insr., Endometriosis & 0.0$\pm$0.0 & 51.3\textless{}58.1 & 7.4\textless{}11.1 & 43.4\textgreater{}0.3 & 3546.8\textgreater{}23.1 & 2 \\
No p.s. education, Osteoarthritis & 0.06$\pm$0.01 & 92.7\textless{}93.9 & 37.8\textless{}38.9 & 123.6\textgreater{}1.3 & 150.5\textgreater{}23.1 & 5 \\
No p.s. education, IBD & 0.0$\pm$0.0 & 80.1\textless{}97.4 & 39.6\textless{}42.6 & 119.7\textgreater{}1.3 & 91.4\textgreater{}23.1 & 4 \\
No p.s. education, COPD & 0.0$\pm$0.0 & 63.5\textless{}71.0 & 41.8\textless{}42.6 & 125.6\textgreater{}0.6 & 676.6\textgreater{}23.1 & 4 \\
Female, Fibromyalgia & -0.08$\pm$0.03 & 75.6\textless{}84.8 & 20.5\textless{}23.7 & 21.2\textgreater{}1.6 & 58.3\textgreater{}23.1 & 4 \\
Female, Chronic kidney disease & -0.09$\pm$0.04 & 61.9\textless{}71.0 & 19.6\textless{}21.0 & 22.0\textgreater{}2.1 & 42.1\textgreater{}23.1 & 4 \\
Female, Migraine & -0.14$\pm$0.06 & 76.3\textless{}79.1 & 22.7\textless{}23.7 & 20.8\textgreater{}1.7 & 43.5\textgreater{}23.1 & 4 \\
Female, Melanoma & 0.02$\pm$0.02 & 84.8\textless{}88.3 & 23.1\textless{}23.7 & 19.7\textgreater{}1.5 & 81.9\textgreater{}23.1 & 4 \\
Black, Heart disease & 0.1$\pm$0.03 & 97.7\textless{}115.4 & 25.5\textless{}31.4 & 9.7\textgreater{}0.2 & 216.8\textgreater{}23.1 & 4 \\
Black, Back pain & 0.16$\pm$0.04 & 84.0\textless{}93.9 & 15.4\textless{}21.0 & 10.1\textgreater{}0.3 & 68.7\textgreater{}23.1 & 5 \\
Black, Fibromyalgia & 0.06$\pm$0.02 & 68.6\textless{}72.2 & 16.1\textless{}21.0 & 9.8\textgreater{}0.3 & 44.4\textgreater{}23.1 & 4 \\
Black, IBD & 0.01$\pm$0.01 & 56.6\textless{}84.8 & 20.8\textless{}21.0 & 9.4\textgreater{}0.3 & 28.0\textgreater{}23.1 & 4 \\
Black, Migraine & 0.08$\pm$0.01 & 104.3\textless{}105.3 & 18.1\textless{}21.0 & 10.9\textgreater{}0.2 & 118.3\textgreater{}23.1 & 5 \\
Black, Melanoma & -0.02$\pm$0.01 & 84.6\textless{}84.8 & 13.3\textless{}21.0 & 10.2\textgreater{}0.3 & 73.1\textgreater{}23.1 & 4 \\
Obese, Heart disease & -0.06$\pm$0.01 & 112.9\textless{}113.1 & 24.8\textless{}30.1 & 67.4\textgreater{}1.6 & 184.1\textgreater{}23.1 & 4 \\
Obese, Fibromyalgia & 0.01$\pm$0.01 & 56.3\textless{}72.2 & 28.5\textless{}28.9 & 63.9\textgreater{}1.6 & 156.0\textgreater{}23.1 & 4 \\
Obese, COPD & -0.08$\pm$0.01 & 95.5\textless{}97.4 & 24.1\textless{}27.6 & 69.0\textgreater{}1.4 & 293.7\textgreater{}23.1 & 4 \\
Asian, Heart disease & 0.09$\pm$0.02 & 108.1\textless{}109.8 & 23.6\textless{}27.6 & 14.2\textgreater{}0.3 & 163.1\textgreater{}23.1 & 5 \\
Asian, Depression & 0.32$\pm$0.05 & 56.2\textless{}73.3 & 14.6\textless{}23.7 & 14.4\textgreater{}0.2 & 54.4\textgreater{}23.1 & 4 \\
Asian, Back pain & 0.09$\pm$0.02 & 96.0\textless{}100.7 & 23.7\textless{}23.7 & 14.4\textgreater{}0.2 & 211.7\textgreater{}23.1 & 4 \\
Asian, Rh. Arthritis & -0.0$\pm$0.0 & 65.9\textless{}66.3 & 11.4\textless{}15.5 & 14.9\textgreater{}0.2 & 689.8\textgreater{}23.1 & 4 \\
Asian, Fibromyalgia & 0.04$\pm$0.01 & 57.3\textless{}61.7 & 15.6\textless{}16.9 & 14.0\textgreater{}0.3 & 102.3\textgreater{}23.1 & 2 \\
Asian, Chronic kidney disease & 0.01$\pm$0.02 & 45.5\textless{}48.6 & 14.4\textless{}16.9 & 14.3\textgreater{}0.3 & 38.1\textgreater{}23.1 & 3 \\
Asian, Endometriosis & 0.02$\pm$0.01 & 86.1\textless{}88.3 & 31.3\textless{}32.7 & 9.9\textgreater{}0.2 & 208.4\textgreater{}23.1 & 3 \\ \bottomrule
\end{tabular}
}
\caption{The remaining 28 valid UK Biobank implicit bias effect estimates after applying our $X,Z$ proxy selection algorithm. Tests (1-5) are detailed in Section \ref{main:tests}, where \textit{statistic} is the given data's statistic and \textit{critical} is the necessary critical value to be greater or less than to pass. $V = \tilde{D}-\gamma^\top \tilde{Z}$. COPD = Chronic obstructive pulmonary disease. IBD = Inflammatory bowel disease. p.s. = post-secondary. insr. = insurance.}
\label{table:proxyrm_alldy}
\end{table}

\begin{figure}[ht]
\centerline{\includegraphics[width=.9\textwidth]{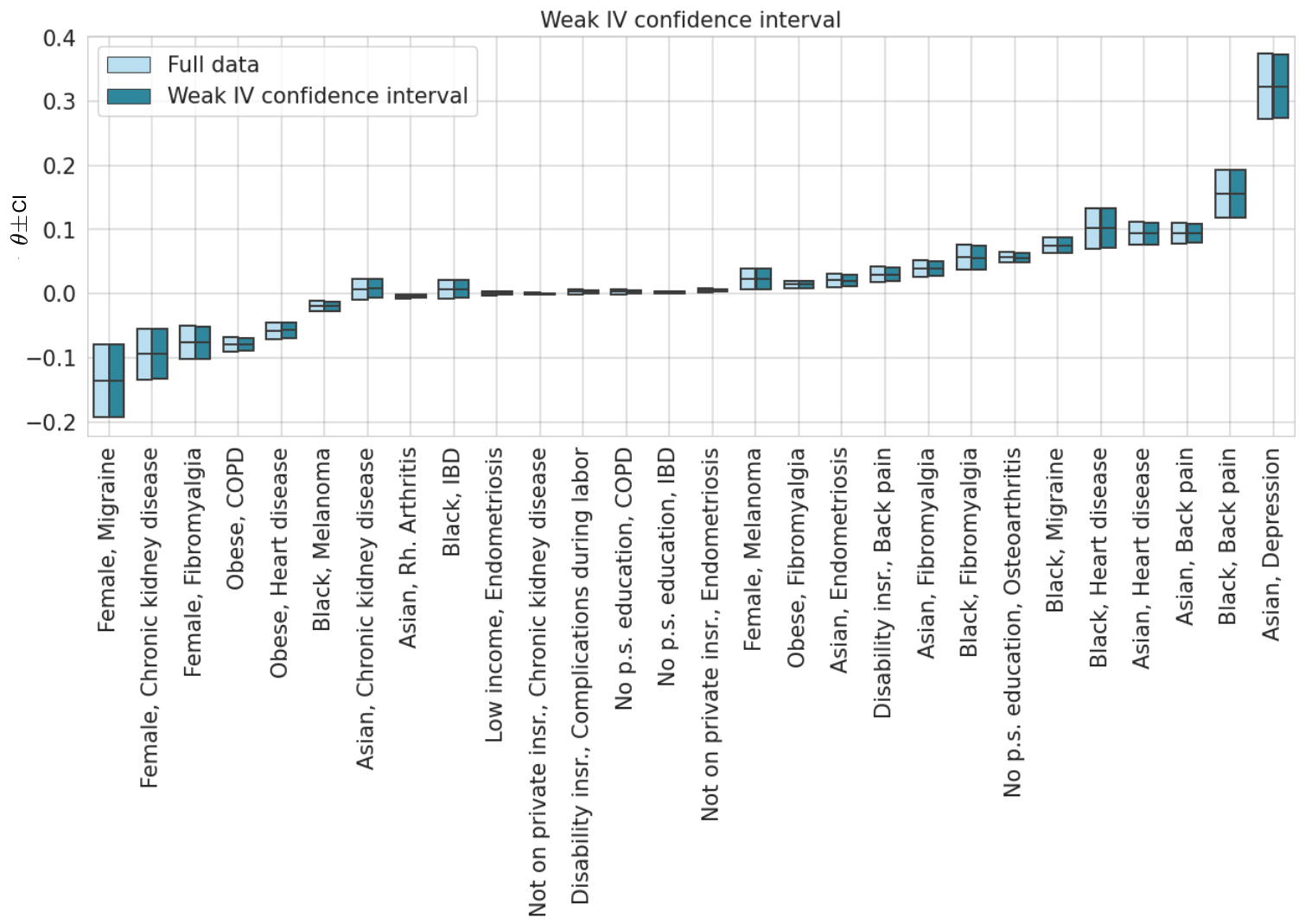}}
\caption{Weak instrumental variable (IV) confidence interval test. COPD = Chronic obstructive pulmonary disease. IBD = Inflammatory bowel disease. p.s. = post-secondary. insr. = insurance.}
\label{fig:weakiv_all}
\end{figure}

\subappendix{Weak instrument confidence interval for all valid estimates}
We ran the weak instrument confidence interval test on the remaining 28 $(D,Y)$ pairs after proxy removal. We see in Table \ref{fig:weakiv_all} that the test yields intervals consistent with those from our method, demonstrating our estimate is robust to weak instruments. 

\subappendix{Additional bootstrap analysis }\label{appendix:bs_samplesize}

\begin{figure}[ht]
\centerline{\includegraphics[width=\textwidth]{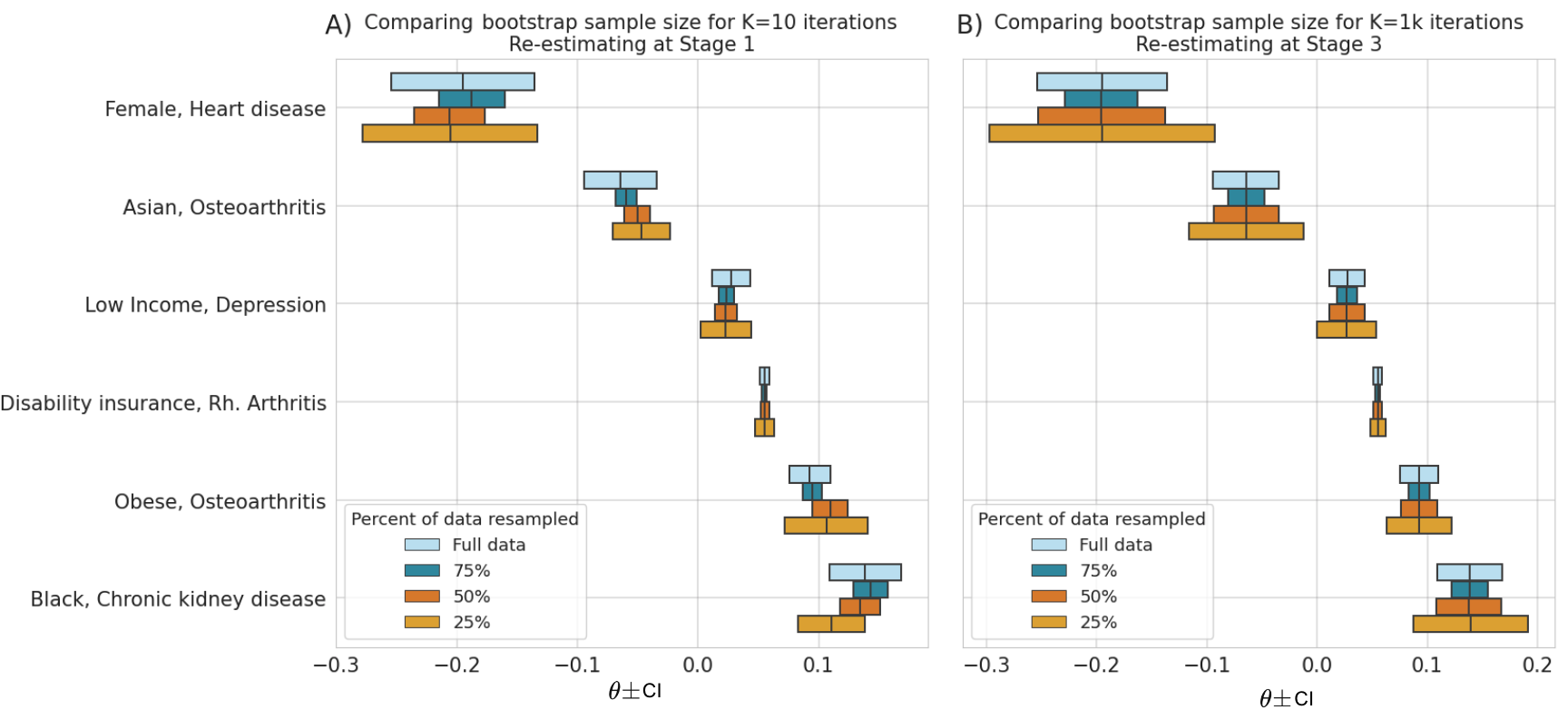}}
\caption{Bootstrap analyses for six $(D,Y)$ pairs. A) Effect of sample size on bootstrapped estimate for $K$=10 iterations re-estimating from stage 1. B) Effect of sample size on bootstrapped estimate for $K$=1000 iterations re-estimating from stage 3.}
\label{fig:bs_p2}
\end{figure}

Here we show the two other bootstrapping analyses on the original six $(D,Y)$ pairs and mentioned in Section \ref{main:bs}. In Figure \ref{fig:bs_p2}A we show the influence of sampling $10\%, 25\%, 50\%$ or $75\%$ of the original data size for $K=10$ bootstrapped iterations, re-estimating over the full pipeline (stage 1). Similarly, in Figure \ref{fig:bs_p2}B we show the influence of sampling $10\%, 25\%, 50\%$ or $75\%$ of the original data size for $K=1000$ bootstrapped iterations, re-estimating over the final estimate only (stage 3). We see relatively consistent estimates regardless of sample size and re-estimation stage. As expected, the confidence interval increases as sample size decreases. 

\subappendix{Additional influence points results}
In Table \ref{table:infxzdim}, we list the size of the high-influence set for each $(D,Y)$ pair such that the estimated estimate after the set's removal is $\theta_{\text{inf}}=0$. Most of the sizes of the minimally influential set are decently small (recall $n=502411$).

\begin{figure}[ht]
\centerline{\includegraphics[width=1.1\textwidth]{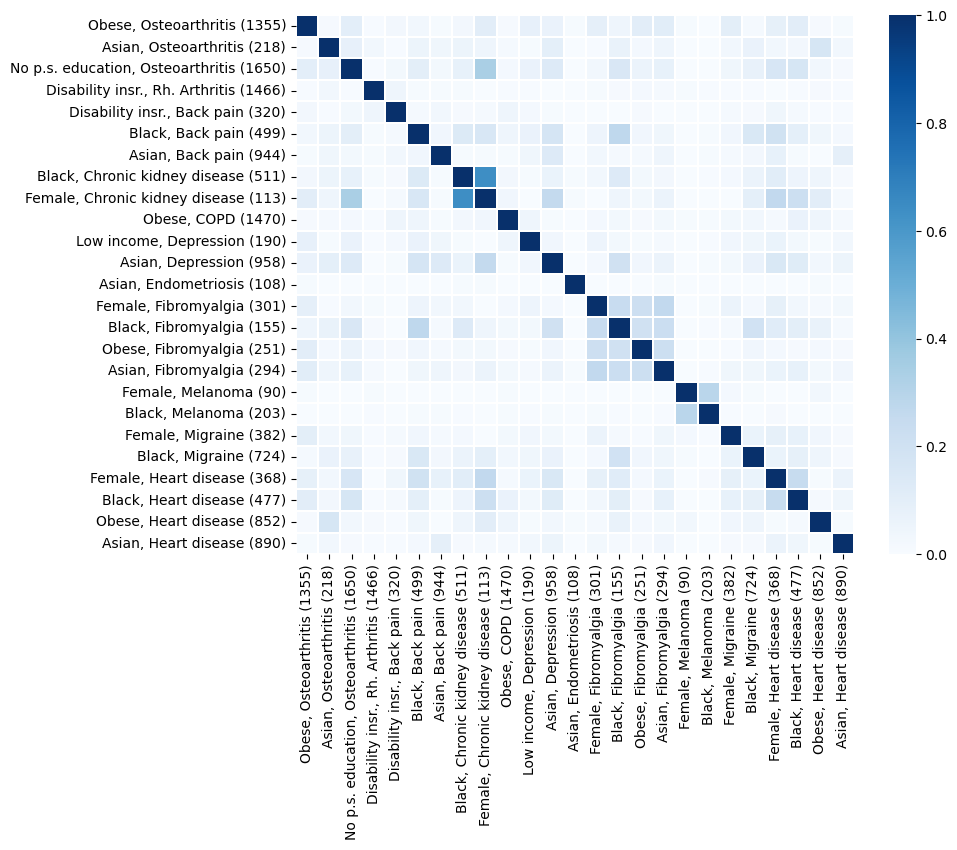}}
\caption{Heatmap of the Szymkiewicz-Simpson coefficient across all ($D,Y$) pairs' high-influence sets for which $\mid \theta \mid > 0.01$. The set size of the high-influence set is reported in parentheses. COPD = Chronic obstructive pulmonary disease. IBD = Inflammatory bowel disease. p.s. = post-secondary. insr. = insurance.}
\label{fig:heatinf}
\end{figure}

We perform two interpretability analyses to better understand these results. First, we compare high-influence sets to see if similar points are being scored as highly influential across $(D,Y)$ pairs. We take the Szymkiewicz-Simpson coefficient ($\frac{\mid A \cap B \mid}{\text{min}(\mid A \mid, \mid B \mid)}$) for the high-influence set across all $(D,Y)$ pairs where $\mid \theta \mid > .01$  The resulting heatmap in Figure \ref{fig:heatinf} shows a potential pattern for overlap among diagnoses $Y$, particularly for $Y\in$ [Fibromyalgia, Chronic kidney disease, Melanoma]. This could potentially indicate that, for these $Y$ diagnoses, the same ``outlier'' patients are driving the detected implicit biases.

\begin{figure}[ht]
\centerline{\includegraphics[width=.9\textwidth]{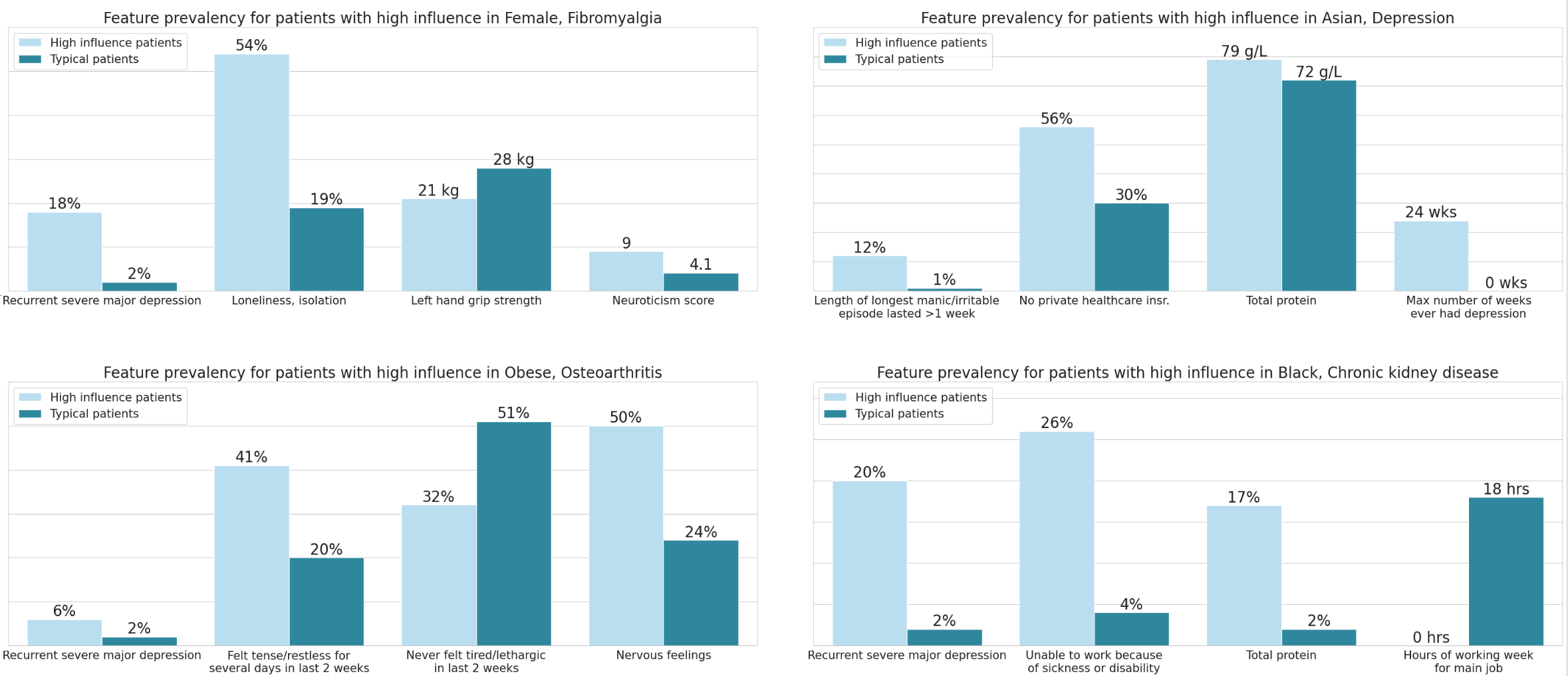}}
\caption{Feature prevalency between high-influence patients and the rest of the data for four $(D,Y$) pairs. Four features were selected randomly for visualization. Categorical or binary features are aggregated by mean, continuous features are by median.}
\label{fig:inf2}
\end{figure}

To investigate what phenotypes characterize these identified high-influence patients, we perform a simple feature interpretability analysis, as described in Appendix \ref{appendix:infint}. Similar to Figure \ref{fig:inf}A, we report significant features between high-influence set versus the rest of the data for several other $(D,Y)$ pairs in Figure \ref{fig:inf2}. These features discrepancies could be potential phenotypes that characterize ``outlier'' patient groups that drive the implicit bias effect estimate. 
\begin{table}[ht]
\centering
\resizebox{.9\textwidth}{!}{%
\begin{tabular}{@{}lcccc@{}}
\toprule
\multicolumn{1}{c}{\textbf{($D, Y$)}} & \multicolumn{1}{c}{\textbf{$\theta\pm 95\%$ CI}} & \multicolumn{1}{c}{\textbf{Inf. set size}} & \multicolumn{1}{c}{\textbf{\#$Z$ proxies}} & \multicolumn{1}{c}{\textbf{\#$X$ proxies}} \\ \midrule
Low income, Depression & 0.03$\pm$0.02 & 190 & 43 & 27 \\
Disability insr., Rh. Arthritis & 0.06$\pm$0.0 & 1466 & 56 & 5 \\
Female, Heart disease & -0.19$\pm$0.06 & 368 & 94 & 14 \\
Black, Chronic kidney disease & 0.14$\pm$0.03 & 511 & 41 & 12 \\
Obese, Osteoarthritis & 0.09$\pm$0.02 & 1355 & 78 & 18 \\
Asian, Osteoarthritis & -0.06$\pm$0.03 & 218 & 79 & 22 \\
Low income, Endometriosis & 0.0$\pm$0.0 & 9 & 57 & 20 \\
Disability insr., Back pain & 0.03$\pm$0.01 & 320 & 75 & 27 \\
Disability insr., Complications during labor & 0.0$\pm$0.0 & 30 & 48 & 15 \\
Not on private insr., Chronic kidney disease & 0.0$\pm$0.0 & 8 & 21 & 13 \\
Not on private insr., Endometriosis & 0.0$\pm$0.0 & 148 & 41 & 5 \\
No p.s. education, Osteoarthritis & 0.06$\pm$0.01 & 1650 & 72 & 26 \\
No p.s. education, IBD & 0.0$\pm$0.0 & 23 & 75 & 29 \\
No p.s. education, COPD & 0.0$\pm$0.0 & 78 & 52 & 29 \\
Female, Fibromyalgia & -0.08$\pm$0.03 & 301 & 64 & 14 \\
Female, Chronic kidney disease & -0.09$\pm$0.04 & 113 & 52 & 12 \\
Female, Migraine & -0.14$\pm$0.06 & 382 & 59 & 14 \\
Female, Melanoma & 0.02$\pm$0.02 & 90 & 67 & 14 \\
Black, Heart disease & 0.1$\pm$0.03 & 477 & 91 & 20 \\
Black, Back pain & 0.16$\pm$0.04 & 499 & 72 & 12 \\
Black, Fibromyalgia & 0.06$\pm$0.02 & 155 & 53 & 12 \\
Black, IBD & 0.01$\pm$0.01 & 18 & 64 & 12 \\
Black, Migraine & 0.08$\pm$0.01 & 724 & 82 & 12 \\
Black, Melanoma & -0.02$\pm$0.01 & 203 & 64 & 12 \\
Obese, Heart disease & -0.06$\pm$0.01 & 852 & 89 & 19 \\
Obese, Fibromyalgia & 0.01$\pm$0.01 & 251 & 53 & 18 \\
Obese, COPD & -0.08$\pm$0.01 & 1470 & 75 & 17 \\
Asian, Heart disease & 0.09$\pm$0.02 & 890 & 86 & 17 \\
Asian, Depression & 0.32$\pm$0.05 & 958 & 54 & 14 \\
Asian, Back pain & 0.09$\pm$0.02 & 944 & 78 & 14 \\
Asian, Rh. Arthritis & 0.0$\pm$0.0 & 130 & 48 & 8 \\
Asian, Fibromyalgia & 0.04$\pm$0.01 & 294 & 44 & 9 \\
Asian, Chronic kidney disease & 0.01$\pm$0.02 & 4 & 33 & 9 \\
Asian, Endometriosis & 0.02$\pm$0.01 & 108 & 67 & 21 \\ \bottomrule
\end{tabular}
}
\caption{All 34 valid UK Biobank implicit bias effect estimates with the size of the minimum highly-influential set such that its removal leads to an estimate $\theta_{\text{inf}}=0$. We also show the number of $Z$ (out of 112) and $X$ (out of 65) proxy features selected by the proxy removal algorithm in producing a valid effect estimate. COPD = Chronic obstructive pulmonary disease. IBD = Inflammatory bowel disease. p.s. = post-secondary. insr. = insurance.}
\label{table:infxzdim}
\end{table}

\begin{figure}[ht]
\centerline{\includegraphics[width=.50\textwidth]{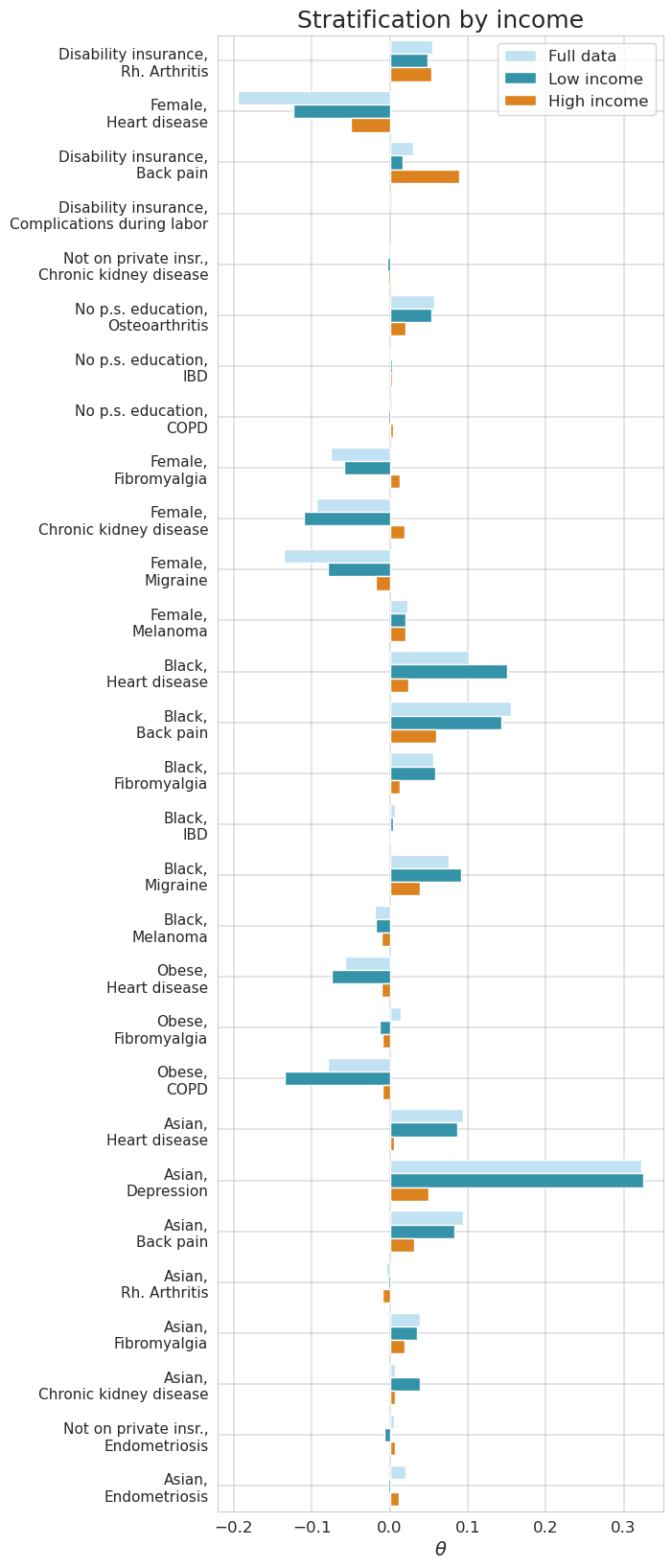}}
\caption{Point re-estimates from stage 2 after stratification by income. COPD = Chronic obstructive pulmonary disease. IBD = Inflammatory bowel disease. p.s. = post-secondary. insr. = insurance.}
\label{fig:incstrat}
\end{figure}

\subappendix{Additional income stratification results}
In Figure \ref{fig:inf}B we show the intersectional effect of income on implicit bias for three ($D,Y$) pairs where $D\neq$Income. We re-estimate the new estimate $\theta$ from stage 2 of the pipeline (see Section \ref{main:bs}) on the patients corresponding to one of two income strata: Low income=Average total household income before tax is less than 18,000\pounds and High income=Average total household income before tax is greater than 100,000\pounds. We report the remaining $(D,Y)$ pairs in Figure \ref{fig:incstrat} and show there are variations in implicit bias estimates based on income strata.

\subappendix{Partial non-linearity of $W$}\label{appendix:nonlinear}
We model the residuals $\E[V | W]$ for $V \in [Y,D,X,Z]$ using XGBoost regression\cite{xgb} where the learning rate is chosen via semi-cross fitting. For the six $(D,Y)$ pairs we use the same admissible proxy sets for $X$ and $Z$ as mentioned previously and recompute the estimate $\theta$ and the 95\% confidence interval using the new XGBoost residuals. In Table \ref{table:xgb2}, we see the XGBoost residuals perform equivalently to using Lasso residuals for our data. 

\begin{table}[ht]
\centering
\resizebox{.7\textwidth}{!}{%
\begin{tabular}{@{}lcc@{}}
\toprule
\multicolumn{1}{c}{\textbf{($D, Y$)}} & \multicolumn{1}{c}{\textbf{\begin{tabular}[c]{@{}c@{}}$\theta\pm95\%$ CI \\ $\E[ \cdot \mid W]$ = Lasso \end{tabular}}} & \multicolumn{1}{c}{\textbf{\begin{tabular}[c]{@{}c@{}}$\theta\pm95\%$ CI\\ $\E[ \cdot \mid W]$ = XGBoost \end{tabular}}}\\ 
\midrule
Low income, Depression & 0.03$\pm$0.02 & \hspace{10pt}0.03$\pm$0.01 \\
Disability insurance, Rh. Arthritis & 0.06$\pm$0.00 & \hspace{10pt}0.04$\pm$0.00 \\
Female, Heart disease & -0.19$\pm$0.06 & \hspace{10pt} -0.17$\pm$0.04 \\
Black, Chronic kidney disease & 0.14$\pm$0.03 & \hspace{10pt}0.10$\pm$0.03 \\
Obese, Osteoarthritis & 0.09$\pm$0.02 & \hspace{10pt} 0.08$\pm$0.02 \\
Asian, Osteoarthritis & -0.06$\pm$0.03 & \hspace{10pt} -0.06$\pm$0.03 \\ \bottomrule
\end{tabular}}
\caption{Comparing Lasso versus XGBoost regression for estimating the residuals of confounders $W$ for six UK Biobank $(D,Y)$ pairs. We use the same $X,Z$ proxy selection sets.}
\label{table:xgb2}
\end{table}

\end{document}